\documentclass[final]{cvpr}

\usepackage{times}
\usepackage{epsfig}
\usepackage{graphicx}
\usepackage{amsmath}
\usepackage{amssymb}
\usepackage{booktabs}
\usepackage{algorithm}
\usepackage{algorithmic}
\usepackage{multirow}
\usepackage{subfigure}
\usepackage{amsthm}
\usepackage{url}
\usepackage{rotate}
\usepackage{bm}
\usepackage{setspace}
\usepackage{color}

\def\bomega{\bm{\omega}}
\def\balpha{\bm{\alpha}}

\def\ttt{\mathtt{t}}
\def\tn{\mathtt{n}}

\usepackage[pagebackref=true,breaklinks=true,colorlinks,bookmarks=false]{hyperref}

\begin{document}

\title{Retinex-inspired Unrolling with Cooperative Prior Architecture Search \\for Low-light Image Enhancement}

\author{Risheng Liu$^{1,3}$, Long Ma$^{2,3}$, Jiaao Zhang$^{2,3}$, Xin Fan$^{1,3}$, Zhongxuan Luo$^{1,3}$\\
 \normalsize $^1$International School of Information Science \& Engineering, Dalian University of Technology\\
 \normalsize$^2$School of Software Technology, Dalian University of Technology\\
 \normalsize $^3$Key Laboratory for Ubiquitous Network and Service Software of Liaoning Province
\\
 {\tt \small \{rsliu, xin.fan, zxluo\}@dlut.edu.cn, \{longma, jiaaozhang\}@mail.dlut.edu.cn}\
}

\maketitle

\begin{abstract}
	Low-light image enhancement plays very important roles in low-level vision field. Recent works have built a large variety of deep learning models to address this task. However, these approaches mostly rely on significant architecture engineering and suffer from high computational burden. In this paper, we propose a new method, named Retinex-inspired Unrolling with Architecture Search (RUAS), to construct lightweight yet effective enhancement network for low-light images in real-world scenario.
	Specifically, building upon Retinex rule, RUAS first establishes models to characterize the intrinsic underexposed structure of low-light images and unroll their optimization processes to construct our holistic propagation structure. Then by designing a cooperative reference-free learning strategy to discover low-light prior architectures from a compact search space, RUAS is able to obtain a top-performing image enhancement network, which is with fast speed and requires few computational resources. Extensive experiments verify the superiority of our RUAS framework against recently proposed state-of-the-art methods. 
\end{abstract}

\begin{figure}[t]
	\begin{center}
		\begin{tabular}{c@{\extracolsep{0.7em}}c}
			\includegraphics[width=0.467\linewidth]{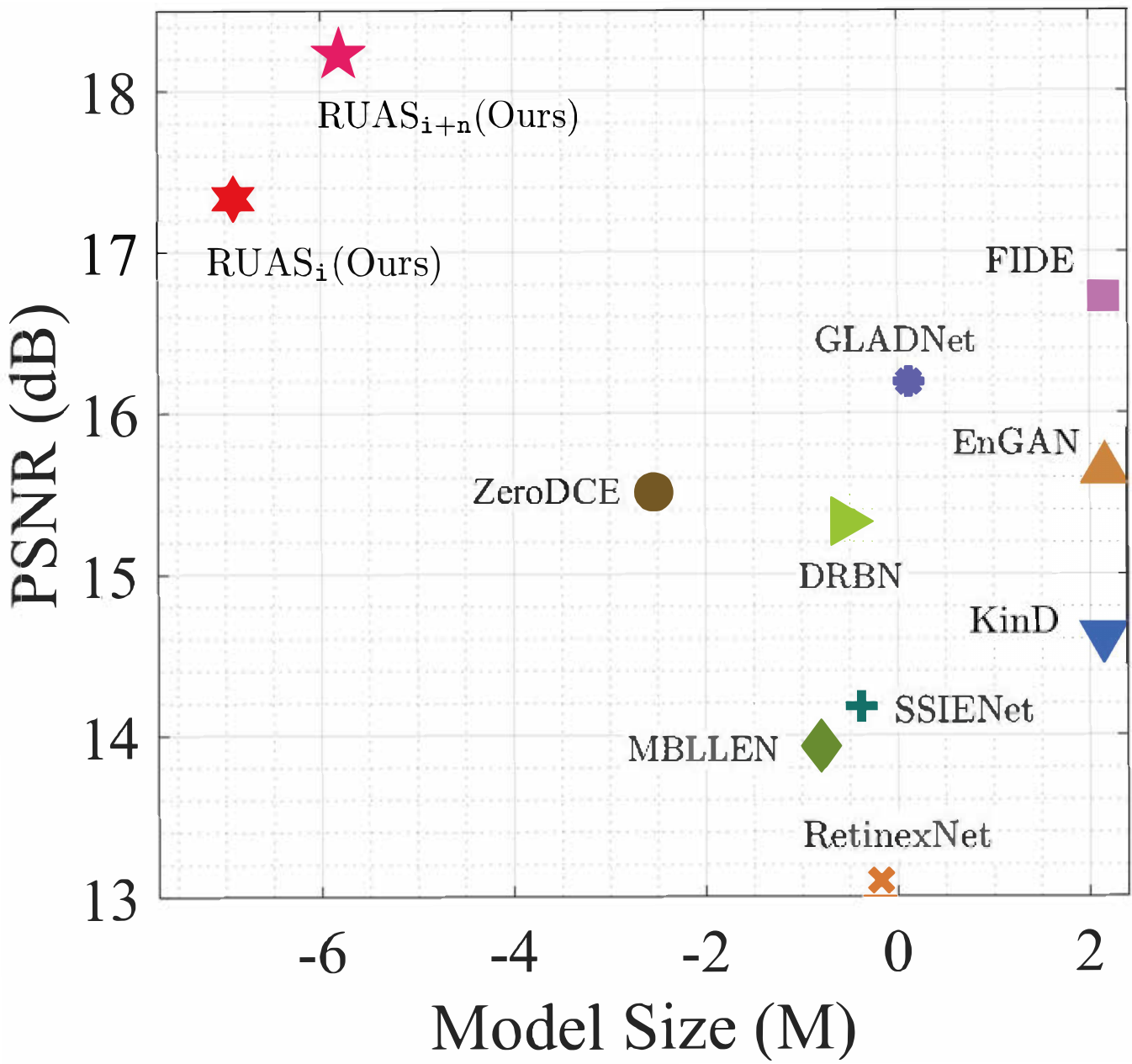}&
			\includegraphics[width=0.467\linewidth]{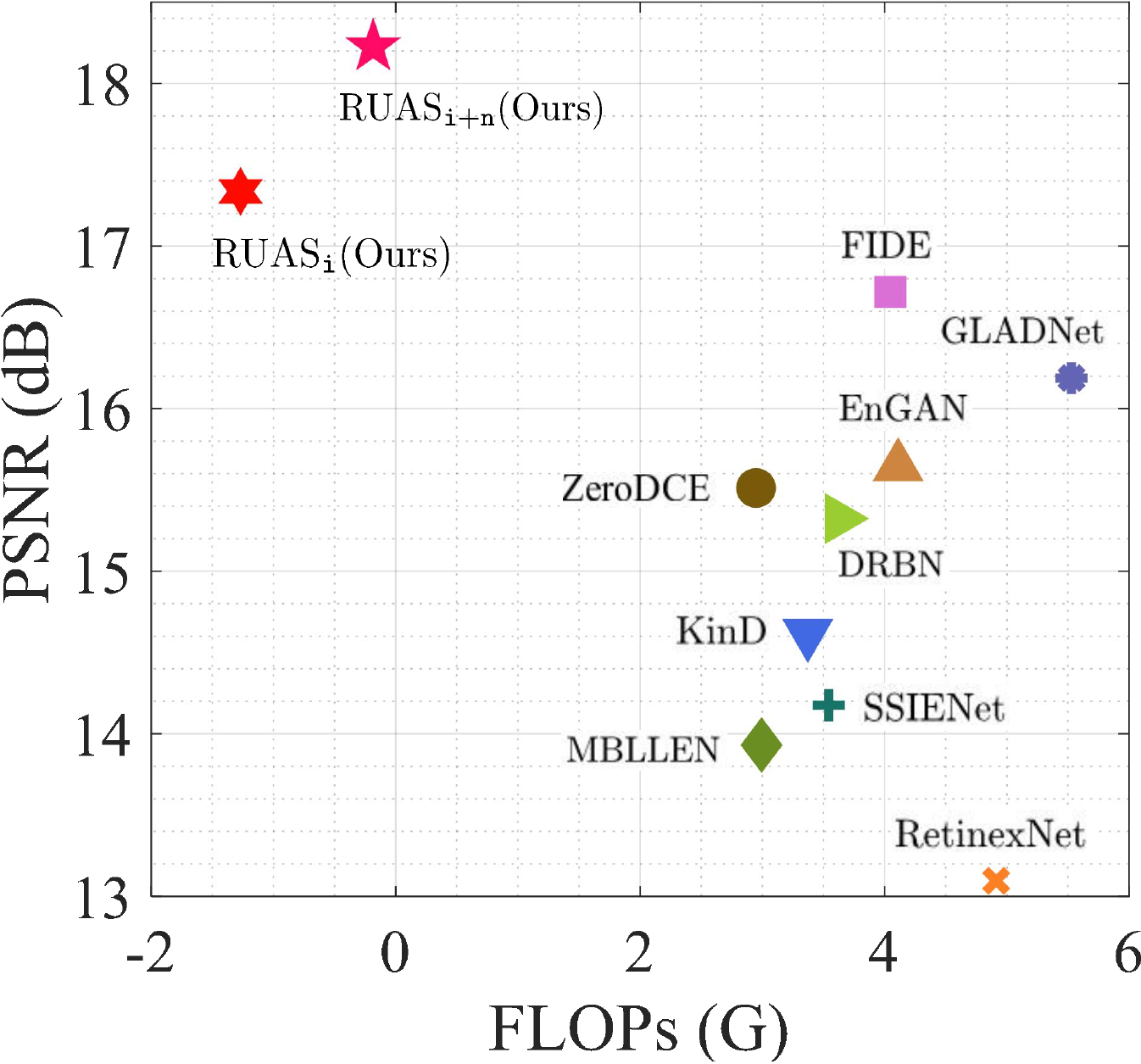}\\
			\footnotesize (a) & \footnotesize (b) \\
		\end{tabular}
	\end{center}
	\vspace{-0.25cm}
	\caption{Averaged quantitative performance (PSNR) vs model size (a), FLOPs (b) on LOL dataset~\cite{Chen2018Retinex}. We plot results of our methods (i.e, RUAS$_{\mathtt{i}}$ and RUAS$_{\mathtt{i}+\mathtt{n}}$) and some recently proposed state-of-the-art CNN-based approaches, including MBLLEN~\cite{lv2019attention}, GLADNet~\cite{wang2018gladnet}, RetinexNet~\cite{Chen2018Retinex}, EnGAN~\cite{jiang2019enlightengan}, SSIENet~\cite{zhang2020self}, KinD~\cite{zhang2019kindling}, ZeroDCE~\cite{guo2020zero}, FIDE~\cite{xu2020learning}, and DRBN~\cite{yang2020fidelity}. Here a log scale is used on the x-axis for illustration.}
	\label{fig:FirstFigure}
\end{figure}

\section{Introduction}
High quality images are critical to a large amount of computer vision and machine learning applications, such as object detection~\cite{liu2020deep}, segmentation~\cite{yu2020context} and recognition~\cite{zhang2019two}, just name a few. Unfortunately, images captured in low-light environments usually suffer multiple degradations, including poor visibility, low contrast and unexpected noise, etc. Therefore, it is necessary to enhance those low-light images before further processing and analysis. Existing Low-light Image Enhancement (LIE) techniques can be roughly divided into two major categories: classical methods and deep learning methods. 

In the past decades, classical methods often perform histogram equalization~\cite{sheet2010brightness,cheng2004simple,yun2011Contrast} or gamma correction~\cite{huang2012efficient,singh2017novel,wang2019variational} to enhance low-light images. There are also various classical methods consider Retinex theory~\cite{rahman2004retinex} and introduce different prior regularized optimization models to characterize the structures of the illumination and reflectance image layers~\cite{fu2015probabilistic,guo2017lime,li2018structure,zhang2020enhancing}. However, these hand-crafted constraints/priors are not adaptive enough and their results may present intensive noises and/or suffer from over- and under- enhancement. 

In recent years, great progress has also been made on designing CNN-based models for LIE problems. Most CNN-based solutions rely on paired data for supervised training~\cite{Chen2018Retinex,zhang2019kindling,wang2019underexposed,xu2020learning}. A few methods also train their networks without paired supervision~\cite{yang2020fidelity,jiang2019enlightengan,guo2020zero}. However, the performances of these deep learning methods heavily rely on their elaborately designed architectures and carefully selected paired/unpaired training data. Moreover, most of these existing CNN-based methods tend to obtain unsatisfactory visual results, when presented with real-world images with various light intensities and intensive noises. This is because the fundamental model they rely on is lacking in principled physical constraints, thus is hard to capture the intrinsic low-light image structure. 

To partially address the above issues, this work proposes Retinex-inspired Unrolling with Architecture Search (RUAS), a principled framework to construct enhancement networks by infusing knowledge of low-light images and searching lightweight prior architectures.  More concretely, taking Retinex rule into consideration, we first design optimization models to exploit the latent structures of the low-light image in the real-world noisy scenario. Then by unrolling the corresponding optimization processes, we establish the holistic propagation structure of our enhancement network. Finally, we provide a reference-free bilevel learning strategy to cooperatively search prior architectures for the illumination map and desired image. Our contributions can be summarized as follows:
\begin{itemize}
	\item In contrast to existing CNN-based LIE methods, that require substantial efforts to heuristically design the whole neural network, RUAS first provides a principled manner to build our fundamental network structure and then automatically discover the embedded atomic prior architectures.
	
	\item 
	We develop a cooperative bilevel search strategy for RUAS, which is able to simultaneously discover architectures from a compact search space for both illumination estimation and noise removal. Furthermore, our strategy does not require any paired/unpaired supervisions during the search process.
	
	
	\item RUAS offers flexibility in searching prior architectures for different kinds of low-light scenarios. Extensive experiments also show that our established enhancement networks are memory and computation efficient, and can perform favorably against state-of-the-art approaches (see Fig.~\ref{fig:FirstFigure}). 
\end{itemize}

\begin{figure*}[t]
	\begin{center}
		\begin{tabular}{c}
			\includegraphics[width=0.95\linewidth]{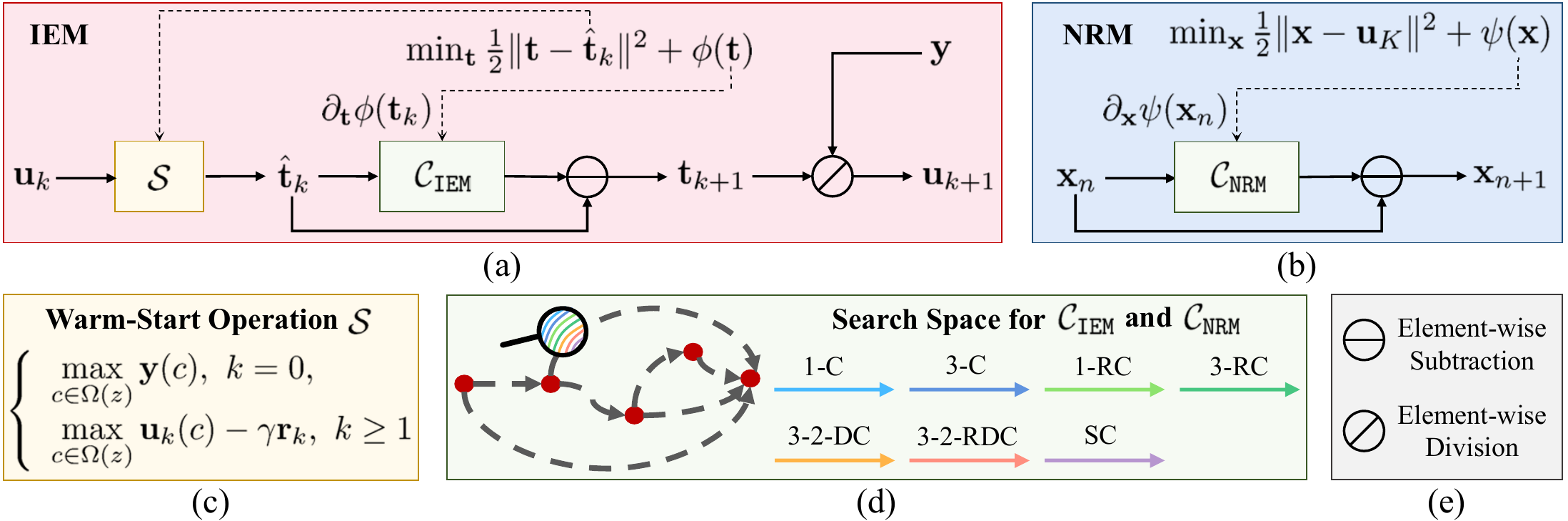}\\
		\end{tabular}	
	\end{center}
	\vspace{-0.15cm}
	\caption{Illustrations of the fundamental components for RUAS. On the top row, we plot the diagrams of IEM (a) and NRM (b). On the bottom row, we show the warm-start operation (c), the search space for $\mathcal{C}_{\mathtt{IEM}}$ and $\mathcal{C}_{\mathtt{NRM}}$ (d) and two element-wise operations (e).}
	\vspace{-0.1cm}
	\label{fig:flowchart}
\end{figure*}

\section{Related Work}

\subsection{CNNs for Low-light Image Enhancement}
In recent years, with the development of deep learning, the problem of LIE has achieved a significant performance boost. A variety of CNN architectures have been designed for solving the LIE problem. For example,
LLNet~\cite{lore2017llnet} utilized a variant of the stacked sparse denoising autoencoder to brighten the low-light images. 
EnGAN~\cite{jiang2019enlightengan} designed an attention module on U-Net~\cite{ronneberger2015u} and can be trained with
only low/normal-light images (unnecessarily paired). The paper in~\cite{xu2020learning} developed a frequency-based decomposition-and-enhancement network based on the attention to context encoding module. RetinexNet~\cite{Chen2018Retinex} combined the Retinex theory with CNNs to estimate the illumination map and enhance the low-light images. KinD~\cite{zhang2019kindling} designed a similar network but connected the feature-level illumination and reflectance in the decomposition step. Wang~\emph{et al.}~\cite{wang2019underexposed} designed an image-to-illumination network architecture based on the bilateral learning framework. Zhang~\emph{et al.}~\cite{zhang2020self} established a self-supervised CNN to simultaneously output the illumination and reflectance. The work in~\cite{guo2020zero} proposed a zero-reference curve estimation CNN to address the LIE task. A recursive band network was proposed in~\cite{yang2020fidelity} and trained by a semi-supervised strategy. Nonetheless, discovering state-of-the-art neural network architectures requires
substantial efforts.

\subsection{Neural Architecture Search (NAS)} 

In past years,  NAS has attracted increasing attentions due to the potential of finding effective and/or efficient architectures that outperform human expertise. Early attempts~\cite{zoph2016neural,so2019evolved} use evolutionary algorithms to optimize architectures and parameters of the networks. Another group of approaches~\cite{cheng2020instanas,tan2020efficientdet} employ reinforcement learning techniques to train a meta-controller to generate potential architectures. However both of these two kinds of methods require a large amount of computations, thus are inefficient in search. Recently, differentiable search methods~\cite{liu2018darts,xu2019pc,wan2020fbnetv2} formulated the super-network into a differentiable form with respect to a set of architectural parameters,  so that they can be optimized by gradient descent and the search cost can be reduced to several hours. The above search algorithms have achieved highly competitive performance in various high-level vision tasks, including image classification~\cite{xu2019pc}, object detection~\cite{wang2020fcos}, and semantic segmentation. Very recently, NAS algorithms~\cite{mozejko2020superkernel,zhang2020memory,li2020all} have also been applied to low-level vision problems, including image denoising, restoration, and deraining, etc. Unfortunately, existing NAS strategies are fully data-driven, thus require a large number of well-prepared paired training data, which is generally impractical for the LIE. Furthermore, due to the lack of principled prior knowledge,
architectures searched by aforementioned methods are still not suitable to exploit the complex statistics of low-light images.

\section{The Proposed Method}

We first establish the enhancement network by unrolling the optimization process of Retinex-inspired models. Then we introduce a distillation cell based search space for the prior modules. Finally, a cooperative bilevel search strategy is proposed to discover desired  architectures for illumination estimation and noise removal.


\subsection{Retinex-Inspired Optimization Unrolling}

Our PUAS enhancement network is built upon the following simple Retinex rule $\mathbf{y} = \mathbf{x}\otimes\mathbf{t}$, where $\mathbf{y}$ and $\mathbf{x}$ are the captured underexposed observation and the desired recovery, respectively. Furthermore, $\mathbf{t}$ denotes the illumination map and the operator $\otimes$ represents the element-wise multiplication. As shown in Fig.~\ref{fig:flowchart} (a) and (b), Illumination Estimation Module (IEM) is devised to estimate the illumination map $\mathbf{t}$ and  
Noise Removal Module (NRM) is designed to suppress noise in some challenging low-light scenario. We next detail these two key components in the following subsections.

%
%

\subsubsection{Illumination Estimation Module (IEM)}


Define an intermediate image $\mathbf{u}$. At $k$-th stage of IEM, we first propose the following strategy to estimate an initial illumination map $\hat{\mathbf{t}}_k$, i.e., $\hat{\mathbf{t}}_k=\mathcal{S}(\mathbf{u}_k)$ with
\begin{equation}
\mathcal{S}(\mathbf{u}_k):=\left\{\begin{array}{l}
\max_{c\in\Omega(z)}\mathbf{y}(c), \ k=0,\\
\max_{c\in\Omega(z)}\mathbf{u}_k(c)-\gamma\mathbf{r}_k, \ k\geq 1.
\end{array}\right.
\end{equation}
Here $\mathbf{u}_k$ is obtained by $\mathbf{u}_k=\mathbf{y}\oslash\mathbf{t}_k$, where $\mathbf{t}_k$ is the estimated illumination map in the last stage and $\oslash$ denotes the element-wise division. 
Furthermore, $\Omega(z)$ is a region centered at pixel $z$ and $c$ is the location index within this region (for three color channels). The principle behind this term is that the illumination is at least the maximal value of a certain location and can be used to handle non-uniform illuminations.
As for the residual $\mathbf{r}_k=\mathbf{u}_k-\mathbf{y}$ (with penalty parameter $0<\gamma\leq1$), we actually introduce this term to adaptively suppress some overexposed pixels for $\hat{\mathbf{t}}_{k}$ during the propagation. 

With the illumination warm-start $\hat{\mathbf{t}}_k$, we further refine $\mathbf{t}$ by solving the following model:
$\min_{\mathbf{t}}\frac{1}{2}\|\mathbf{t}-\hat{\mathbf{t}}_k\|^2+\phi(\mathbf{t}),$
where $\phi(\cdot)$ represents a regularization term of $\mathbf{t}$. Different from classical iterative optimization methods, which interact with the prior term directly, we just write a schematic gradient descent scheme\footnote{Please notice that here we just skip the learning rate (i.e.,  set it as 1).}
\begin{equation}
\mathbf{t}_{k+1}=\hat{\mathbf{t}}_k -\partial_{\mathbf{t}}\phi(\mathbf{t}_k),\label{eq:update_t}
\end{equation}
and parameterize $\partial_{\mathbf{t}}\phi(\mathbf{t}_k)$ by a CNN architecture $\mathcal{C}_{\mathtt{IEM}}({\mathbf{t}}_k)$.
By performing $K$ stages of the above calculations, we can obtain $\mathbf{u}_K=\mathbf{y}\oslash\mathbf{t}_K$ as the output of IEM.
Indeed, the choice of parameterizing each iteration as a separate CNN offers tremendous flexibility. We will demonstrate how to discover proper architectures for the above optimization process at the end of this section. 


\subsubsection{Noise Removal Module (NRM)}

It has been recognized that intensive noise in underexposed images cannot be simply removed by pre-/post-processing with existing denoising methods. Therefore, we intend to introduce another optimization unrolling module (NRM) to suppress noises in real-world low-light scenarios. Similar to IEM, we define a regularized model: $\min_{\mathbf{x}} \frac{1}{2}\|\mathbf{x}-\mathbf{u}_{K}\|^2 + \psi(\mathbf{x})$, where $\psi$ denotes the prior regularization on $\mathbf{x}$. By adopting the same unrolling strategy used in IEM, we can update our desired image $\mathbf{x}$ by
\begin{equation}
\mathbf{x}_{n+1}={\mathbf{u}}_{K} -\partial_{\mathbf{x}}\psi(\mathbf{x}_n).\label{eq:update_x}
\end{equation}
Here we write $\mathcal{C}_{\mathtt{NRM}}({\mathbf{x}}_{n})$ as the parameterization (i.e., CNN architecture) of $\partial_{\mathbf{x}}\psi(\mathbf{x}_n)$ and denote the output of NRM (with $N$ stages) as $\mathbf{x}_{N}$ in parallel.


\subsection{Cooperative Architecture Search}

In this part, we present a new search strategy to cooperatively discover architectures for both IEM and NRM.

\subsubsection{Compact Search Space for Low-light Priors}

We start with defining the search space for low-light prior modules ($\mathcal{C}_{\mathtt{IEM}}$ and $\mathcal{C}_{\mathtt{NRM}}$). By employing feature distillation techniques~\cite{liu2020residual}, we define our search space as a distillation cell, which is a directed acyclic graph with five nodes and each node connects to the next and the last nodes (see Fig.~\ref{fig:flowchart} (d)). In fact, each node in the cell is a latent representation and each direct edge is associated with some operation. The connection to the last node just realizes the feature information distillation. The candidate operations include 
1$\times$1 and 3$\times$3 Convolution (1-C and 3-C), 1$\times$1 and 3$\times$3 Residual Convolution (1-RC and 3-RC), 3$\times$3 Dilation Convolution with dilation rate of 2 (3-2-DC), 3$\times$3 Residual Dilation Convolution with dilation rate of 2 (3-2-RDC), and Skip Connection (SC). By adopting the continuous relaxation technique used in differentiable NAS literature~\cite{liu2018darts,xu2019pc,liang2019darts+}, we introduce the vectorized form $\bm{\alpha}=\{\bm{\alpha}_{\ttt},\bm{\alpha}_{\tn}\}$ to encode architectures in our search space (denoted as $\mathcal{A}$) for $\mathcal{C}_{\mathtt{IEM}}$ and $\mathcal{C}_{\mathtt{NRM}}$, respectively.
Denote by $\bomega=\{\bomega_{\ttt},\bomega_{\tn}\}$ the weight parameters associated with the architecture $\bm{\alpha}$.
Then the search task reduces to jointly learn $\bm{\alpha}$ and $\bm{\omega}$ within all the mixed operations.

\subsubsection{Differentiable Search with Cooperation}

The above search space can make our entire framework differentiable to both layer weights $\bomega$
and hyper-parameters $\balpha$, so that the most straightforward idea is to apply gradient-based NAS approaches for our problem. However, these classical methods can only learn $\bomega$ and $\balpha$ in an end-to-end fashion, which completely ignore the important light enhancement factors (e.g., illuminations and noises).
Our main idea to address this issue is to discover architectures that can properly reveal low-light prior information for underexposed images in real-world noisy scenarios. This is achieved by searching architectures for IEM and NRM by cooperation. Specifically, we formulate the search process of these two modules as a cooperative game and aim to solve the following optimization model for ${\bm{\alpha}_{\ttt}}$ (IEM) and ${\bm{\alpha}_{\tn}}$ (NRM):
\begin{equation}
\min\limits_{{\bm{\alpha}_{\tn}}\in\mathcal{A}}\left\{\min\limits_{{\bm{\alpha}_{\ttt}}\in\mathcal{A}}
\mathcal{L}_{\mathtt{val}}(\bm{\alpha}_{\ttt},\bm{\alpha}_{\tn};\bomega_{\ttt}^*,\bomega_{\tn}^*)\right\}.\label{eq:loss_val}
\end{equation}
We denote $\mathcal{L}_{\mathtt{val}}$ as a cooperative loss on the validation dataset, i.e., 
\begin{equation}
\mathcal{L}_{\mathtt{val}}:=\mathcal{L}^{\ttt}_{\mathtt{val}}(\balpha_{\ttt};\bomega_{\ttt}^{*})+\beta\mathcal{L}^{\tn}_{{\mathtt{val}}}(\balpha_{\tn}(\balpha_{\ttt});\bomega_{\tn}^{*}),
\end{equation} 
where $\mathcal{L}^{\ttt}_{\mathtt{val}}$ and $\mathcal{L}^{\tn}_{{\mathtt{val}}}$ respectively denote the losses on IEM and NRM and $\beta\geq 0$ is a trade-off parameter. Since NRM is based on the output of IRM (see Fig.~\ref{fig:flowchart}), here we should also consider $\balpha_{\ttt}$ as parameters of $\balpha_{\tn}$ in $\mathcal{L}^{\tn}_{{\mathtt{val}}}$. 
In fact, by analogy with the generative adversarial learning task~\cite{goodfellow2014generative}, it should be understood that the optimization problem in Eq.~\eqref{eq:loss_val} actually considers a cooperative (``min-min''), rather than an adversarial (``min-max'') objective.

\begin{algorithm}[t]
	\caption{Cooperative Architecture Search Strategy}\label{alg:search}
	\begin{algorithmic}[1]
		\REQUIRE 
		The search space $\mathcal{A}$, the training and validation datasets $\mathcal{D}_{\mathtt{tr}}$ and $\mathcal{D}_{\mathtt{val}}$ and necessary parameters.
		\ENSURE The searched architecture of RUAS. 
		\STATE Initialize $\balpha=\{\balpha_{\ttt},\balpha_{\tn}\}$ and $\bomega=\{\bomega_{\ttt},\bomega_{\tn}\}$.
		\WHILE{not converged}
		\STATE // Update $\balpha_{\ttt}$ and $\bomega_{\ttt}$ for IEM.
		\WHILE{not converged}
		\STATE  $\balpha_{\ttt}^{\dagger}\leftarrow\balpha_{\ttt}-\nabla_{\balpha_{\ttt}}\mathcal{L}_{\mathtt{val}}^{\ttt}(\balpha_{\ttt};\bomega_{\ttt}-
		\nabla_{\bomega_{\ttt}}\mathcal{L}_{\mathtt{tr}}^{\ttt})-\beta\nabla_{\balpha_{\ttt}}\mathcal{L}_{\mathtt{val}}^{\tn}(\balpha_{\tn}(\balpha_{\ttt});\bomega_{\tn})$.
		\STATE $\bomega_{\ttt}^{\dagger}\leftarrow\bomega_{\ttt}-\nabla_{\bomega_{\ttt}}\mathcal{L}_{\mathtt{tr}}^{\ttt}(\bomega_{\ttt};\balpha_{\ttt}^{\dagger})$.
		\ENDWHILE
		\STATE // Update $\balpha_{\tn}$ and $\bomega_{\tn}$ for NRM.
		\WHILE{not converged}
		\STATE  $\balpha_{\tn}^{\dagger}\leftarrow\balpha_{\tn}-\nabla_{\balpha_{\tn}}\mathcal{L}_{\mathtt{val}}^{\tn}(\balpha_{\tn}(\balpha_{\ttt}^{\dagger});\bomega_{\tn}-
		\nabla_{\bomega_{\tn}}\mathcal{L}_{\mathtt{tr}}^{\tn})$.
		\STATE $\bomega_{\tn}^{\dagger}\leftarrow\bomega_{\tn}-\nabla_{\bomega_{\tn}}\mathcal{L}_{\mathtt{tr}}^{\tn}(\bomega_{\tn};\balpha_{\tn}^{\dagger})$.
		\ENDWHILE
		\ENDWHILE
		\RETURN  Architecture derived based on $\balpha_{\ttt}^*$ and $\balpha_{\tn}^*$.
	\end{algorithmic}
\end{algorithm}

\begin{table*}[t]
	\caption{Quantitative results (PSNR and SSIM) of state-of-the-art methods and ours on the MIT-Adobe 5K and LOL datasets. The best result is in red whereas the second best one is in blue.}
	\vspace{-0.2cm}
	\begin{center}
		\begin{tabular}{|c|c|c@{\extracolsep{0.4em}}c@{\extracolsep{0.4em}}c@{\extracolsep{0.4em}}c@{\extracolsep{0.4em}}c@{\extracolsep{0.4em}}c@{\extracolsep{0.4em}}c@{\extracolsep{0.4em}}c@{\extracolsep{0.4em}}c@{\extracolsep{0.4em}}c@{\extracolsep{0.4em}}c@{\extracolsep{0.4em}}c@{\extracolsep{0.4em}}c|}
			\hline 
			\footnotesize Datasets&\footnotesize Metrics&\footnotesize LIME&\footnotesize SDD&\footnotesize MBLLEN&\footnotesize GALDNet&\footnotesize RetinexNet&\footnotesize EnGAN&\footnotesize SSIENet&\footnotesize KinD&\footnotesize DeepUPE&\footnotesize ZeroDCE&\footnotesize FIDE&\footnotesize DRBN&\footnotesize Ours\\
			\hline
			\multirow{2}{*}{\footnotesize MIT}&\footnotesize PSNR&\footnotesize 17.788&\footnotesize 17.617& \footnotesize 15.587&\footnotesize 16.728&	\footnotesize 12.685&\footnotesize 15.014&\footnotesize 10.324&\footnotesize 17.169&\footnotesize \textcolor{blue}{\textbf{18.779}}&\footnotesize 16.463&\footnotesize 17.170&\footnotesize 15.954&\footnotesize \textcolor{red}{\textbf{20.830}}\\
			\cline{2-15}
			~&\footnotesize SSIM&\footnotesize\textcolor{blue}{\textbf{0.826}}&\footnotesize 0.792&\footnotesize 0.713&\footnotesize 0.764&\footnotesize 0.644&\footnotesize 0.768&\footnotesize 0.620&\footnotesize 0.696&\footnotesize 0.822&\footnotesize 0.764&\footnotesize 0.696&\footnotesize 0.704&\footnotesize \textcolor{red}{\textbf{0.854}}\\
			\hline
			\multirow{2}{*}{\footnotesize LOL}&\footnotesize PSNR&\footnotesize 14.916&\footnotesize 15.484&\footnotesize 13.931&\footnotesize 16.188&\footnotesize 13.096&\footnotesize 15.644&\footnotesize 14.176&\footnotesize 14.616&\footnotesize 13.041&\footnotesize 15.512&\footnotesize \textcolor{blue}{\textbf{16.718}}&\footnotesize 15.324&\footnotesize \textcolor{red}{\textbf{18.226}}\\
			\cline{2-15}  
			~&\footnotesize SSIM&\footnotesize 0.516&\footnotesize 0.578&\footnotesize 0.489&\footnotesize 0.605&\footnotesize 0.429&\footnotesize 0.578&\footnotesize 0.534&\footnotesize 0.636&\footnotesize 0.483&\footnotesize 0.553&\footnotesize 0.673&\footnotesize \textcolor{blue}{\textbf{0.699}}&\footnotesize \textcolor{red}{\textbf{0.717}}\\
			\hline
		\end{tabular}
	\end{center}
	\label{tab: MITquantitative}
	\vspace{-0.2cm}
\end{table*}

\begin{figure*}[t]
	\begin{center}
		\begin{tabular}{c@{\extracolsep{0.3em}}c@{\extracolsep{0.3em}}c@{\extracolsep{0.3em}}c@{\extracolsep{0.3em}}c@{\extracolsep{0.3em}}c@{\extracolsep{0.3em}}c@{\extracolsep{0.3em}}c}
			\includegraphics[width=0.117\linewidth]{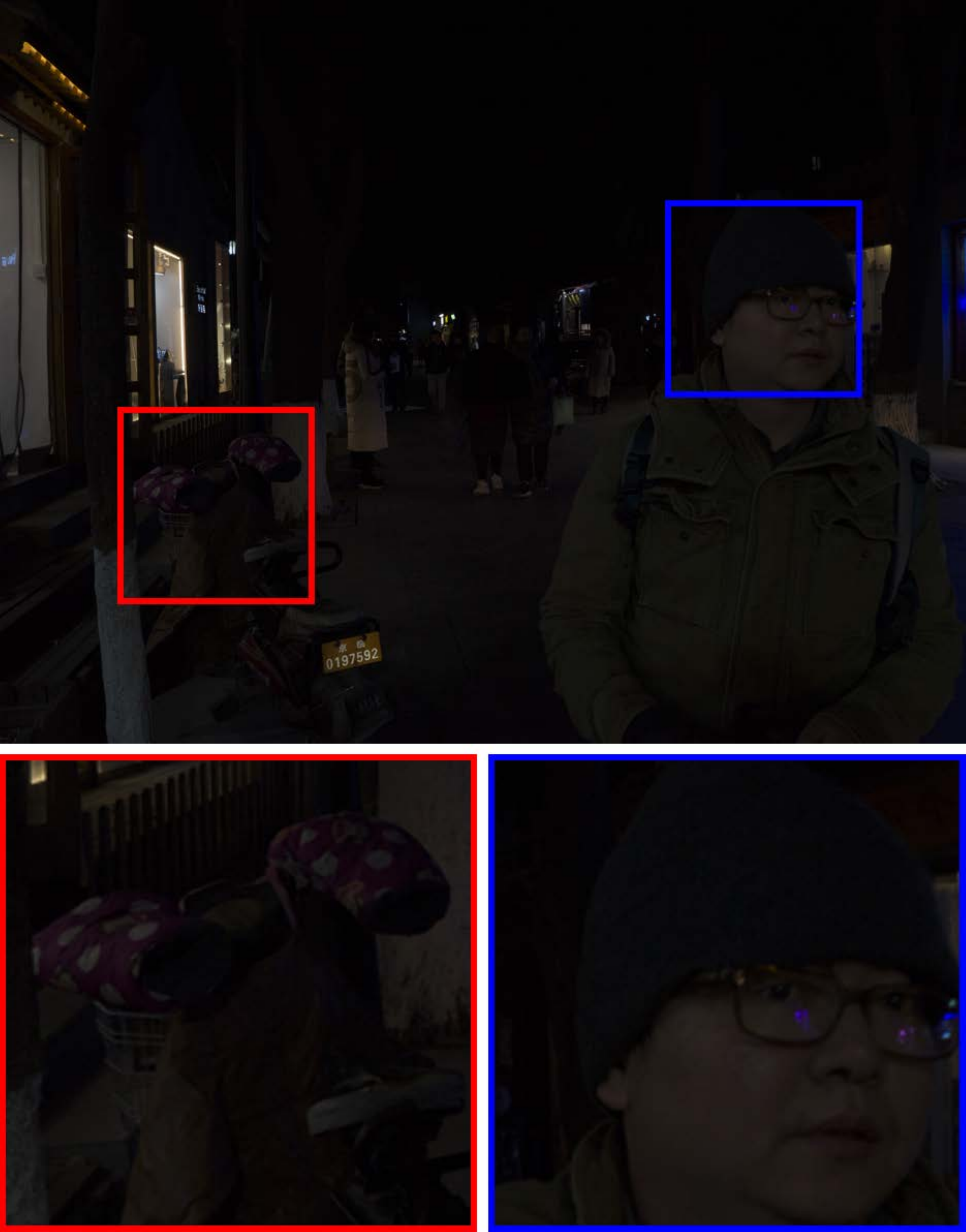}&
			\includegraphics[width=0.117\linewidth]{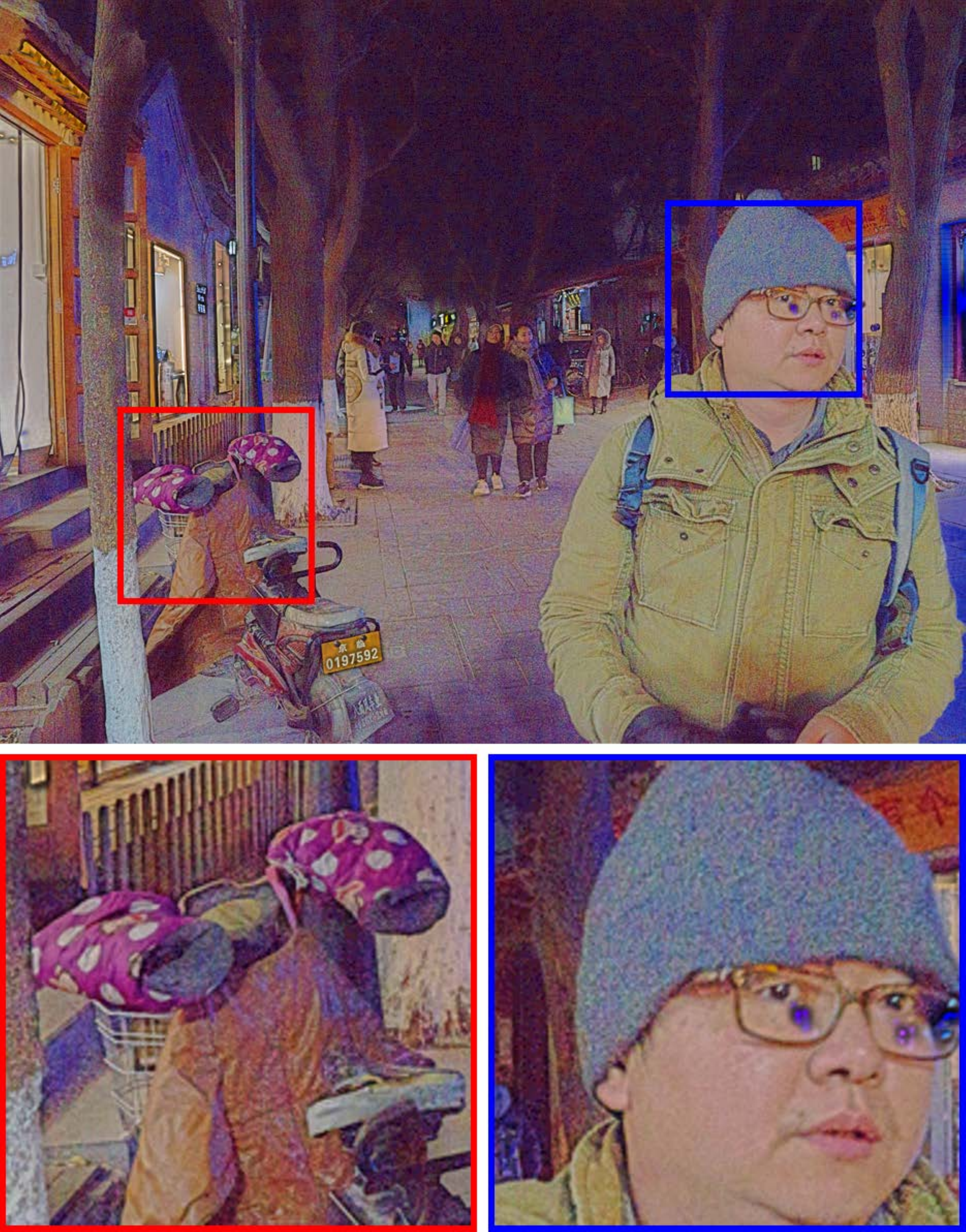}&
			\includegraphics[width=0.117\linewidth]{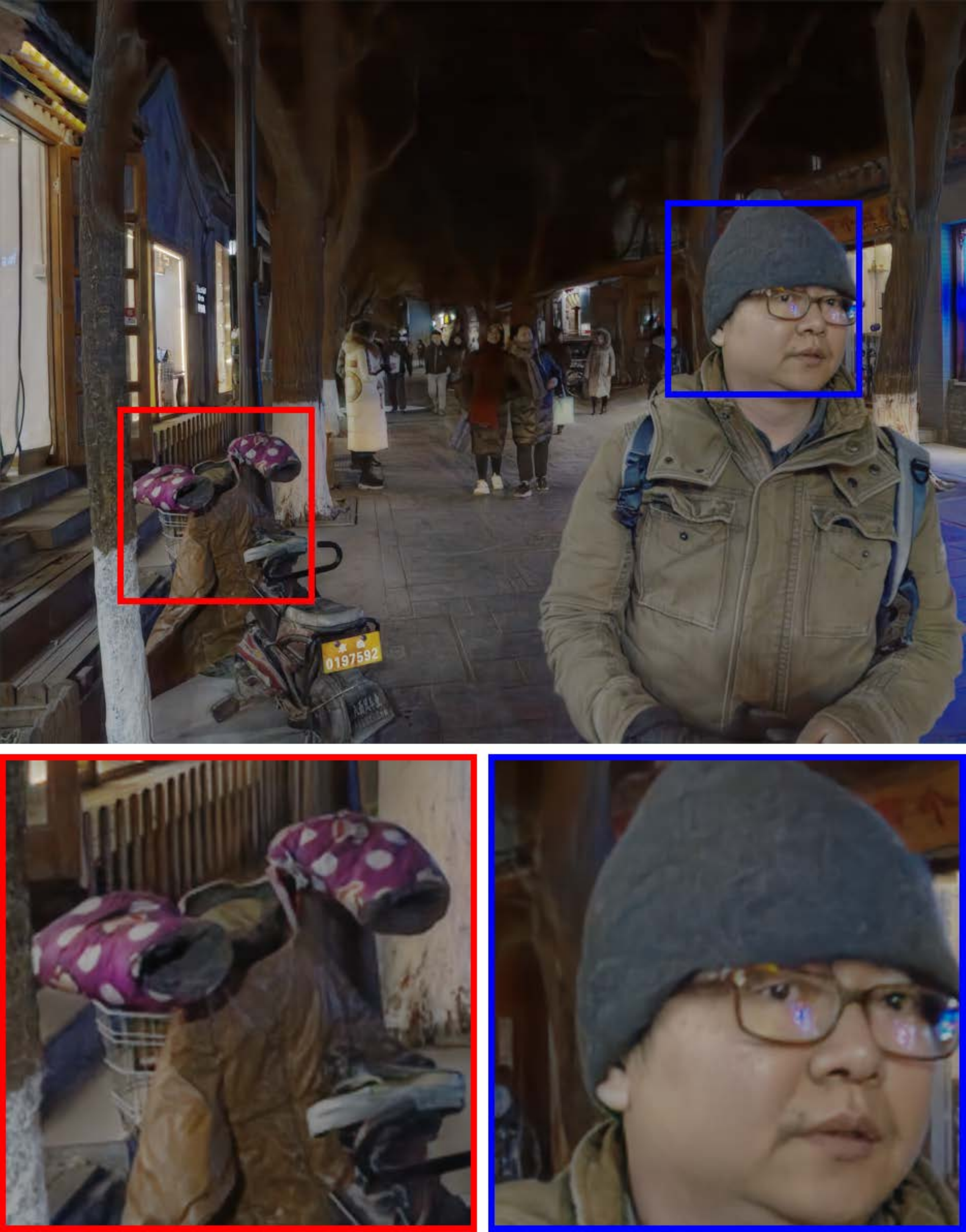}&
			\includegraphics[width=0.117\linewidth]{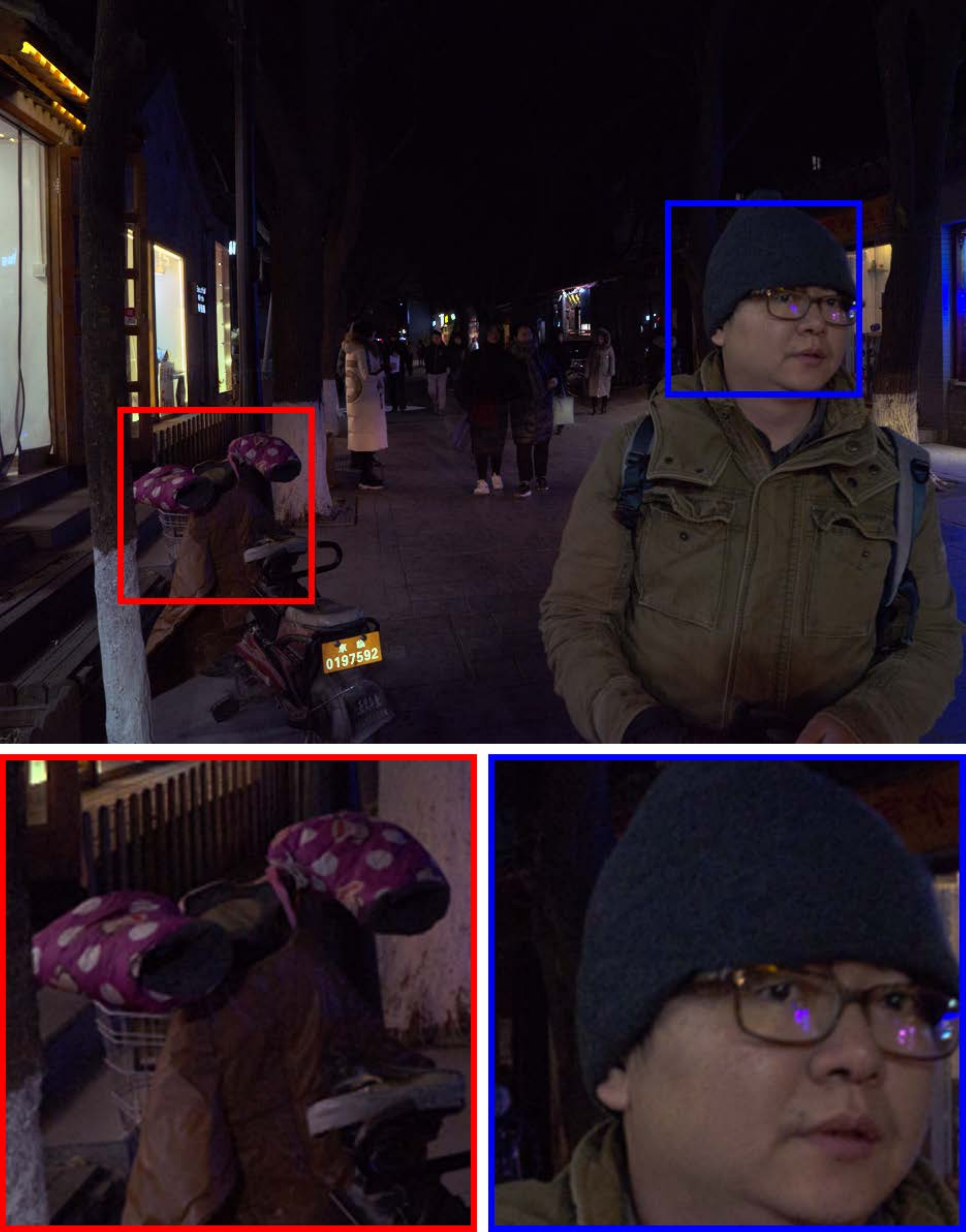}&
			\includegraphics[width=0.117\linewidth]{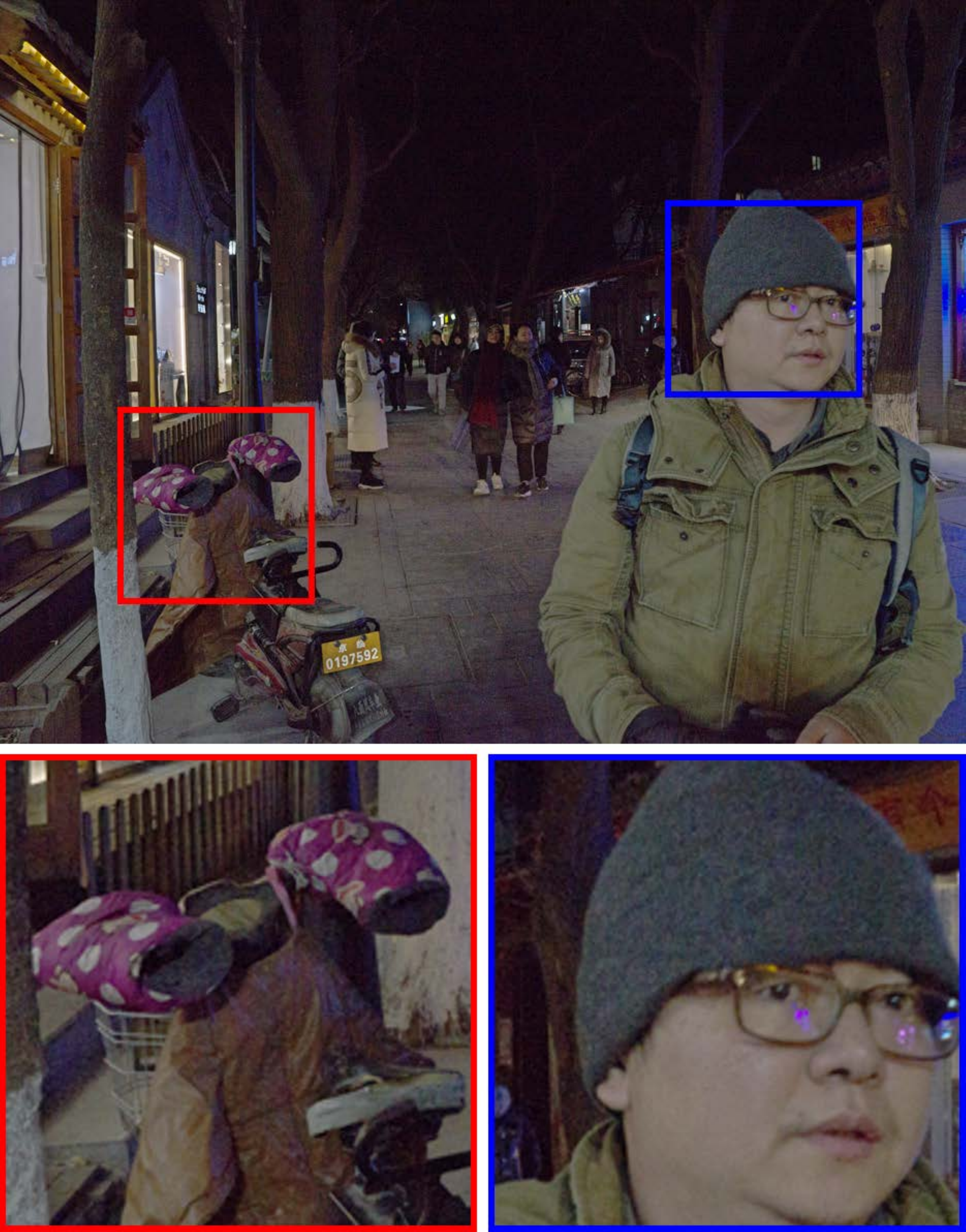}&
			\includegraphics[width=0.117\linewidth]{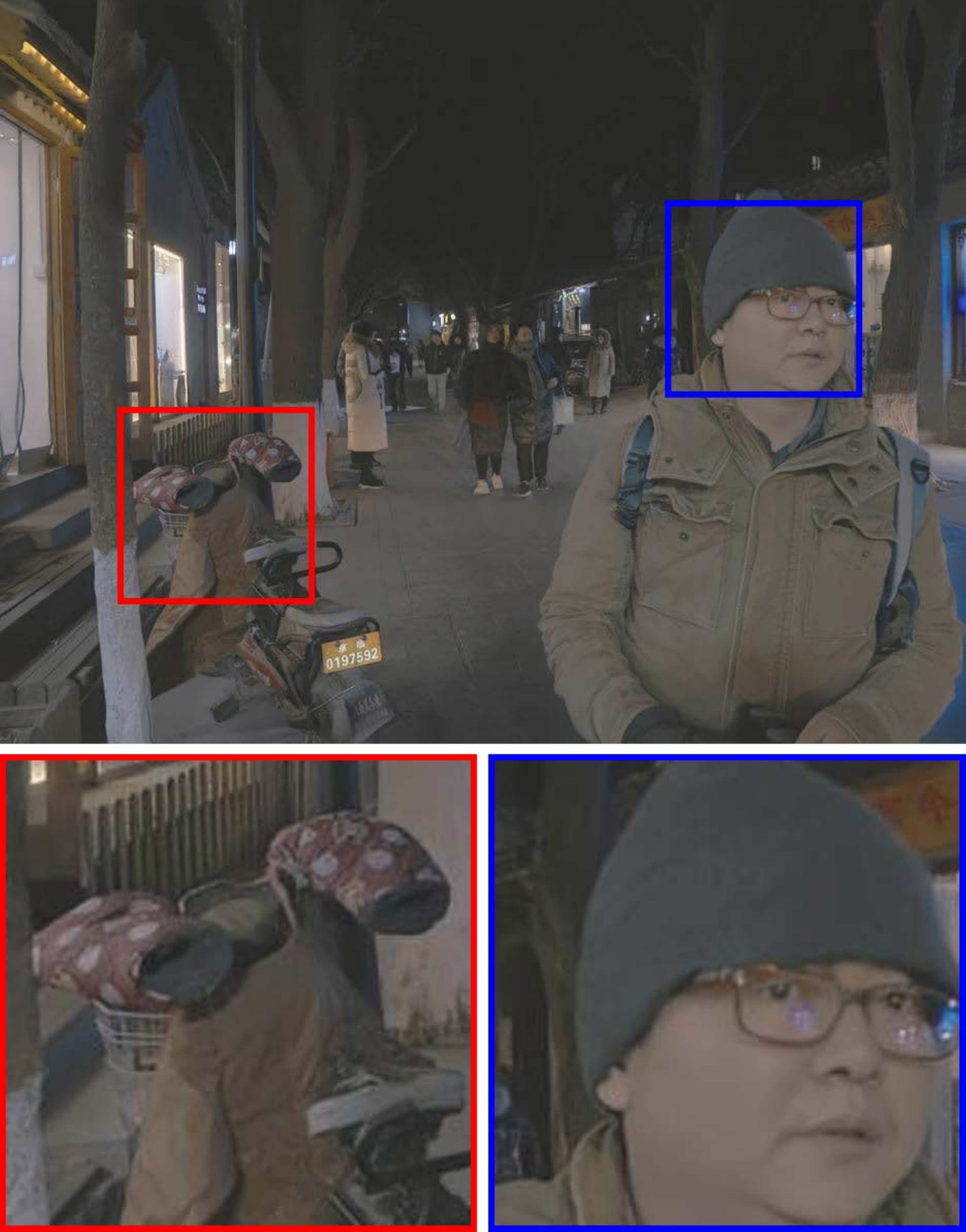}&
			\includegraphics[width=0.117\linewidth]{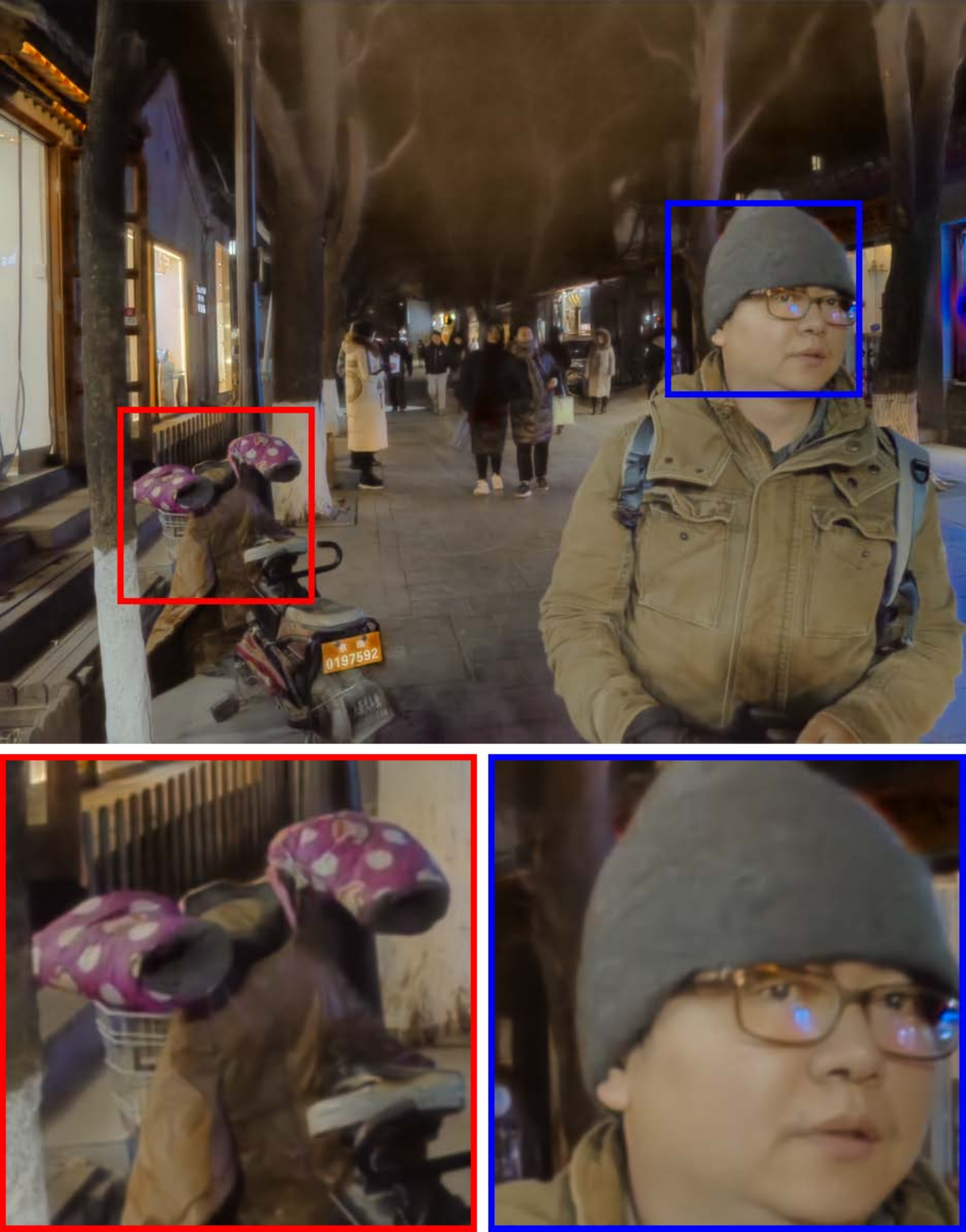}&
			\includegraphics[width=0.117\linewidth]{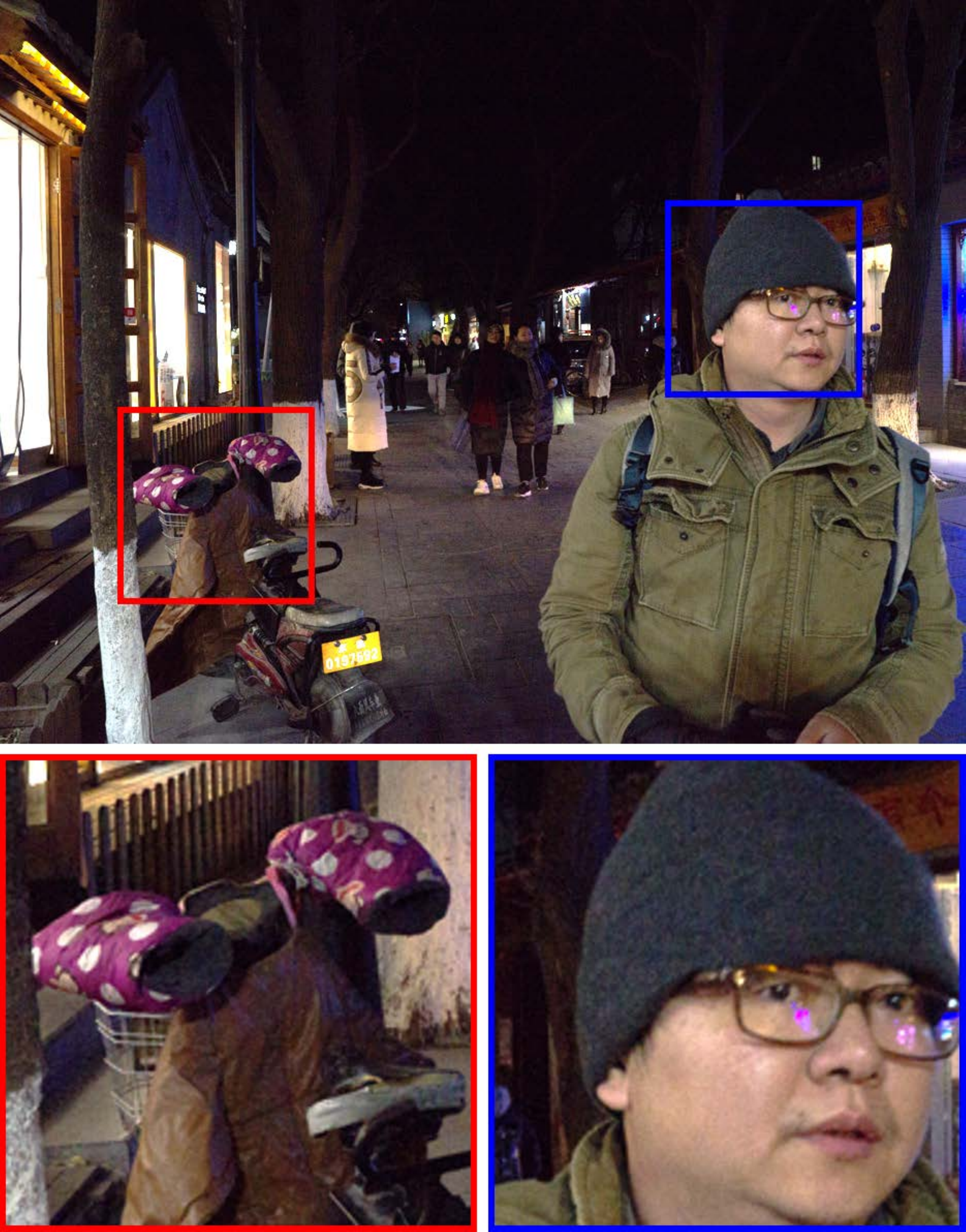}\\
			\includegraphics[width=0.117\linewidth]{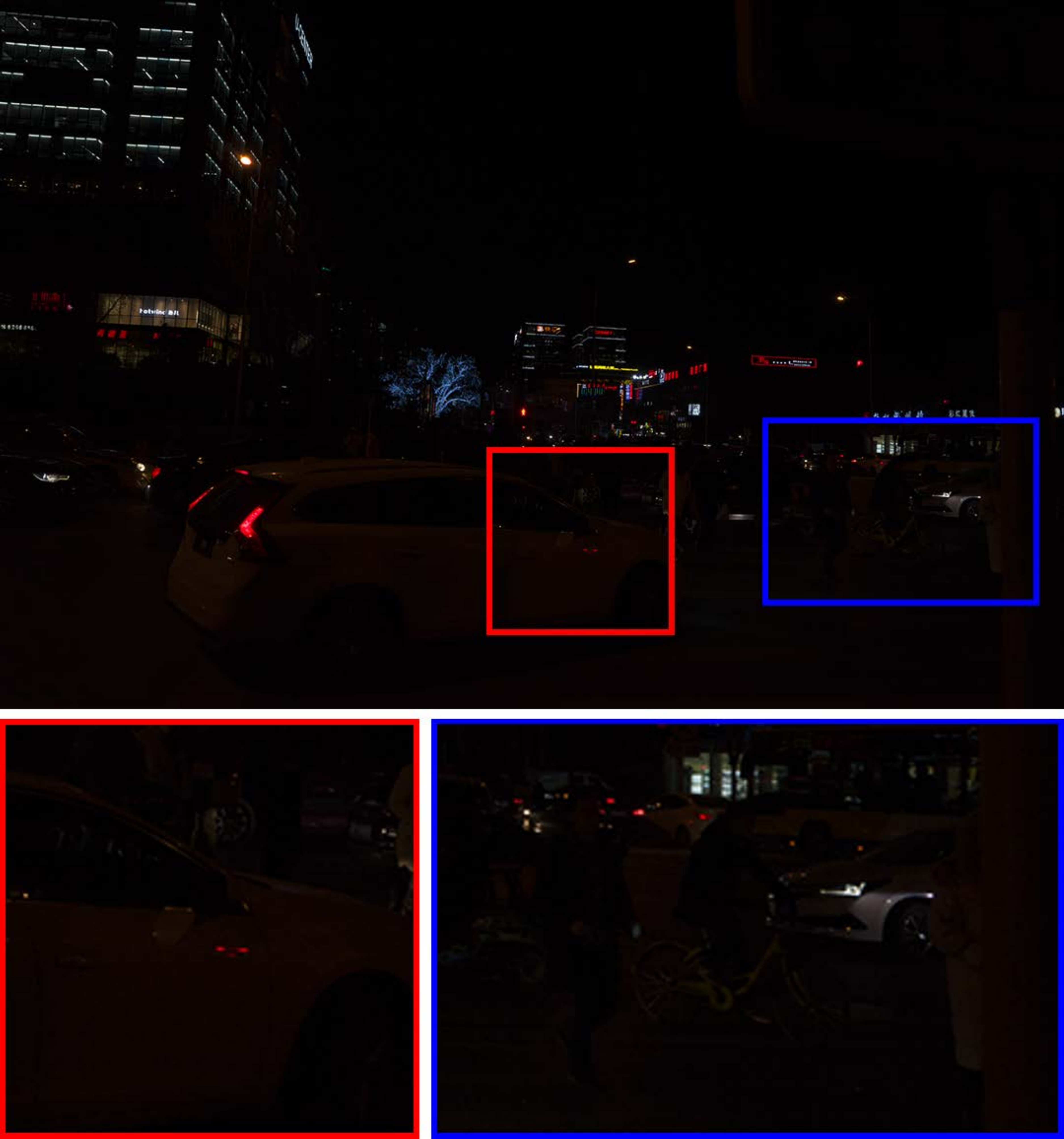}&
			\includegraphics[width=0.117\linewidth]{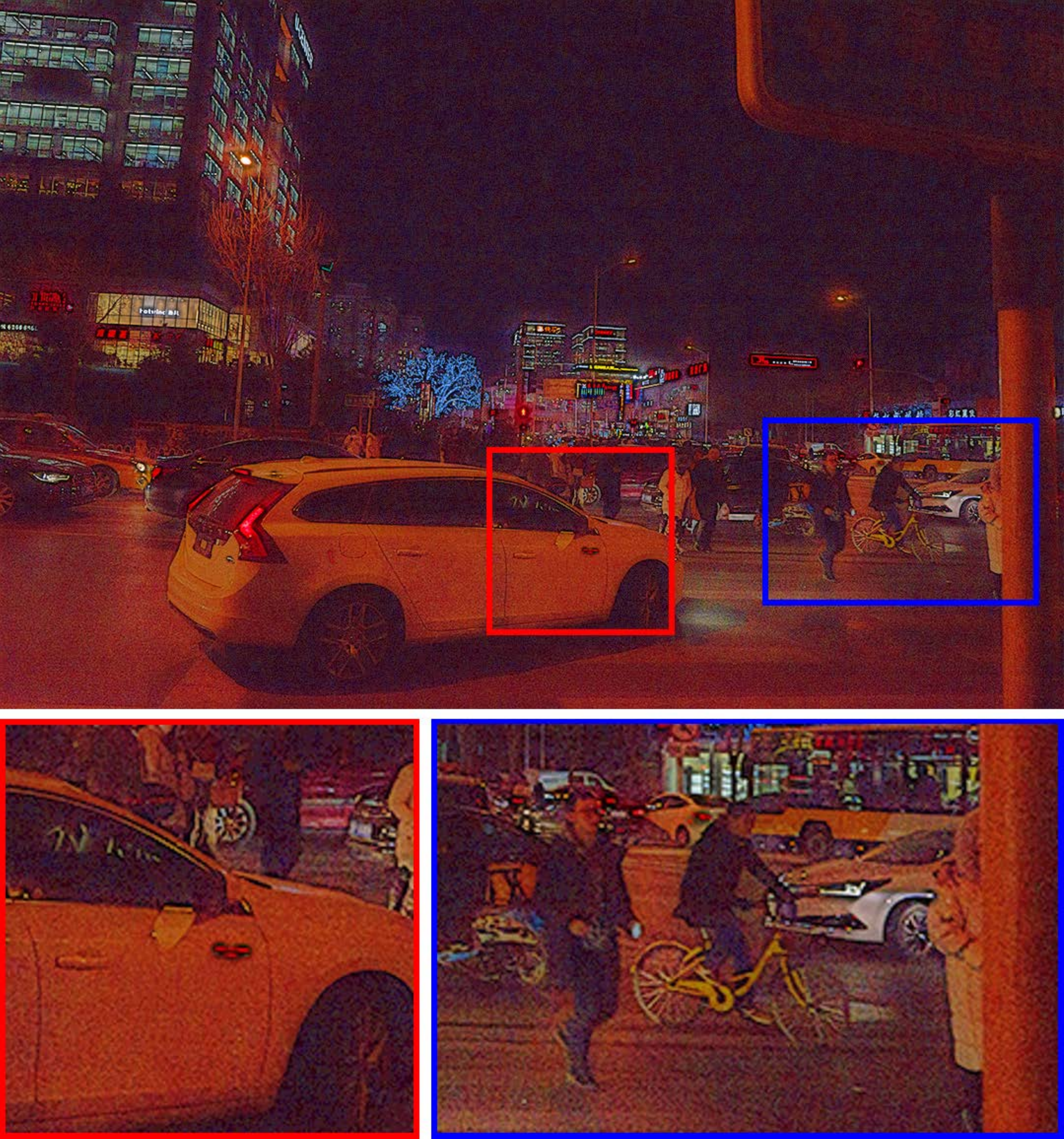}&
			\includegraphics[width=0.117\linewidth]{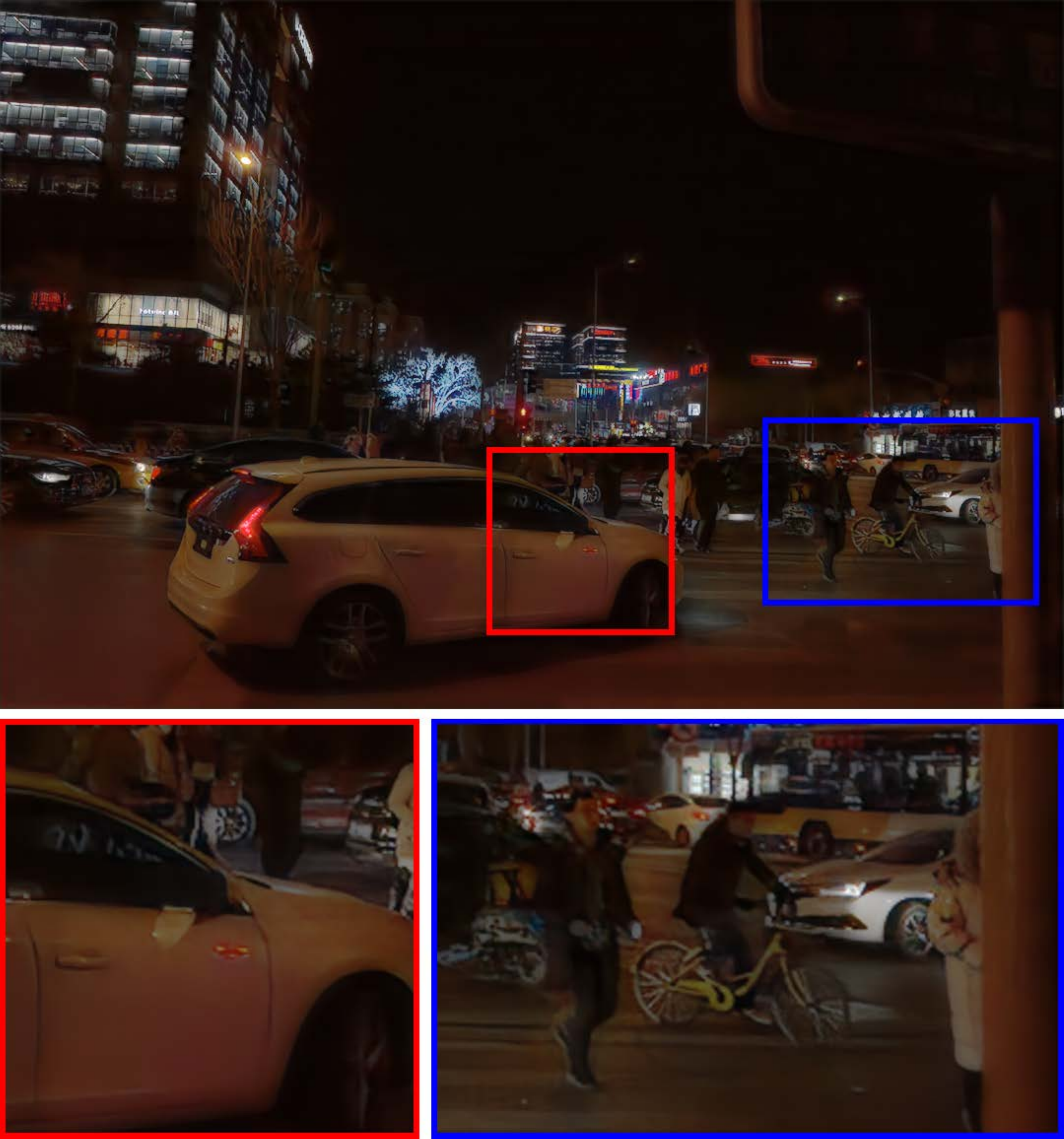}&
			\includegraphics[width=0.117\linewidth]{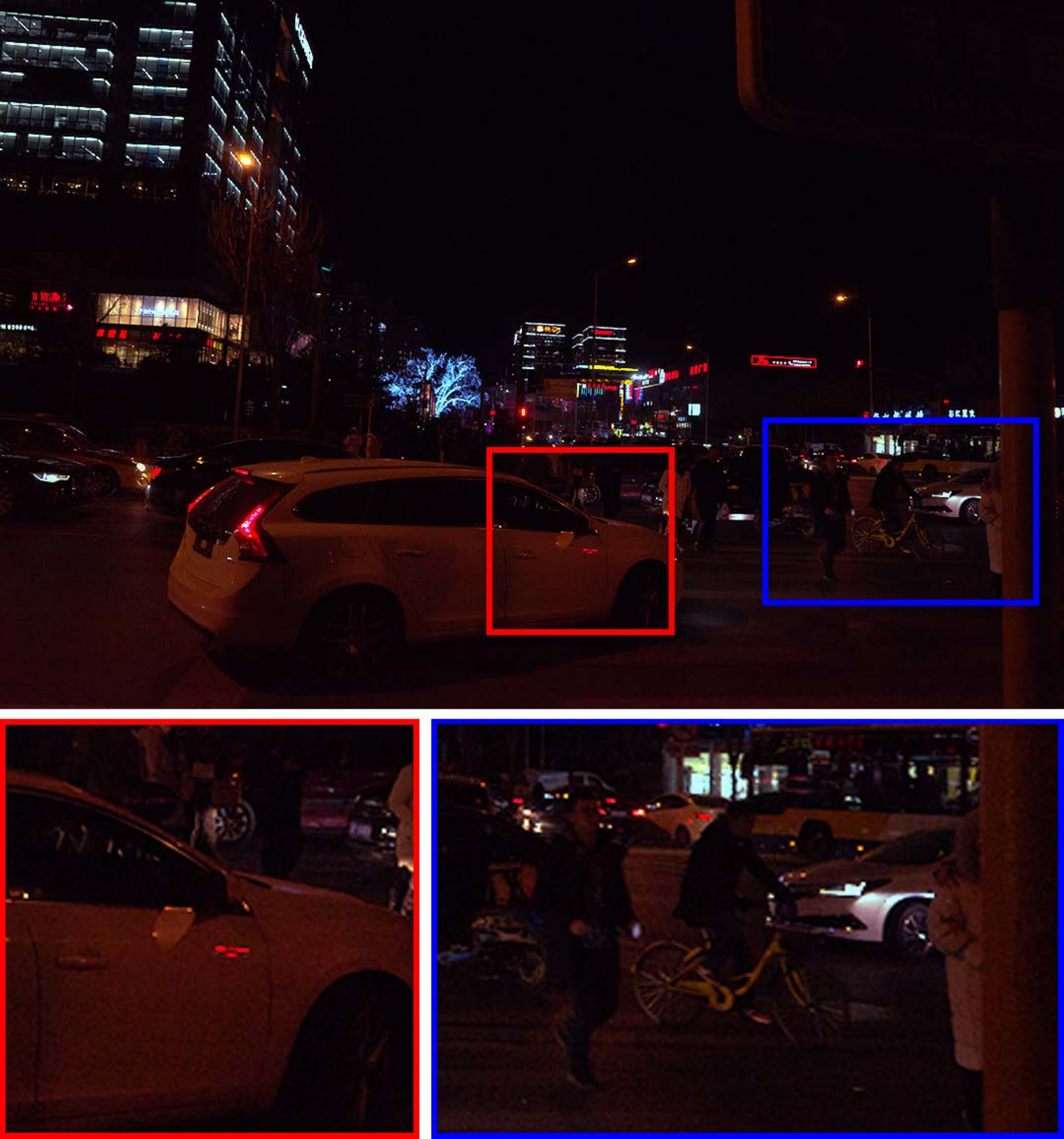}&
			\includegraphics[width=0.117\linewidth]{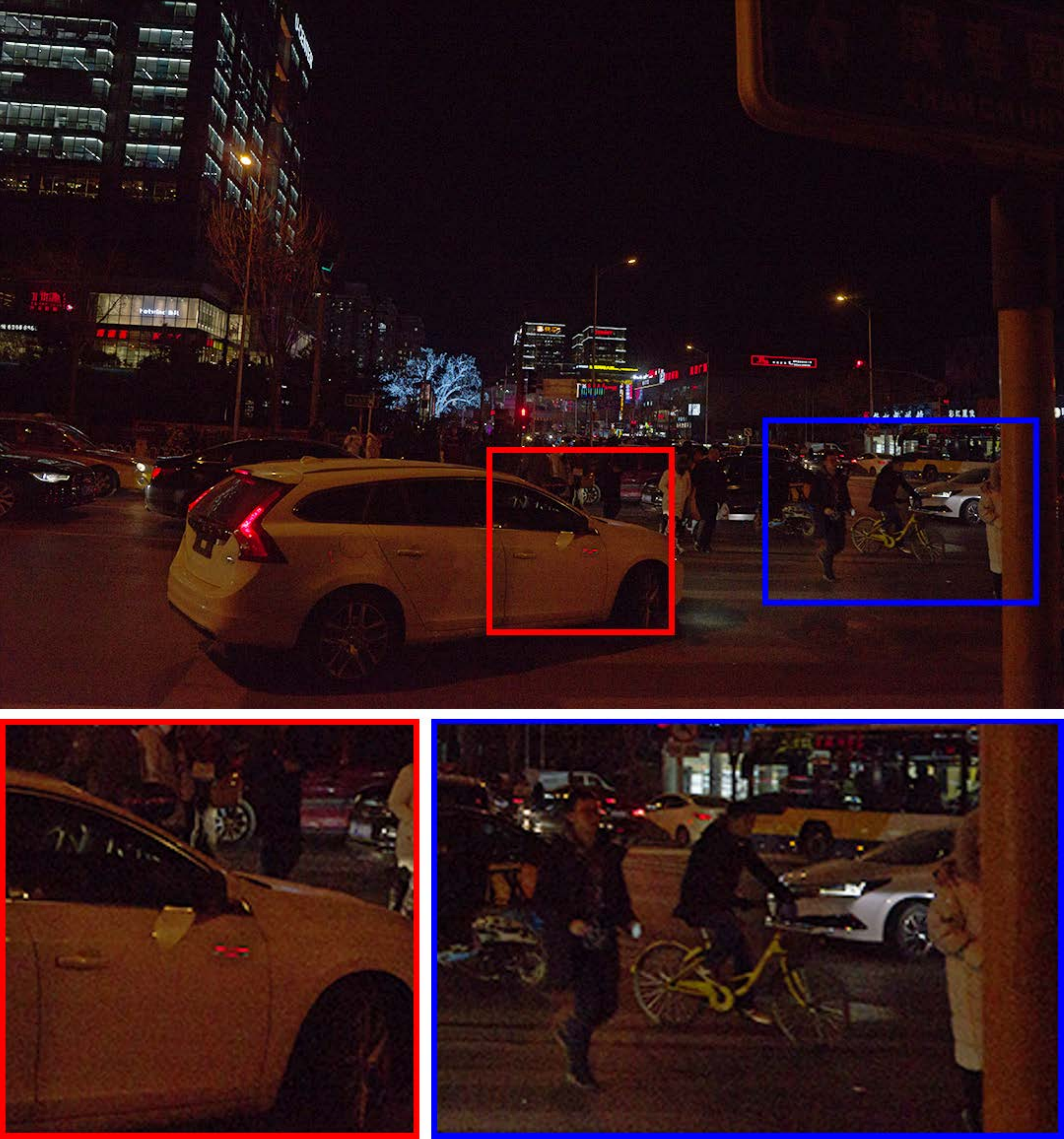}&
			\includegraphics[width=0.117\linewidth]{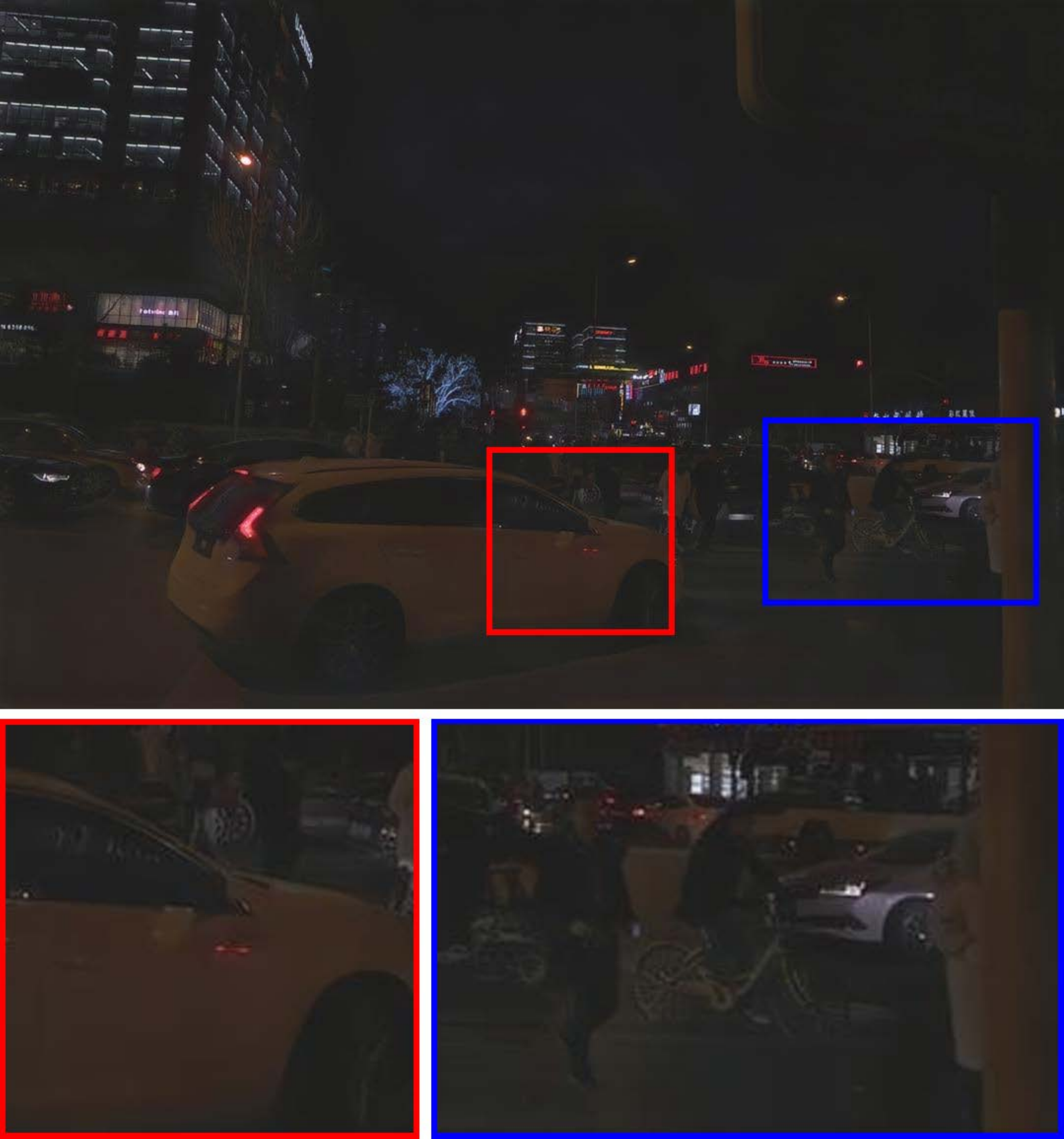}&
			\includegraphics[width=0.117\linewidth]{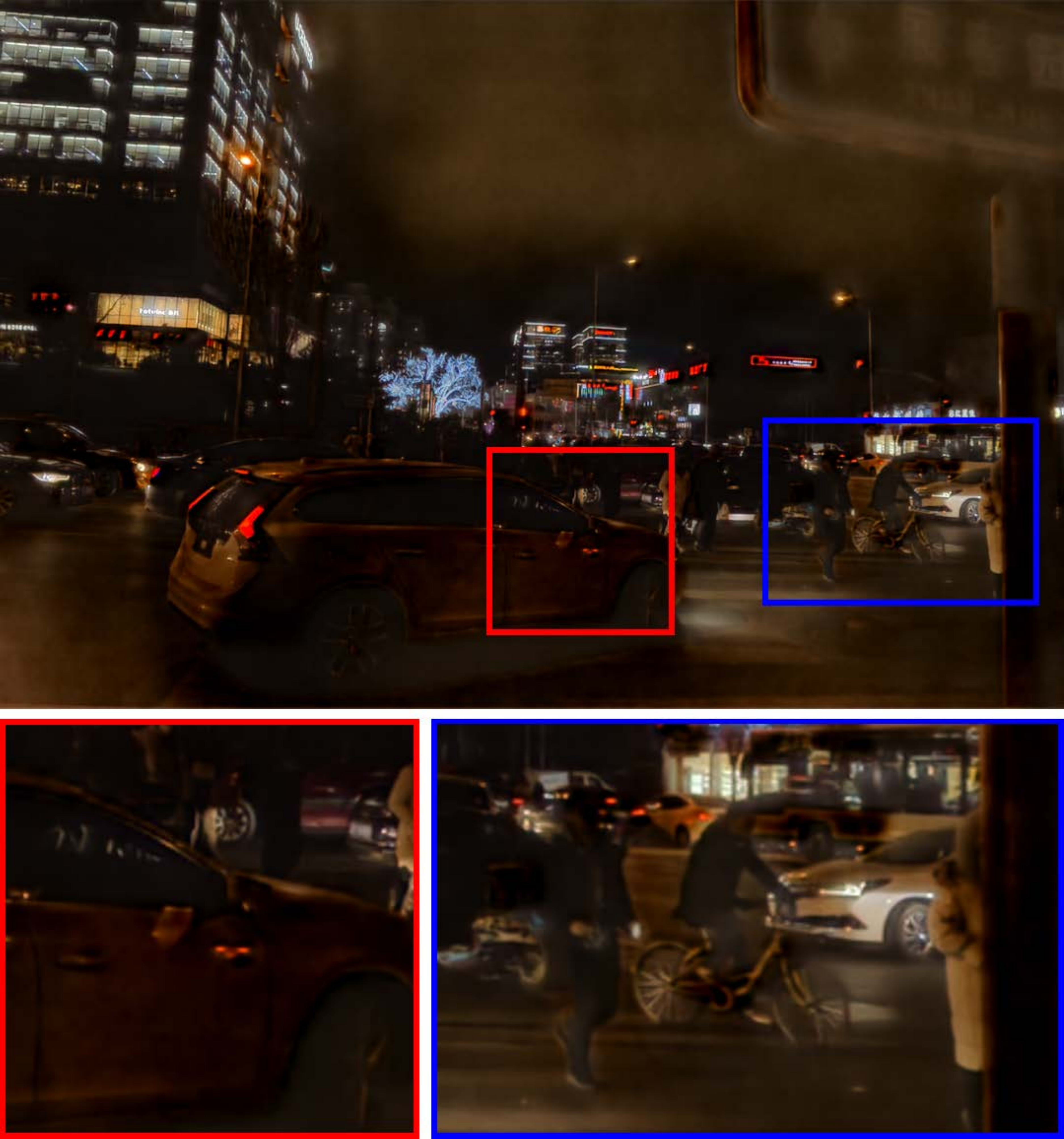}&
			\includegraphics[width=0.117\linewidth]{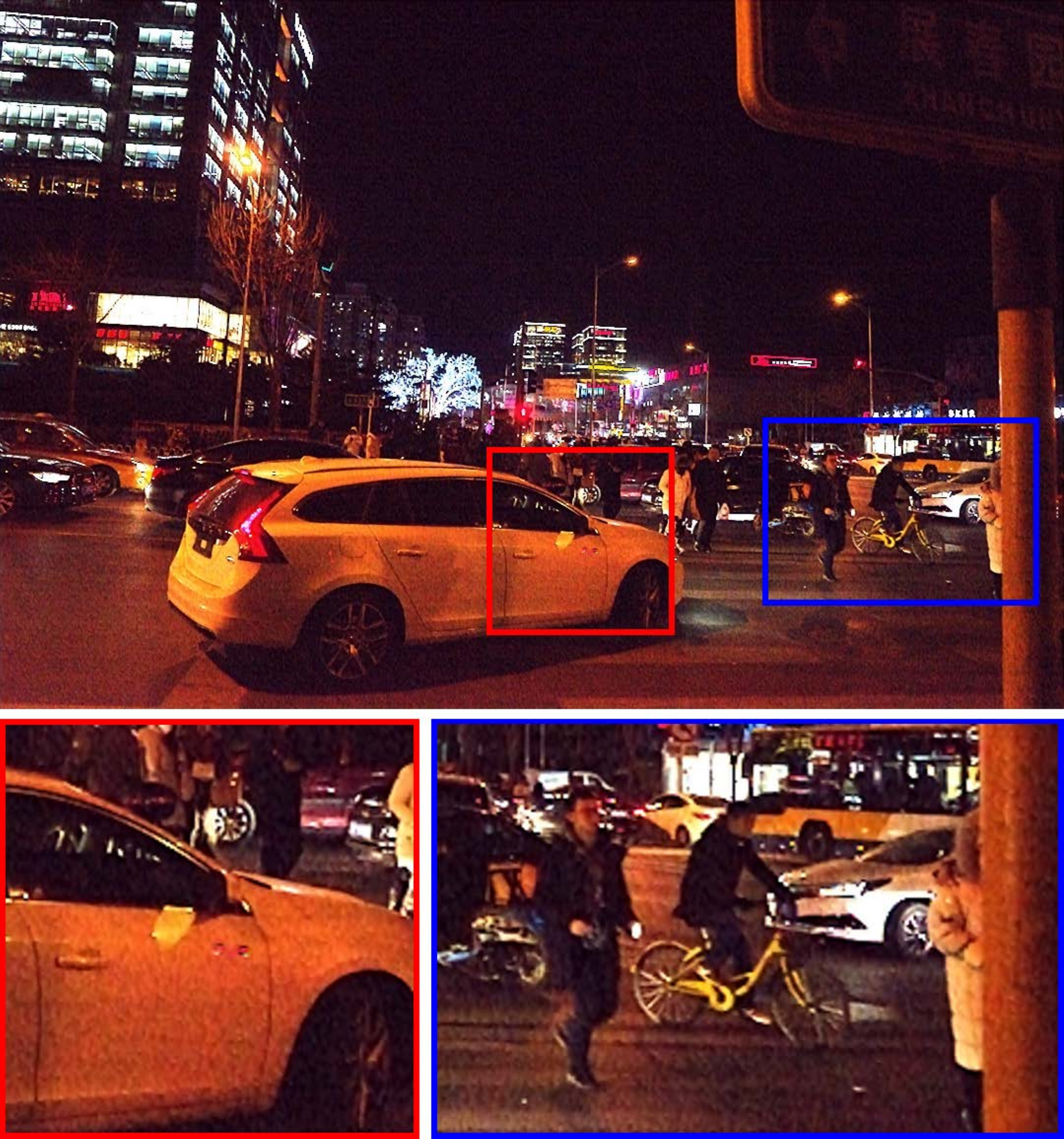}\\
			\includegraphics[width=0.117\linewidth]{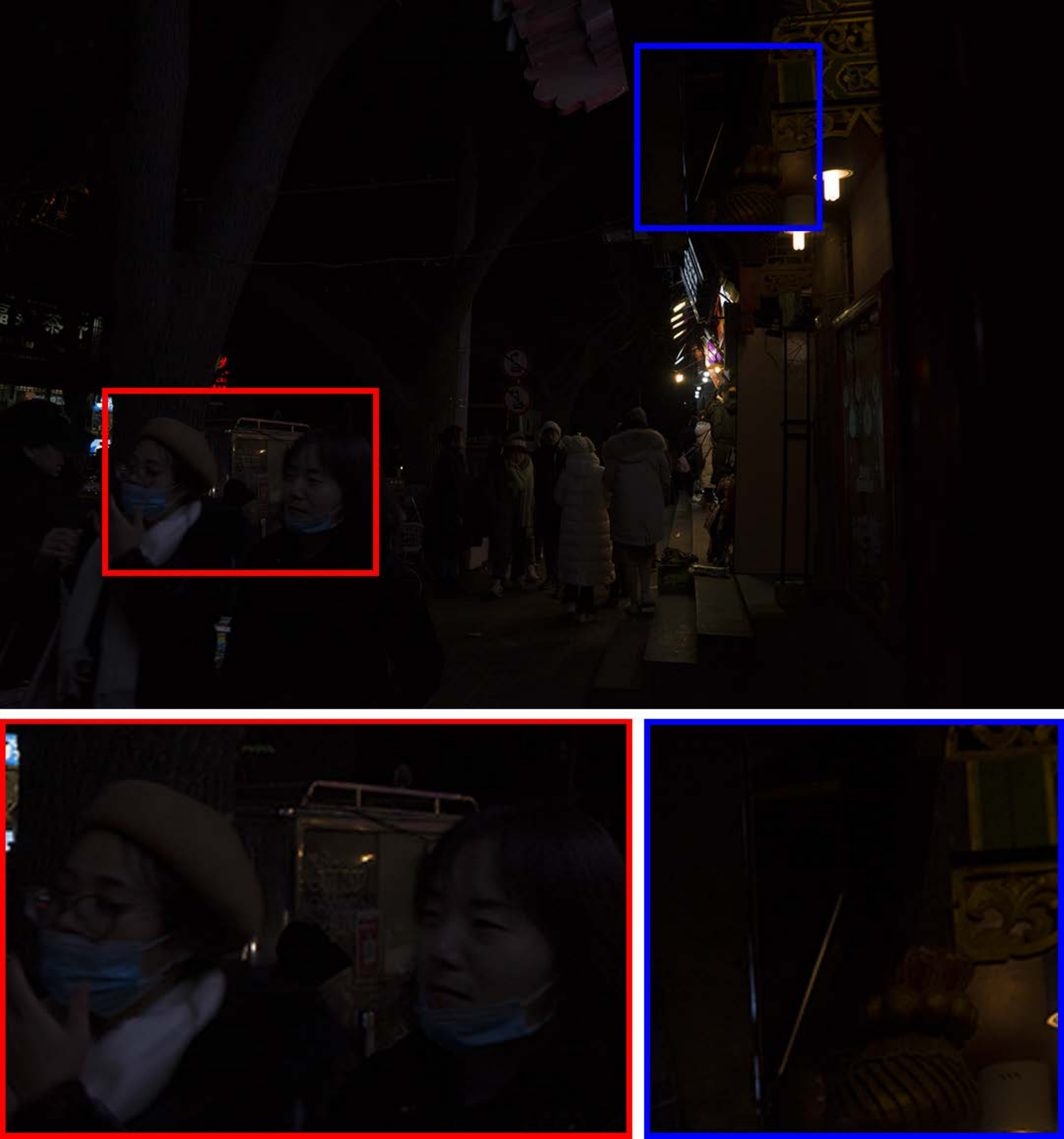}&
			\includegraphics[width=0.117\linewidth]{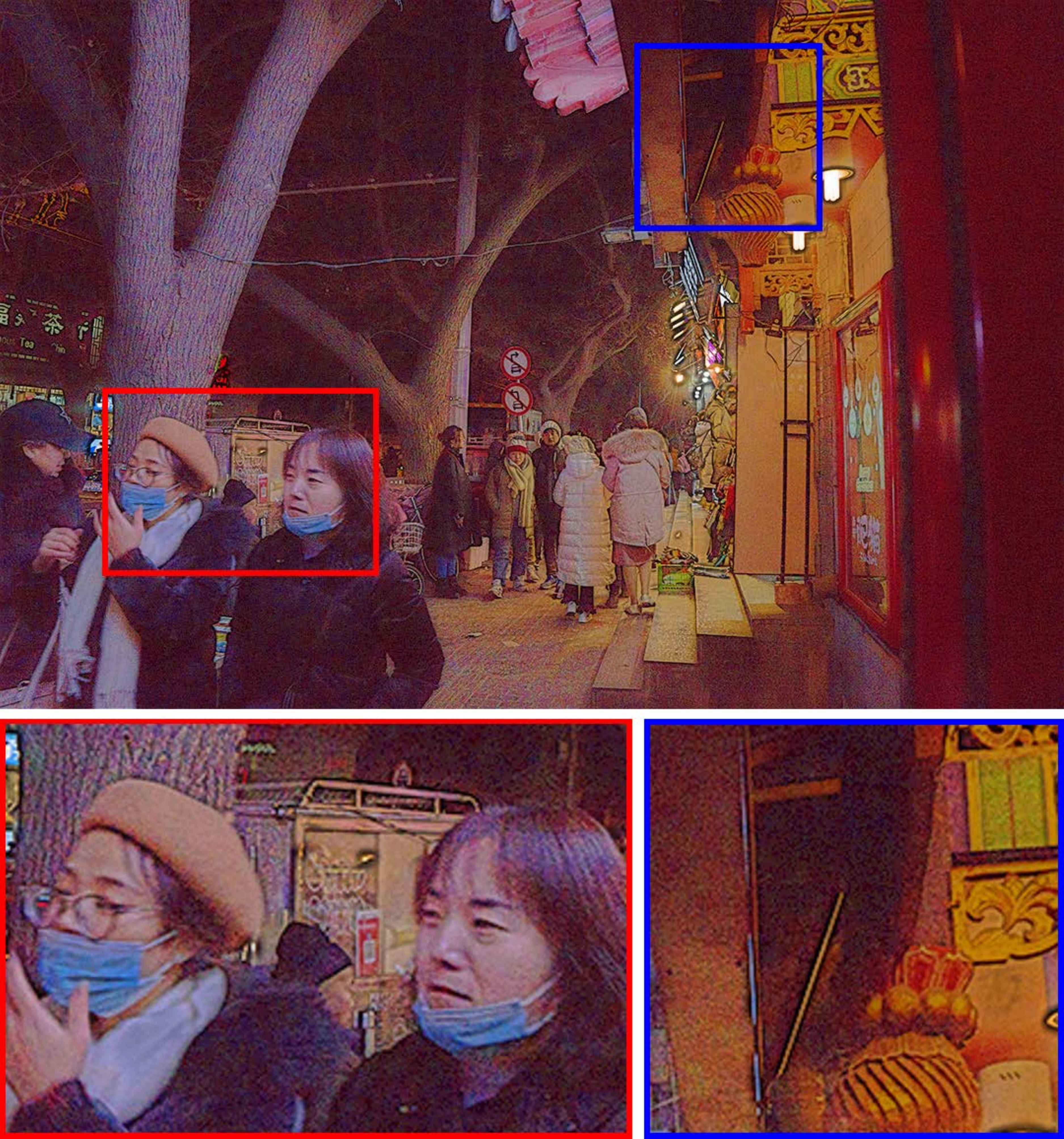}&
			\includegraphics[width=0.117\linewidth]{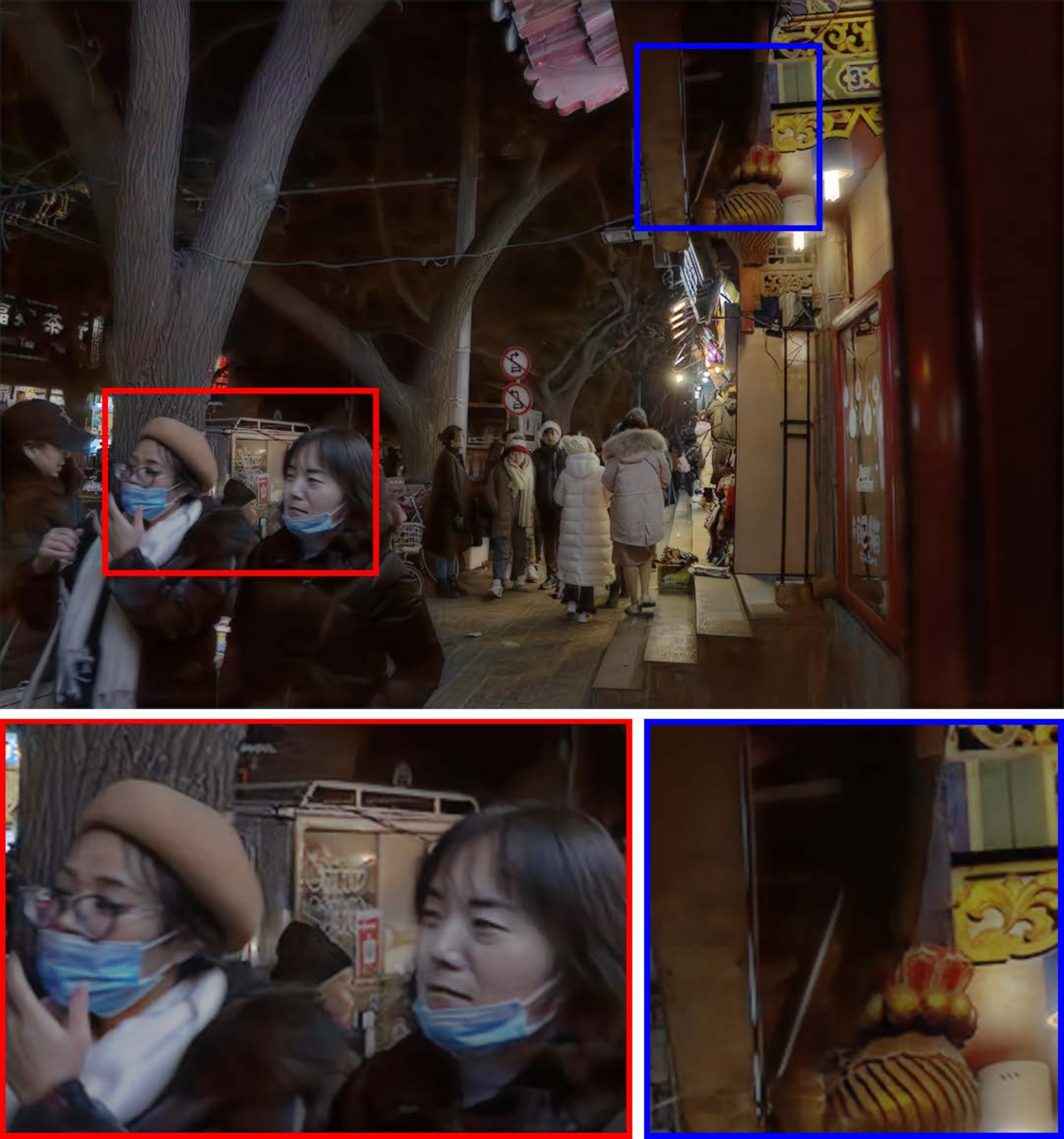}&
			\includegraphics[width=0.117\linewidth]{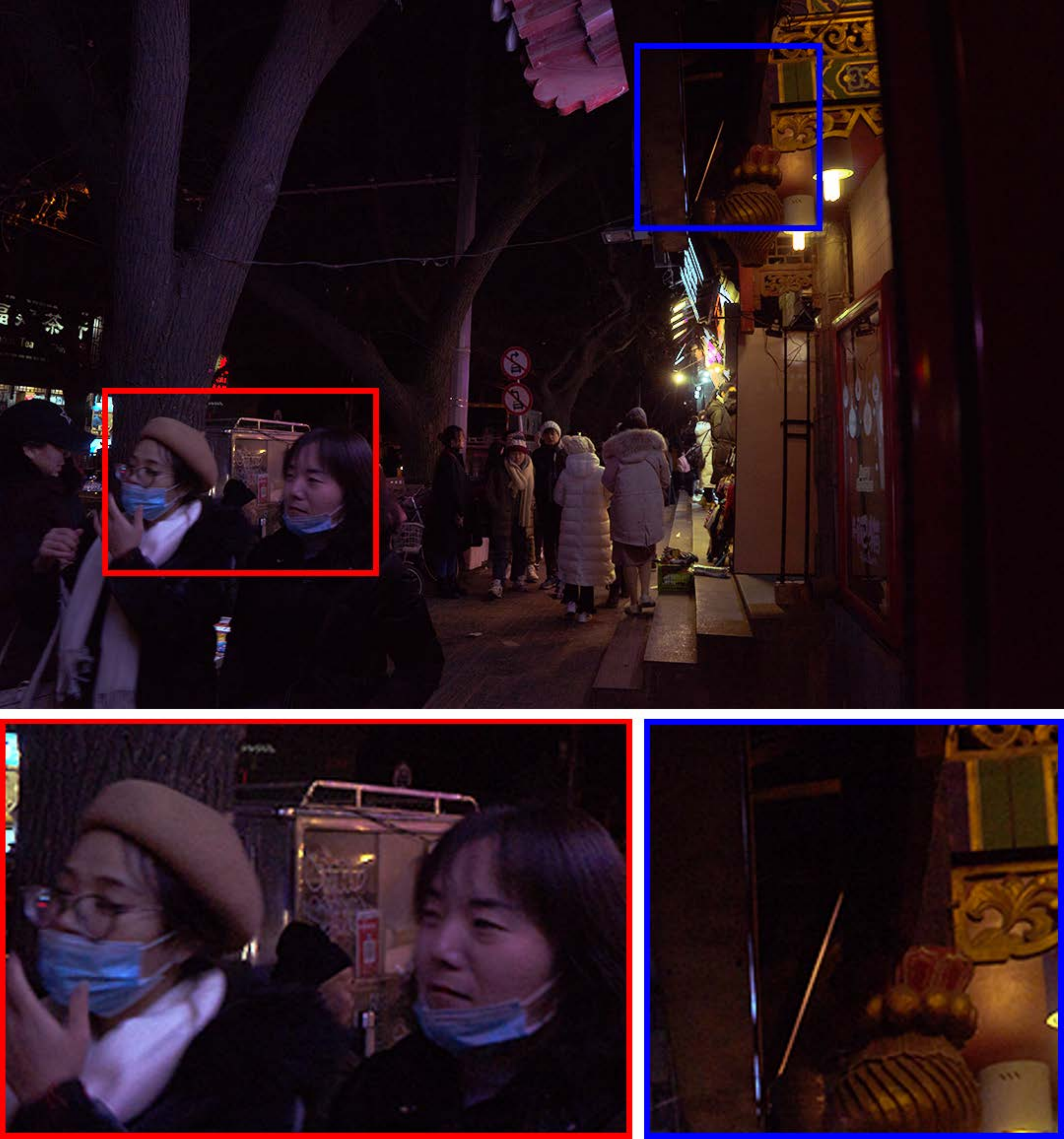}&
			\includegraphics[width=0.117\linewidth]{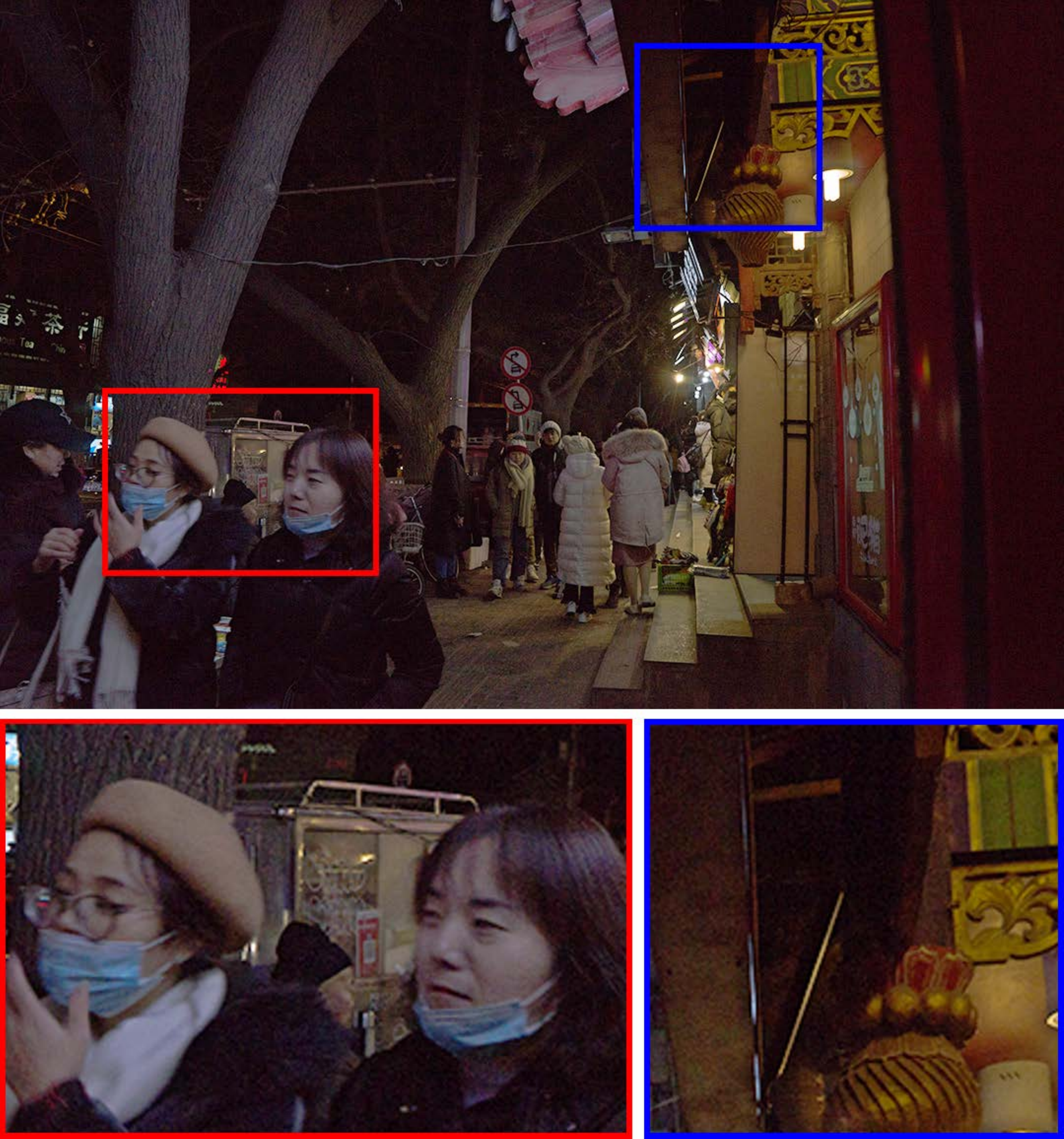}&
			\includegraphics[width=0.117\linewidth]{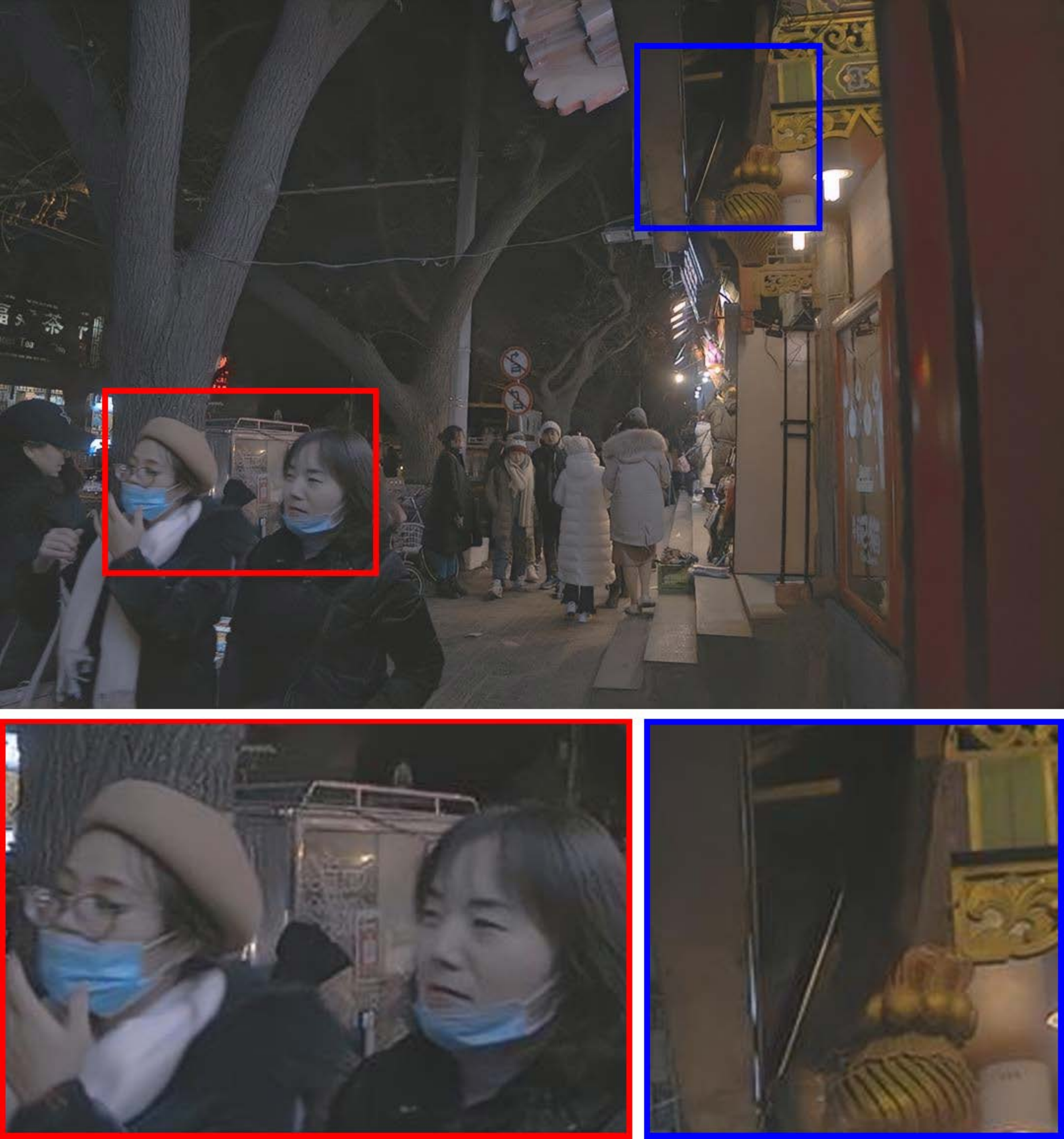}&
			\includegraphics[width=0.117\linewidth]{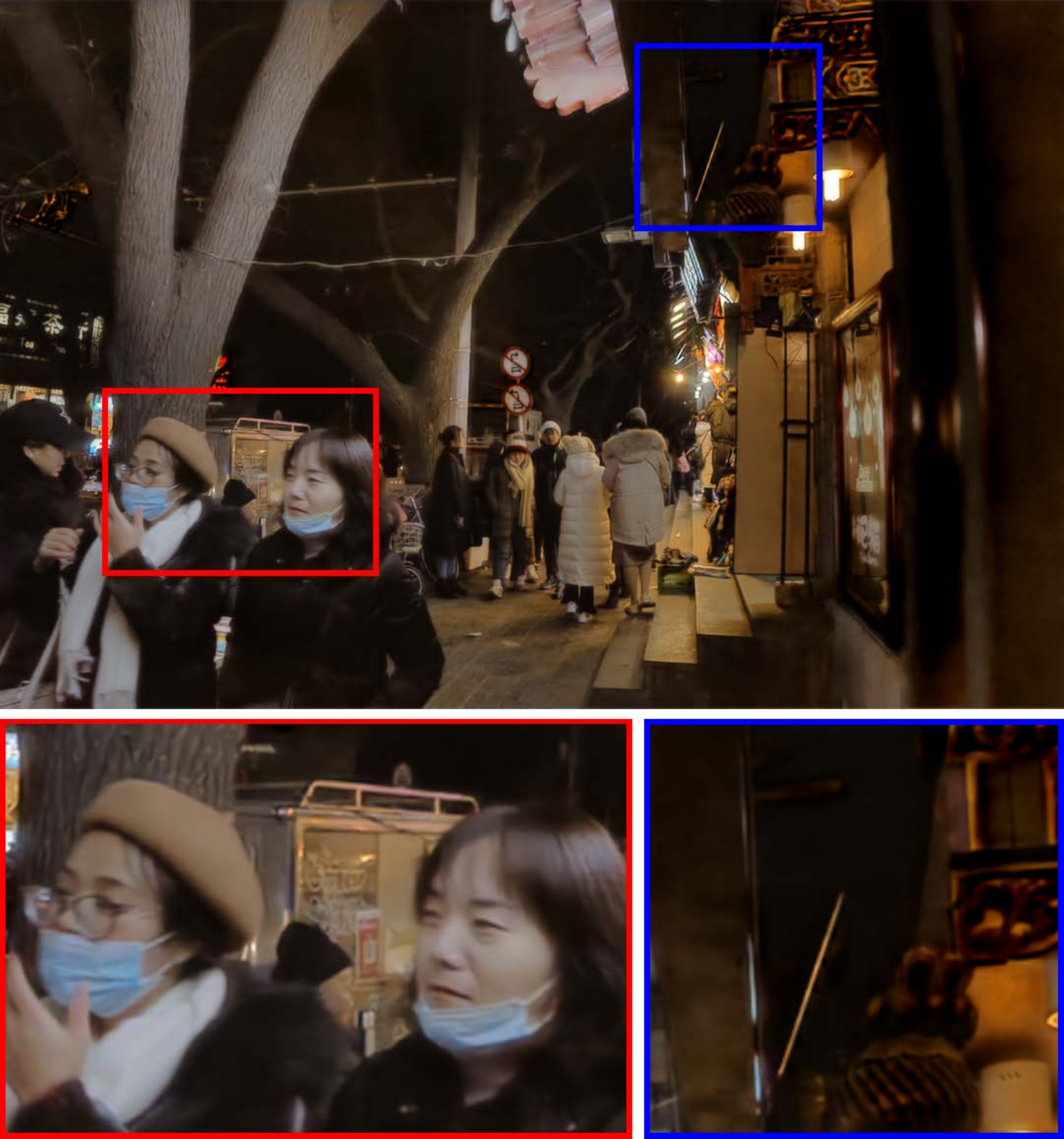}&
			\includegraphics[width=0.117\linewidth]{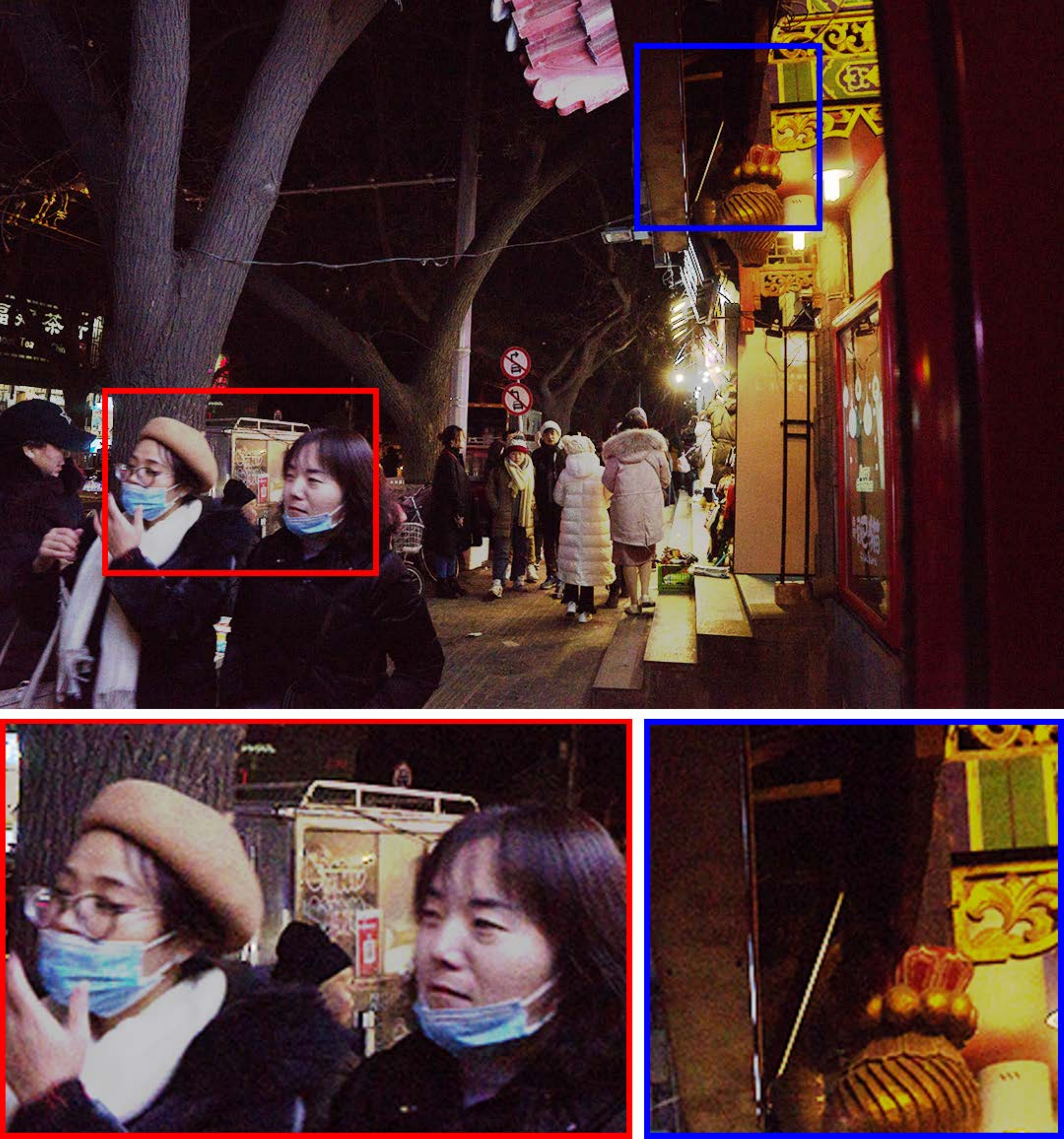}\\
			\footnotesize{Input}&\footnotesize RetinexNet&\footnotesize{KinD}&\footnotesize DeepUPE&\footnotesize{ZeroDCE}&\footnotesize FIDE&\footnotesize DRBN&\footnotesize Ours\\
		\end{tabular}
	\end{center}
\vspace{-0.25cm}
	\caption{Visual results of state-of-the-art methods and ours on the DarkFace dataset. Red boxes indicate the obvious differences. }
	\vspace{-0.25cm}
	\label{fig:DarkFace}
\end{figure*}

As for the weight parameters $\bomega_{\ttt}^*$ (and $\bomega_{\tn}^*$), we assume that they are only associated with the architecture  $\balpha_{\ttt}$ (and $\balpha_{\tn}$). That is, they can be obtained by minimizing the following models 
\begin{equation}
\left\{\begin{array}{l}
\bomega_{\ttt}^*=\arg\min\limits_{\bomega_{\ttt}}\mathcal{L}_{\mathtt{tr}}^{\ttt}(\bomega_{\ttt};\balpha_{\ttt}),\\
\bomega_{\tn}^*=\arg\min\limits_{\bomega_{\tn}}\mathcal{L}_{\mathtt{tr}}^{\tn}(\bomega_{\tn};\balpha_{\tn}),
\end{array}\right.\label{eq:loss_tr}
\end{equation}
where $\mathcal{L}_{\mathtt{tr}}^{\ttt}$ and $\mathcal{L}_{\mathtt{tr}}^{\tn}$ are the training losses for IEM and NRM, respectively. 

Therefore, our search strategy implies a bilevel optimization problem with Eq.~\eqref{eq:loss_val} and Eq.~\eqref{eq:loss_tr} as the upper-level and lower-level subproblems, respectively. Moreover, the upper-level subproblem in Eq.~\eqref{eq:loss_val} should be further separated into two cooperative tasks during the search process.

\begin{figure*}[t]
	\begin{center}
		\begin{tabular}{c@{\extracolsep{0.3em}}c@{\extracolsep{0.3em}}c@{\extracolsep{0.3em}}c@{\extracolsep{0.3em}}c}
			\includegraphics[width=0.188\linewidth]{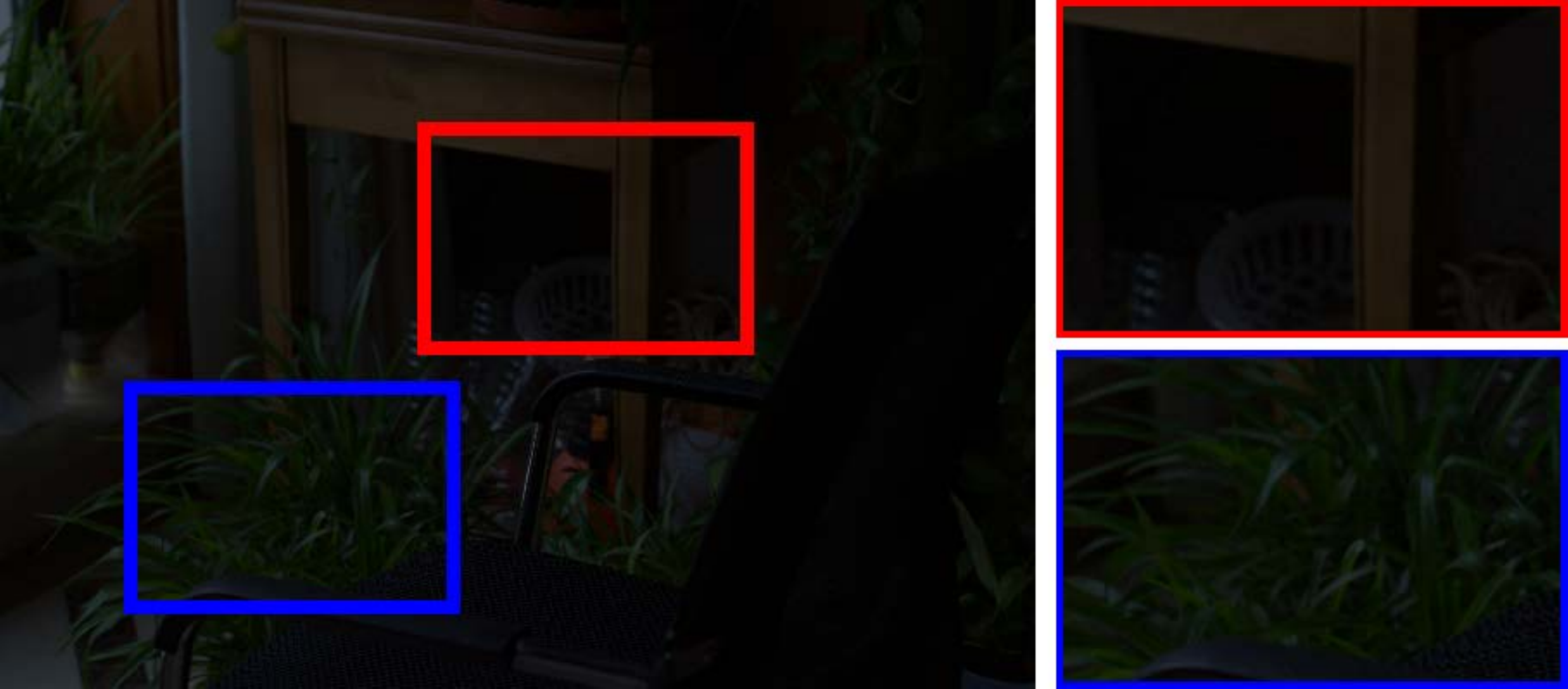}&
			\includegraphics[width=0.188\linewidth]{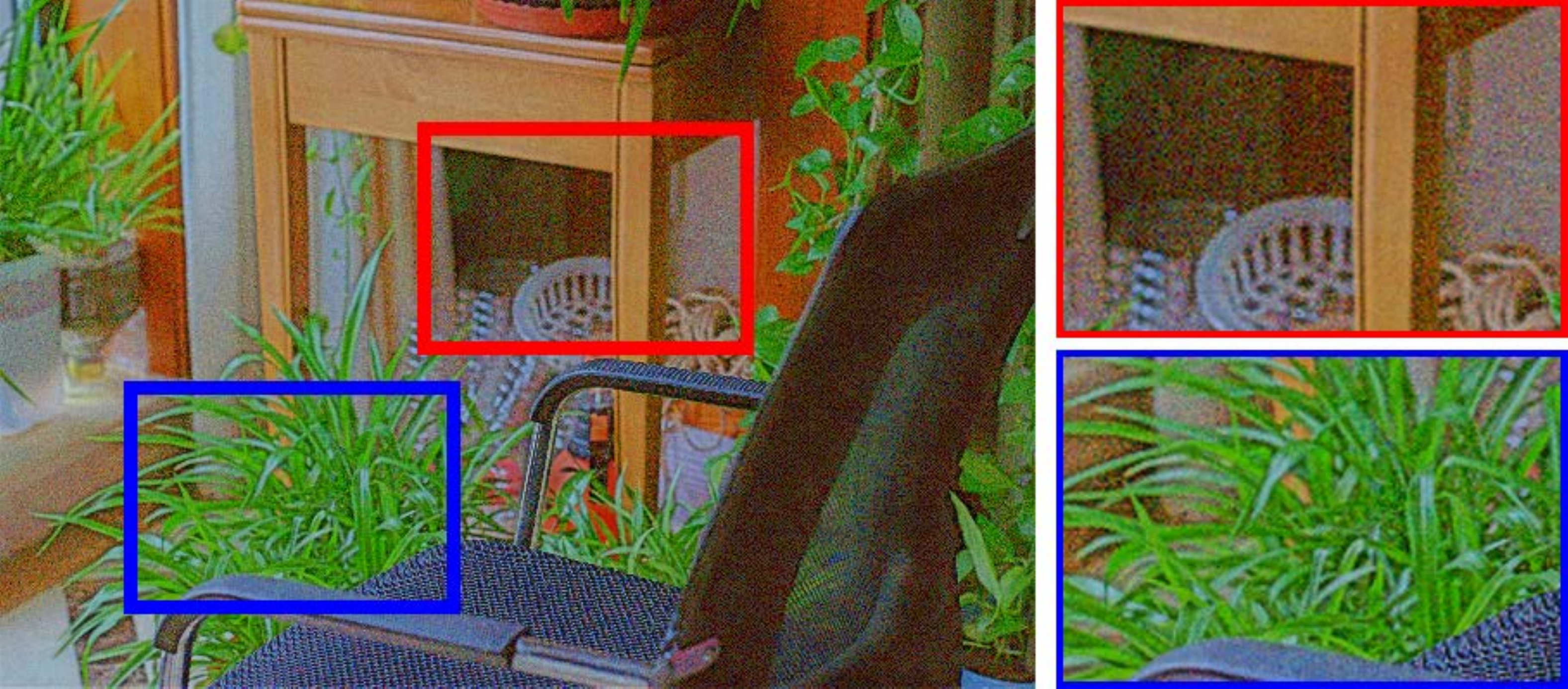}&
			\includegraphics[width=0.188\linewidth]{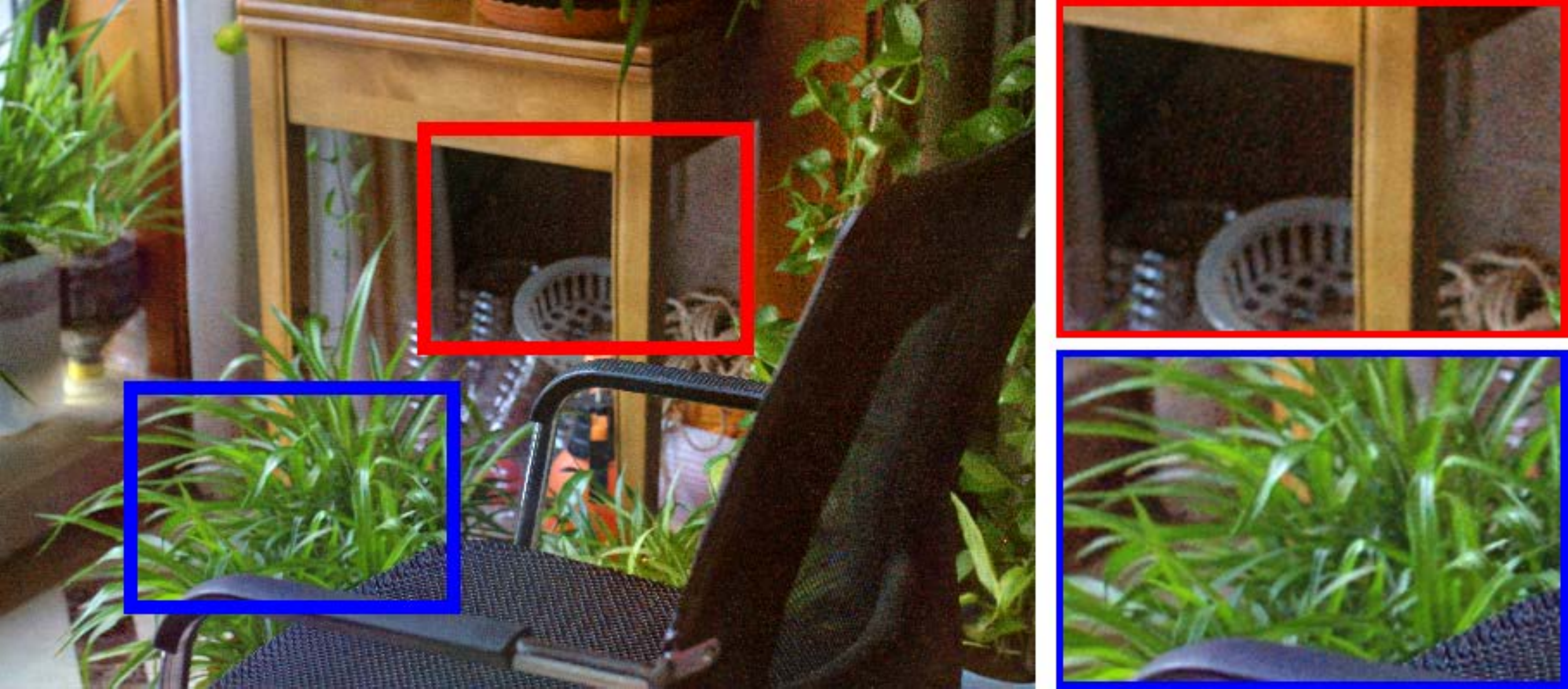}&
			\includegraphics[width=0.188\linewidth]{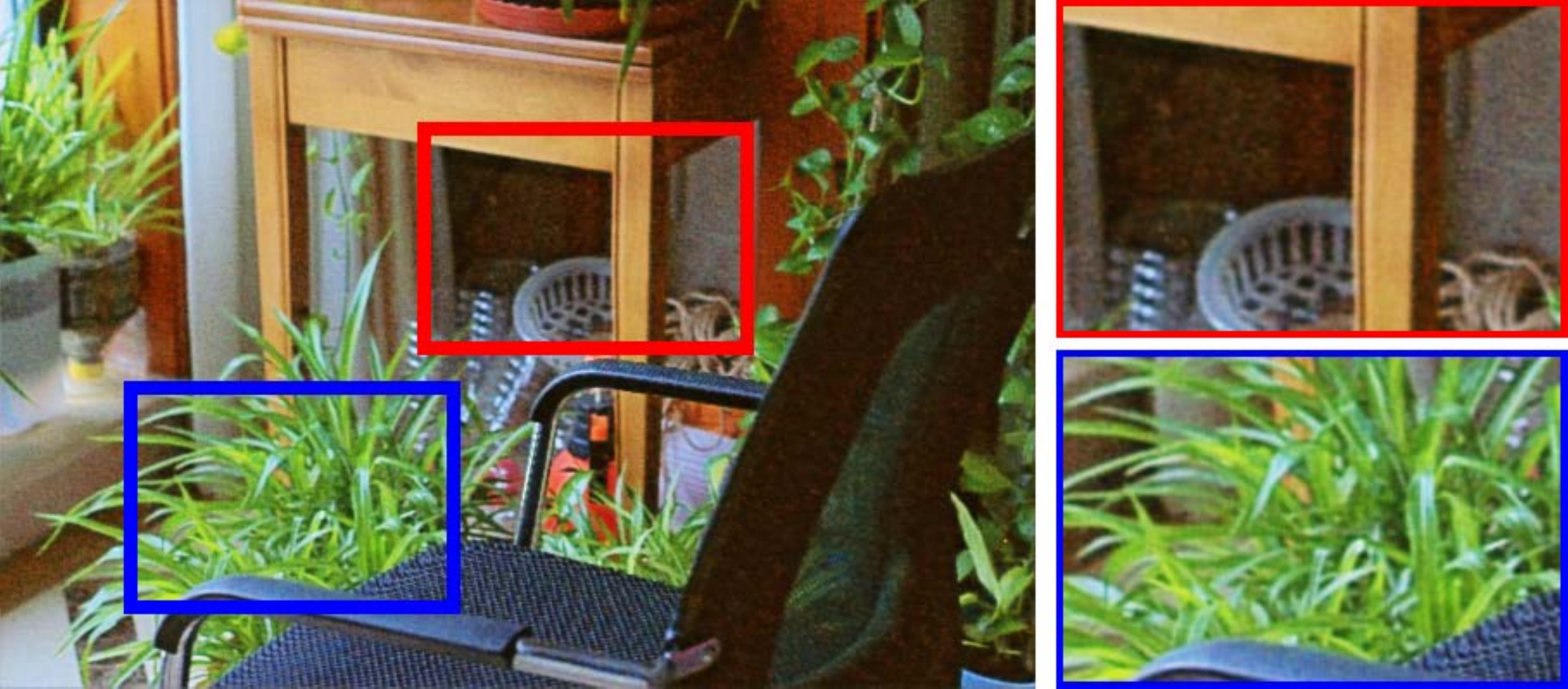}&
			\includegraphics[width=0.188\linewidth]{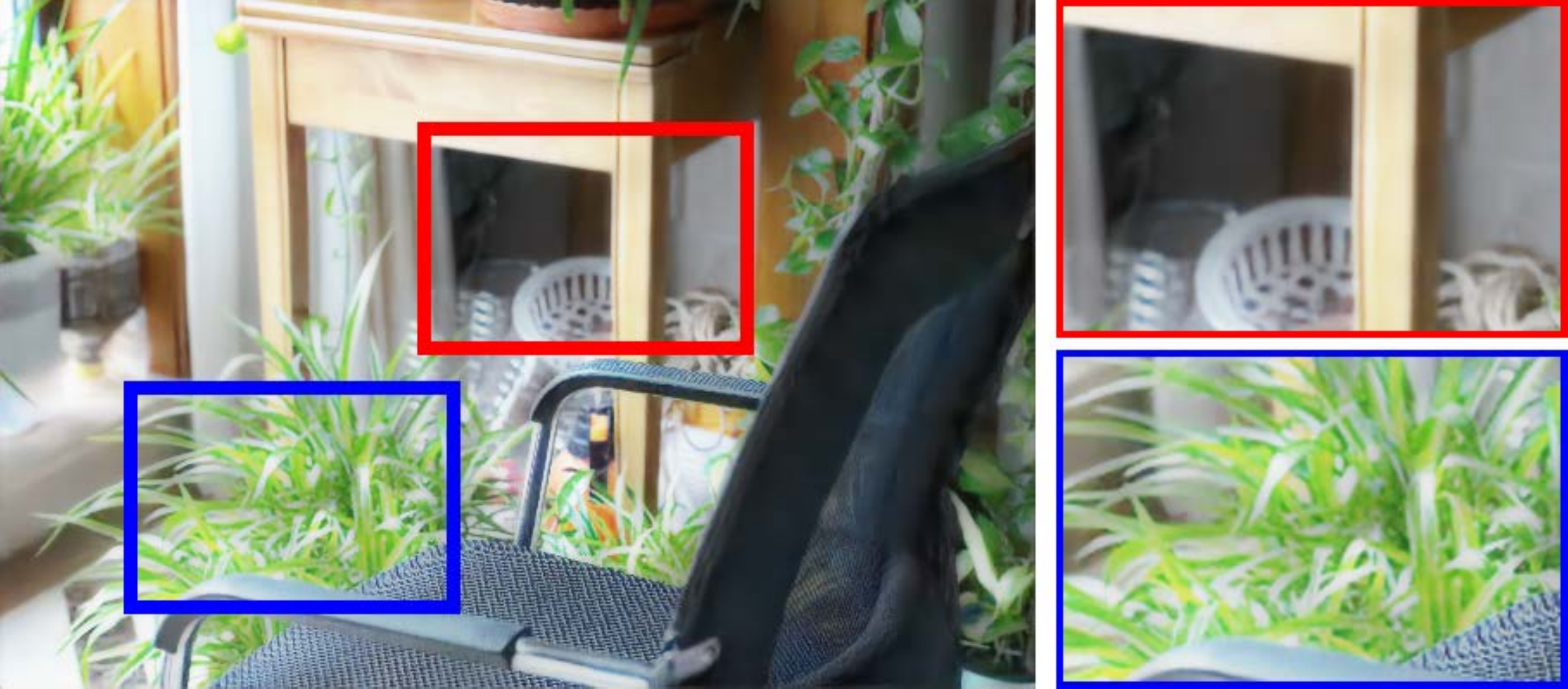}\\
			\footnotesize Input&\footnotesize{RetinexNet}&\footnotesize EnGAN&\footnotesize{SSIENet}&\footnotesize KinD\\
			\includegraphics[width=0.188\linewidth]{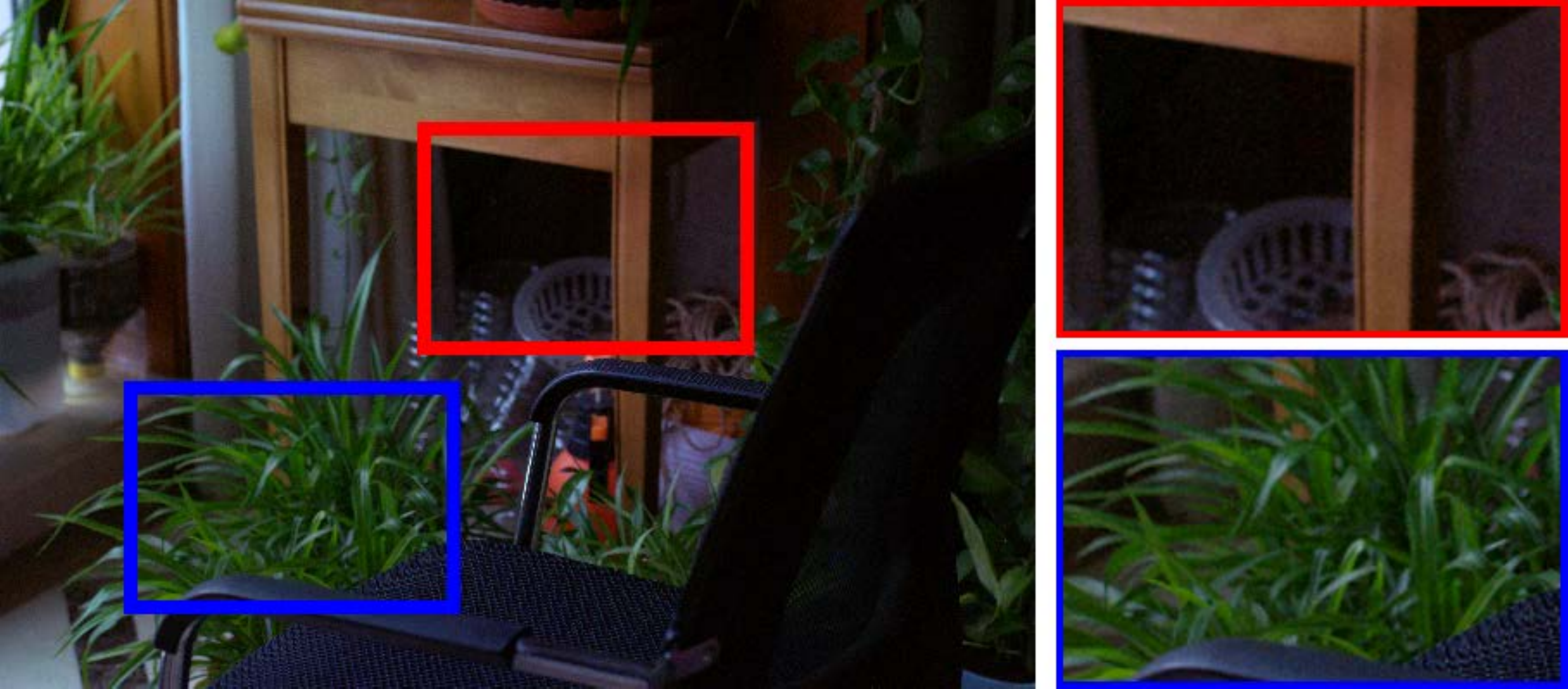}&
			\includegraphics[width=0.188\linewidth]{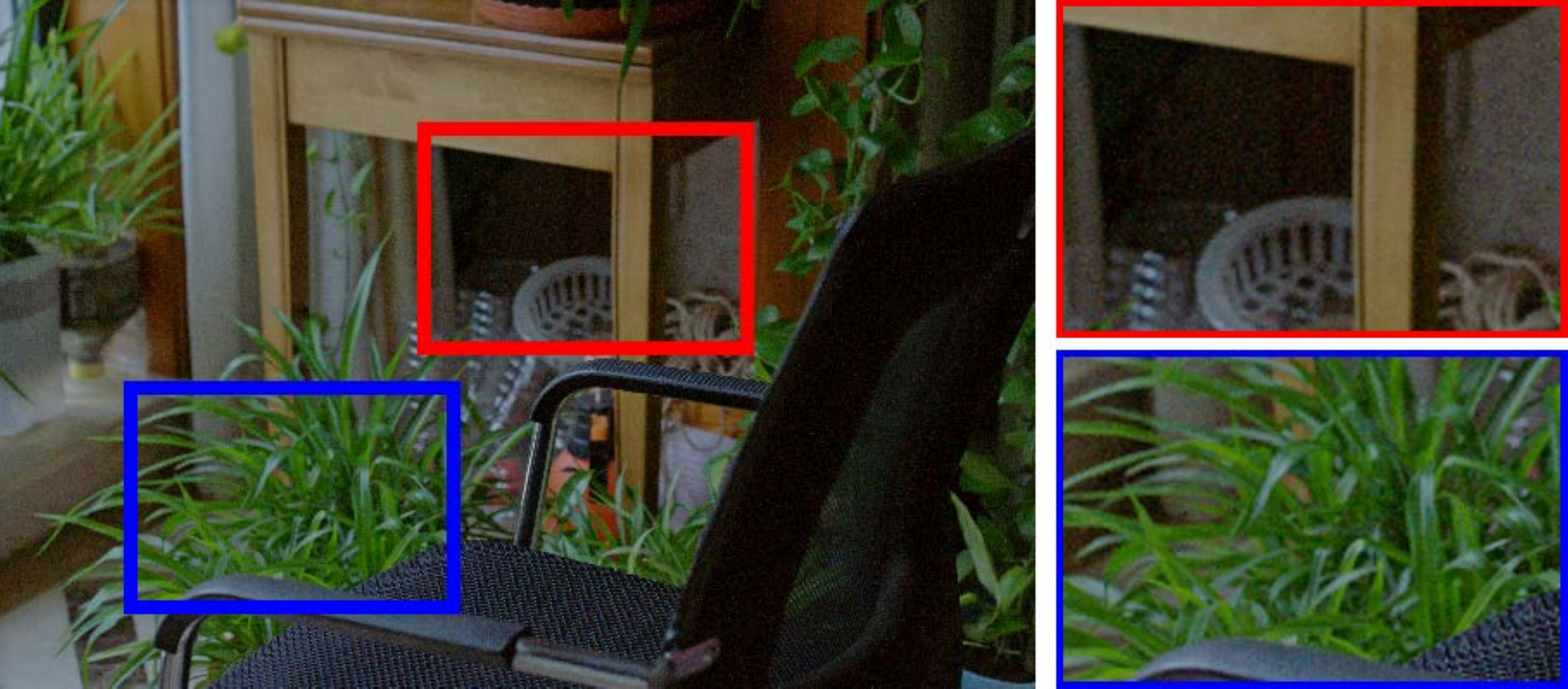}&
			\includegraphics[width=0.188\linewidth]{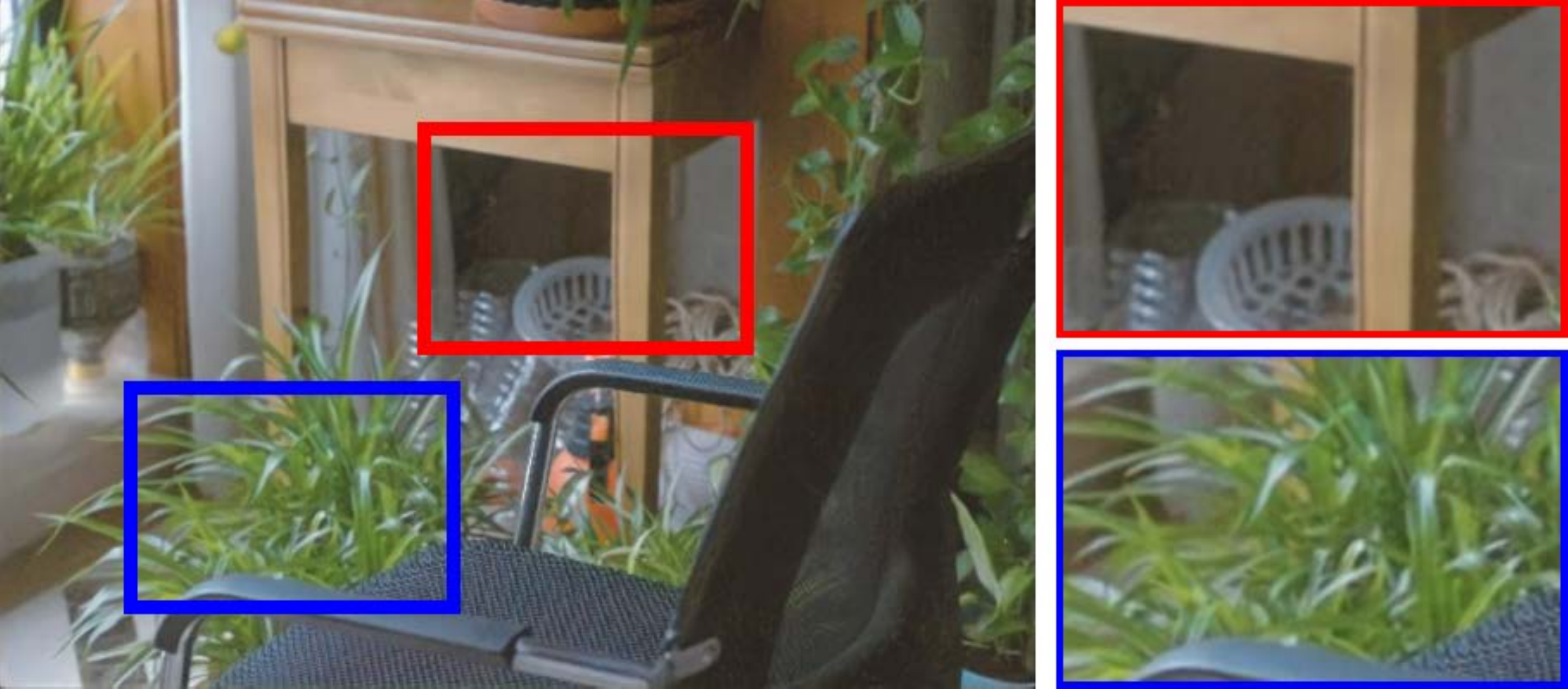}&
			\includegraphics[width=0.188\linewidth]{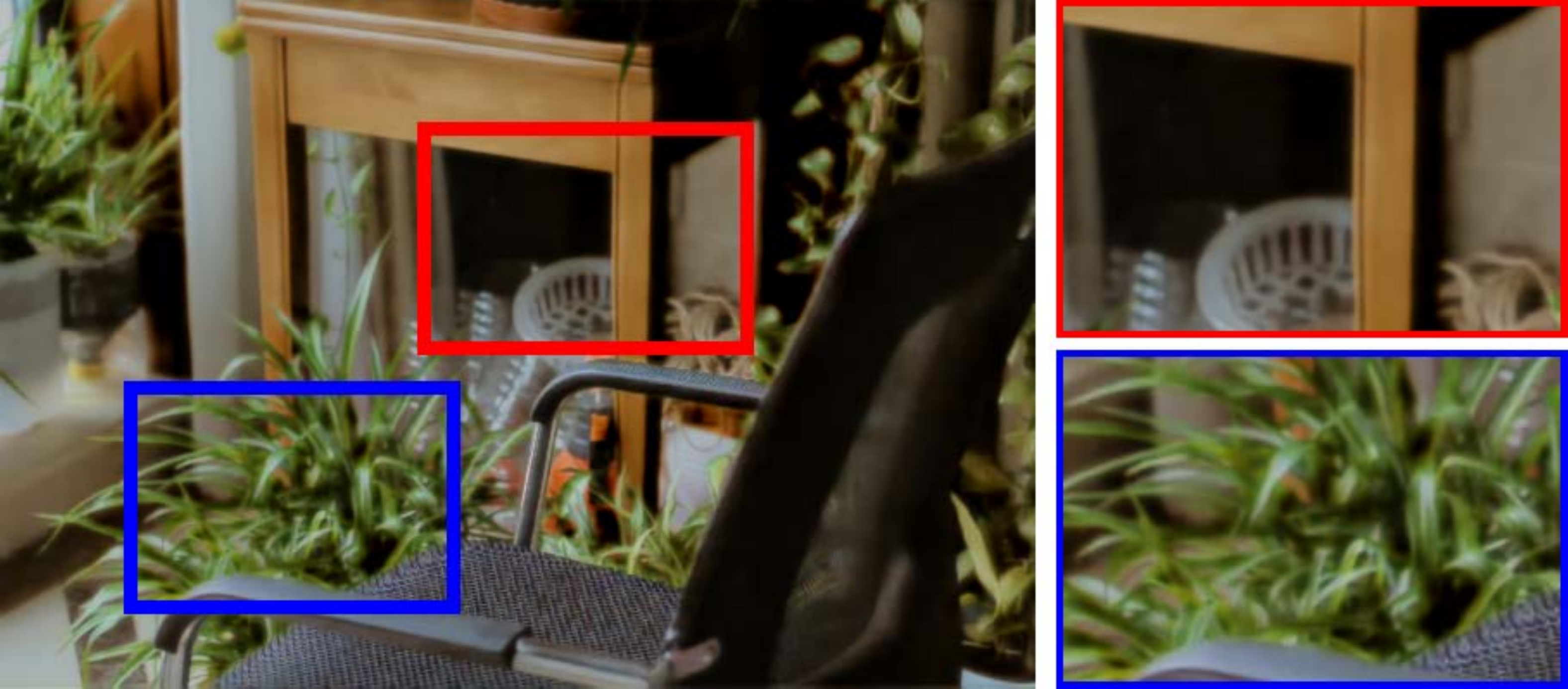}&
			\includegraphics[width=0.188\linewidth]{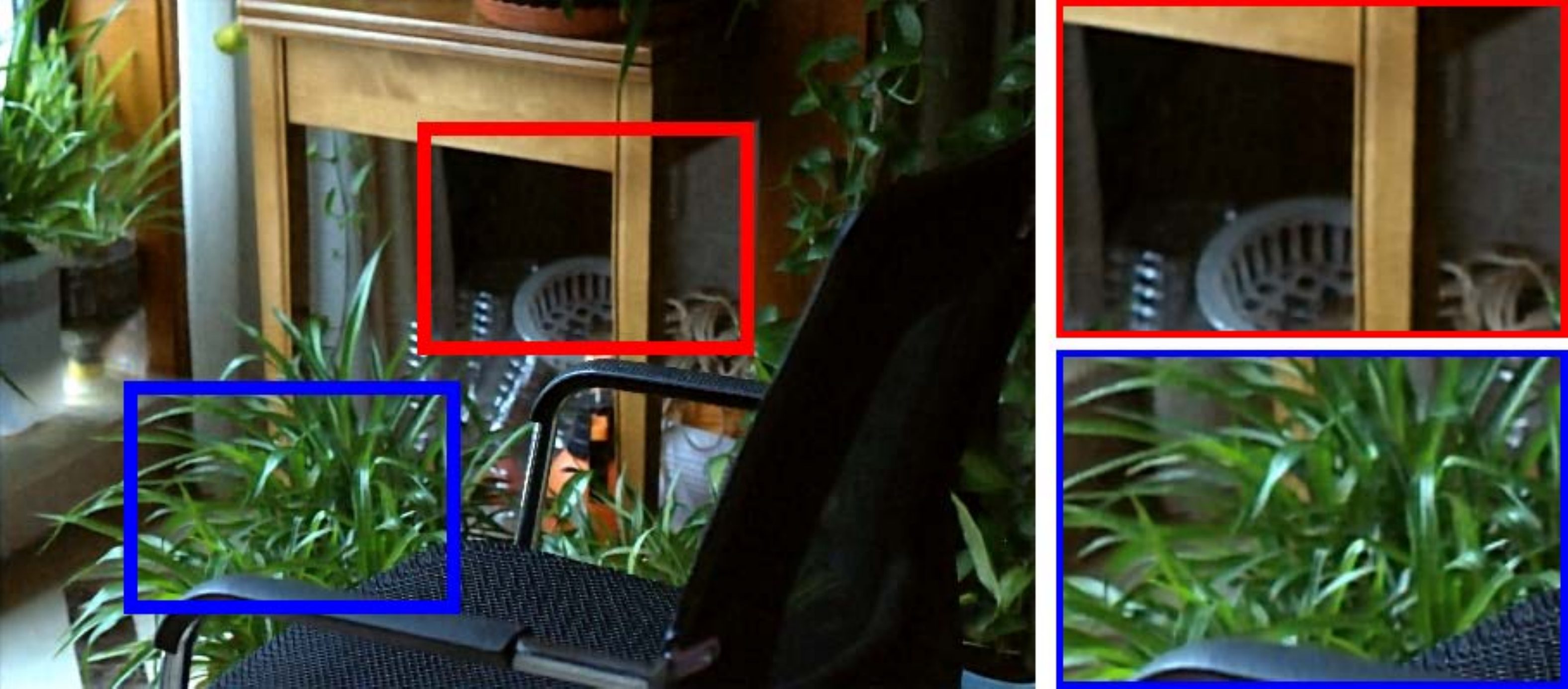}\\
			\footnotesize DeepUPE&\footnotesize{ZeroDCE}&\footnotesize FIDE&\footnotesize DRBN&\footnotesize{Ours}\\	
			\includegraphics[width=0.188\linewidth]{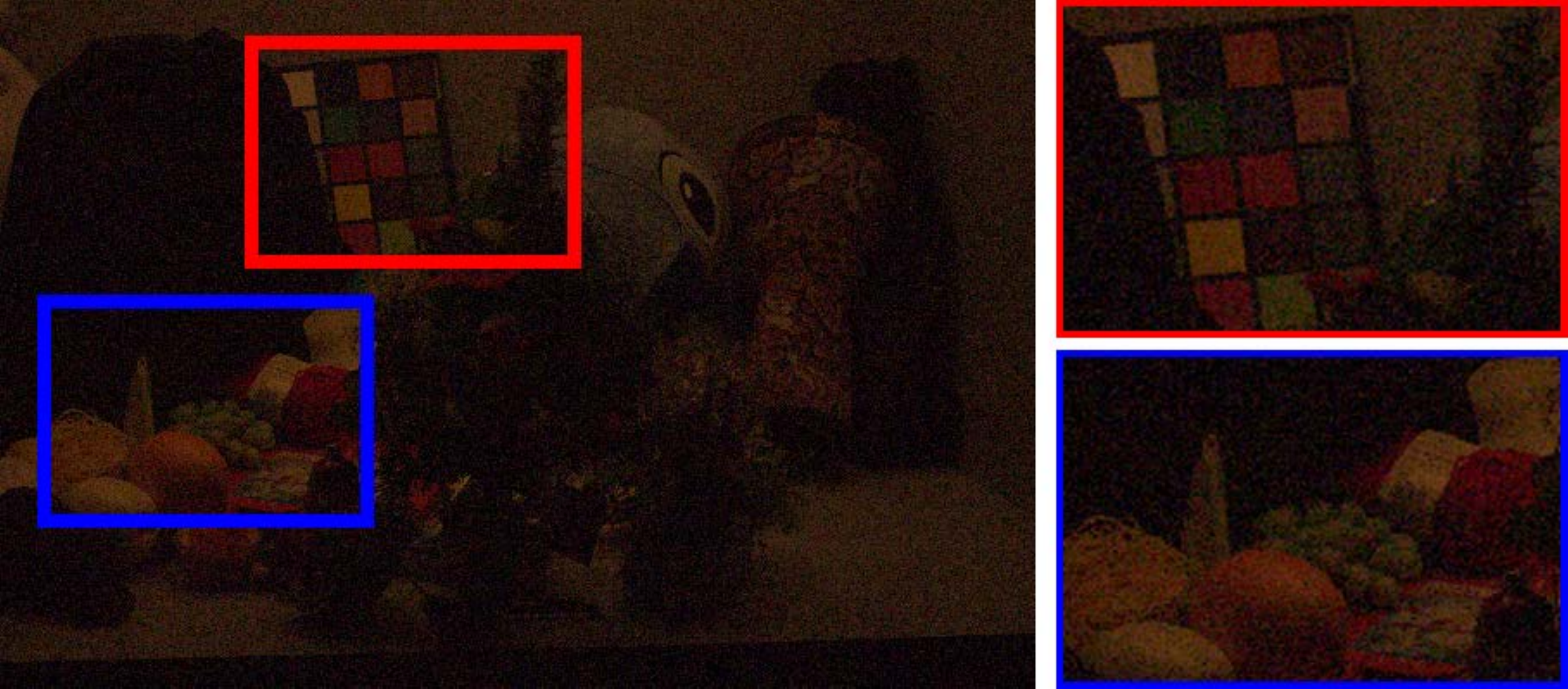}&
			\includegraphics[width=0.188\linewidth]{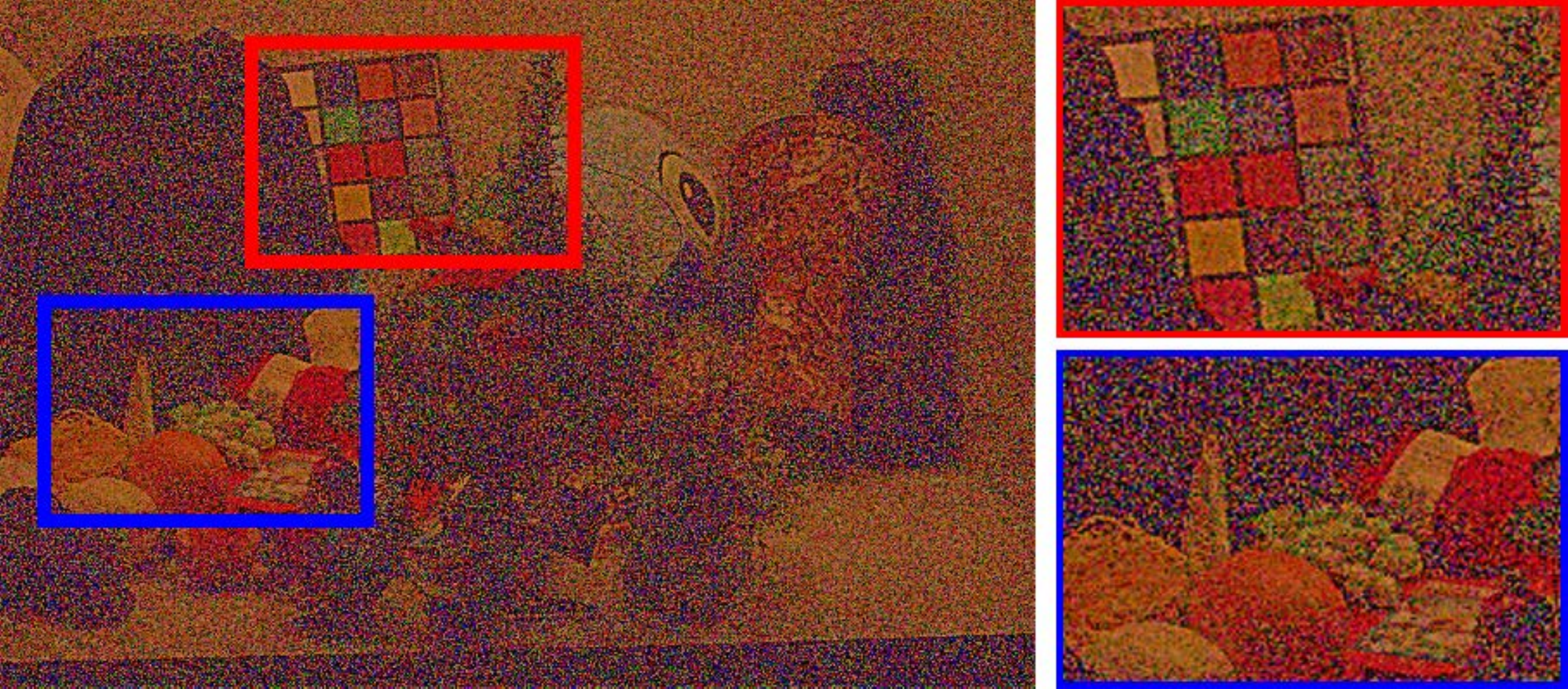}&
			\includegraphics[width=0.188\linewidth]{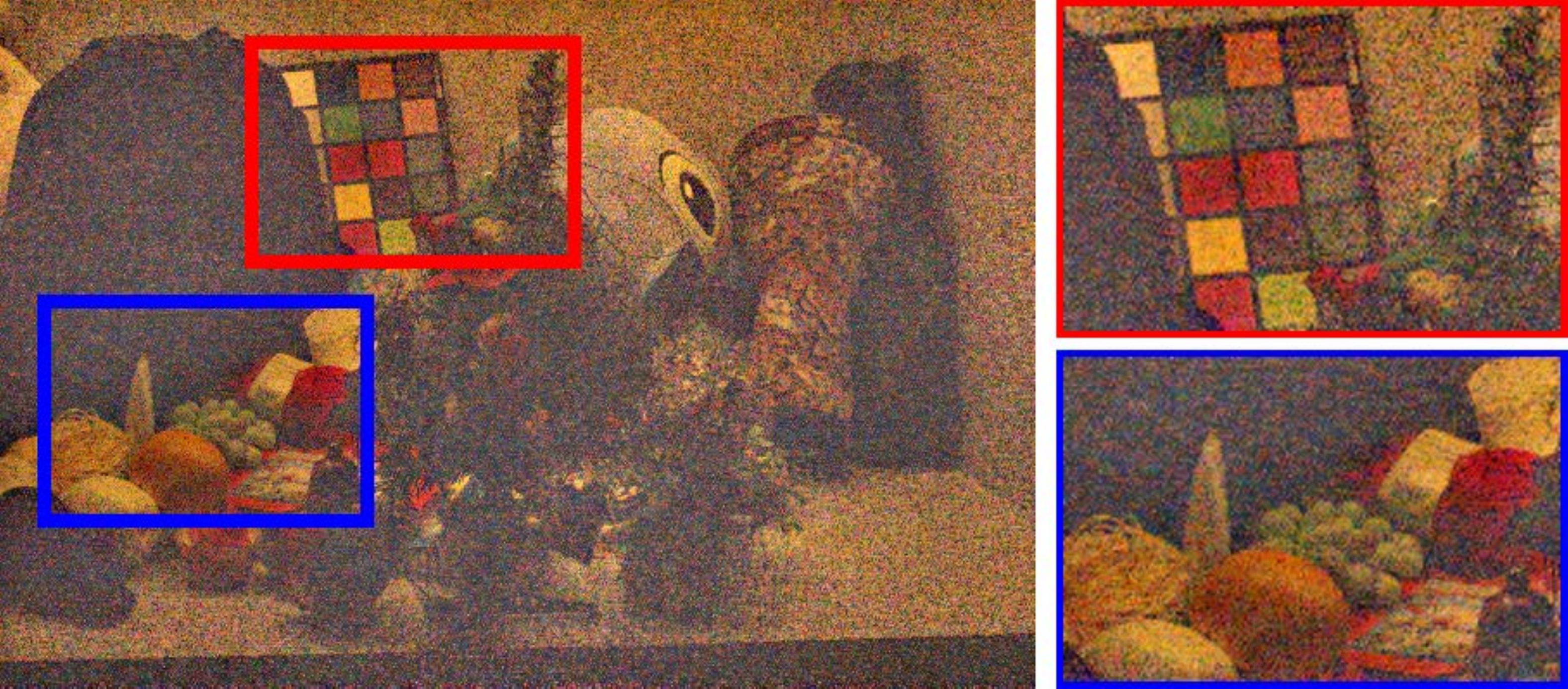}&
			\includegraphics[width=0.188\linewidth]{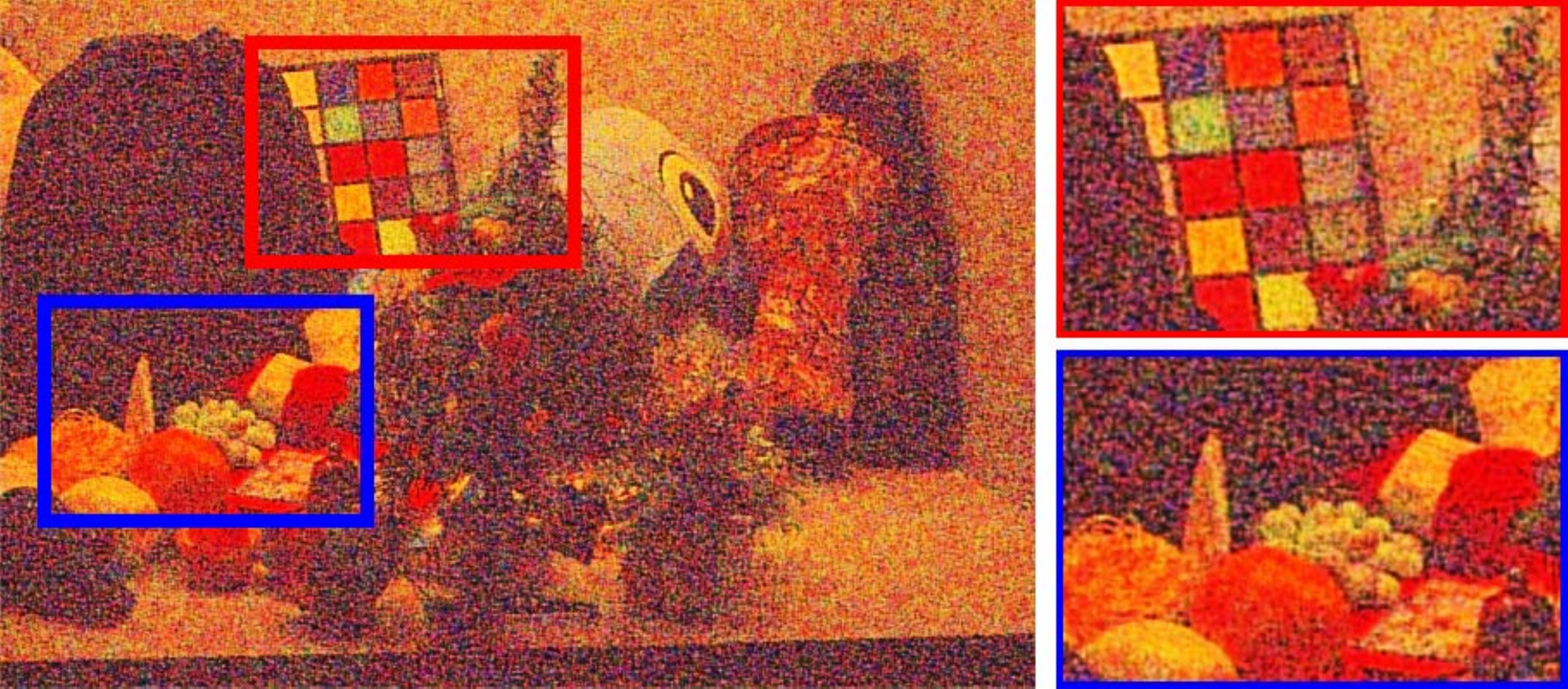}&
			\includegraphics[width=0.188\linewidth]{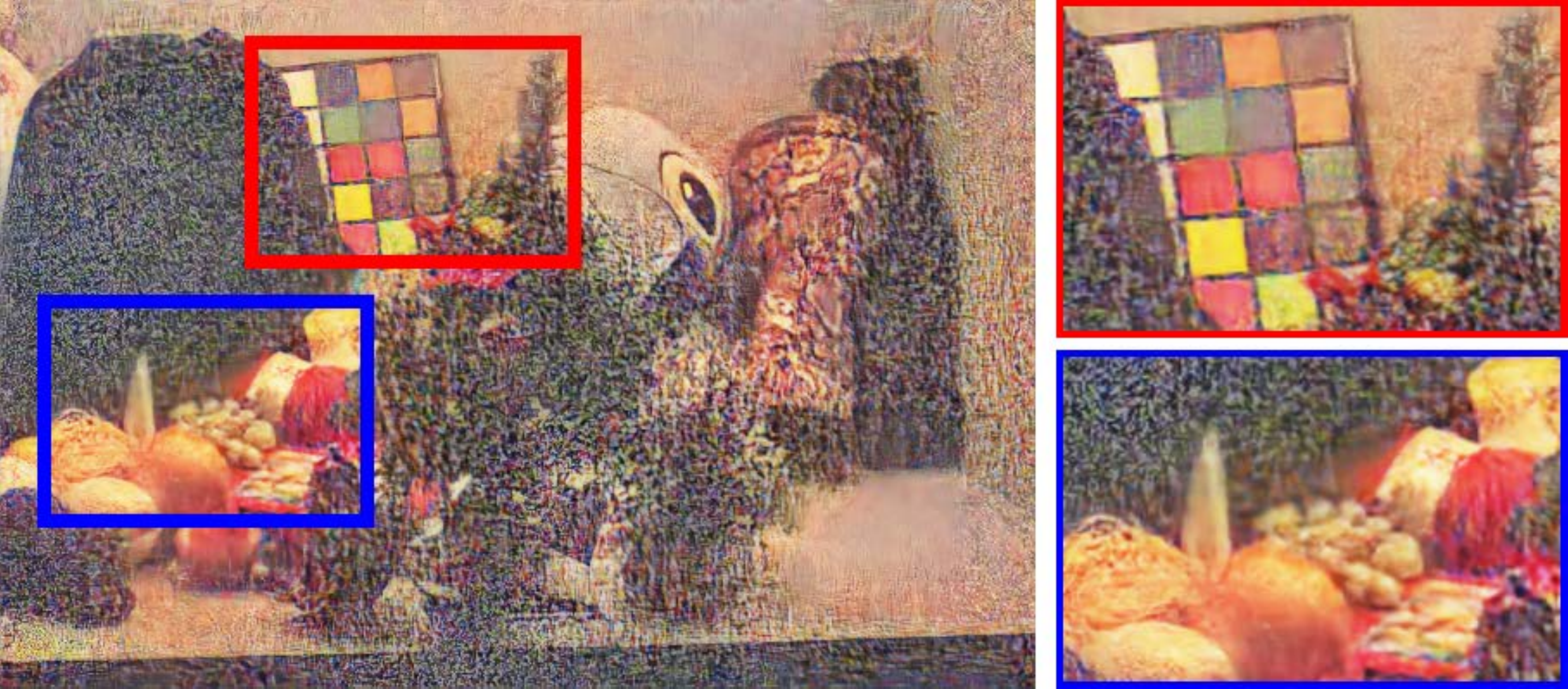}\\
			\footnotesize Input&\footnotesize{RetinexNet}&\footnotesize EnGAN&\footnotesize{SSIENet}&\footnotesize KinD\\
			\includegraphics[width=0.188\linewidth]{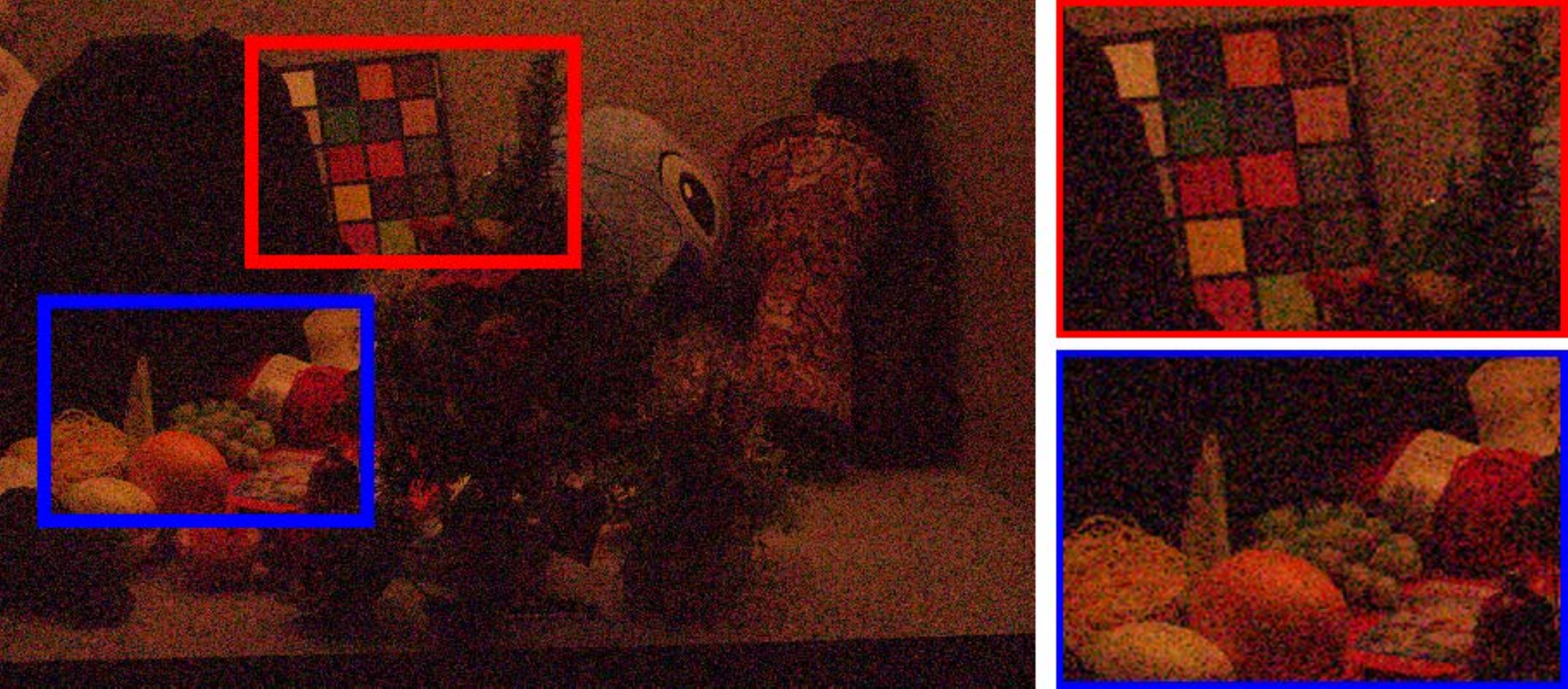}&
			\includegraphics[width=0.188\linewidth]{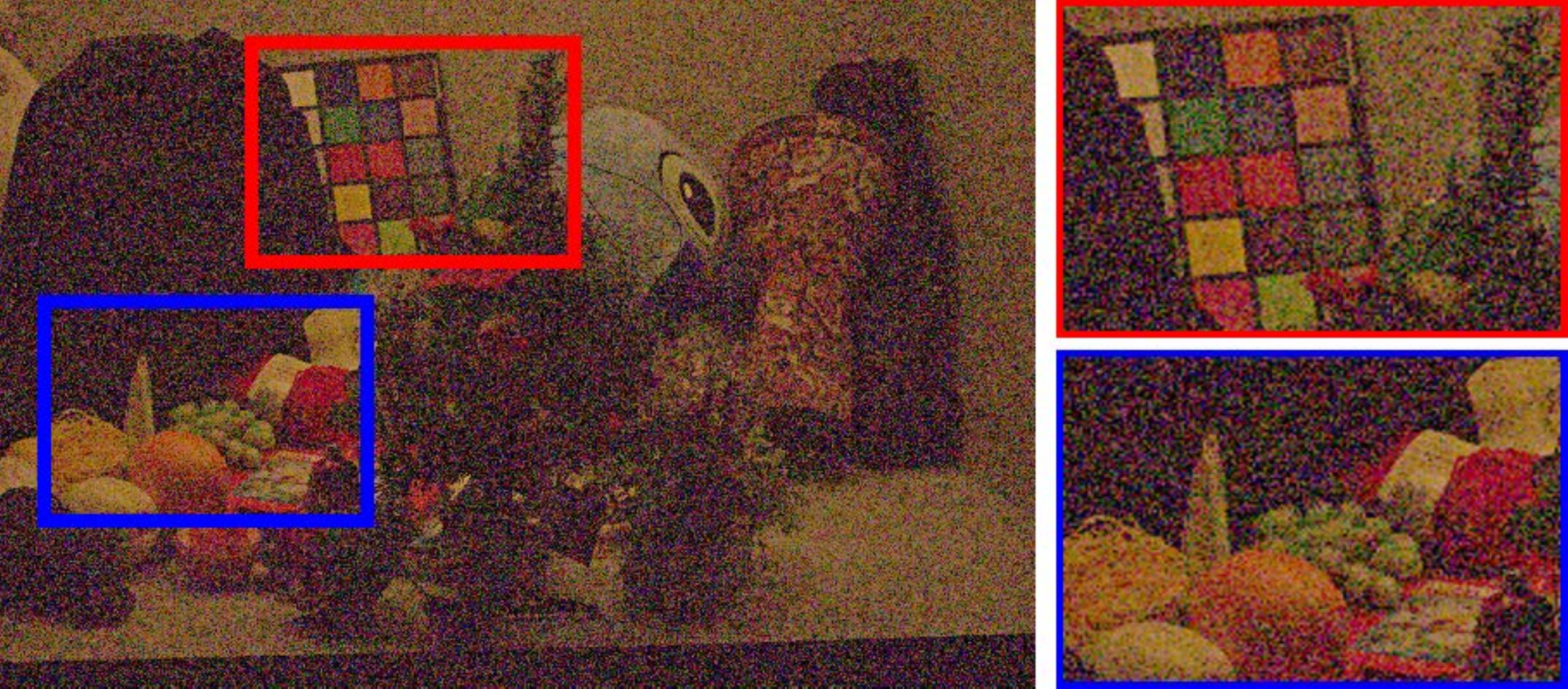}&
			\includegraphics[width=0.188\linewidth]{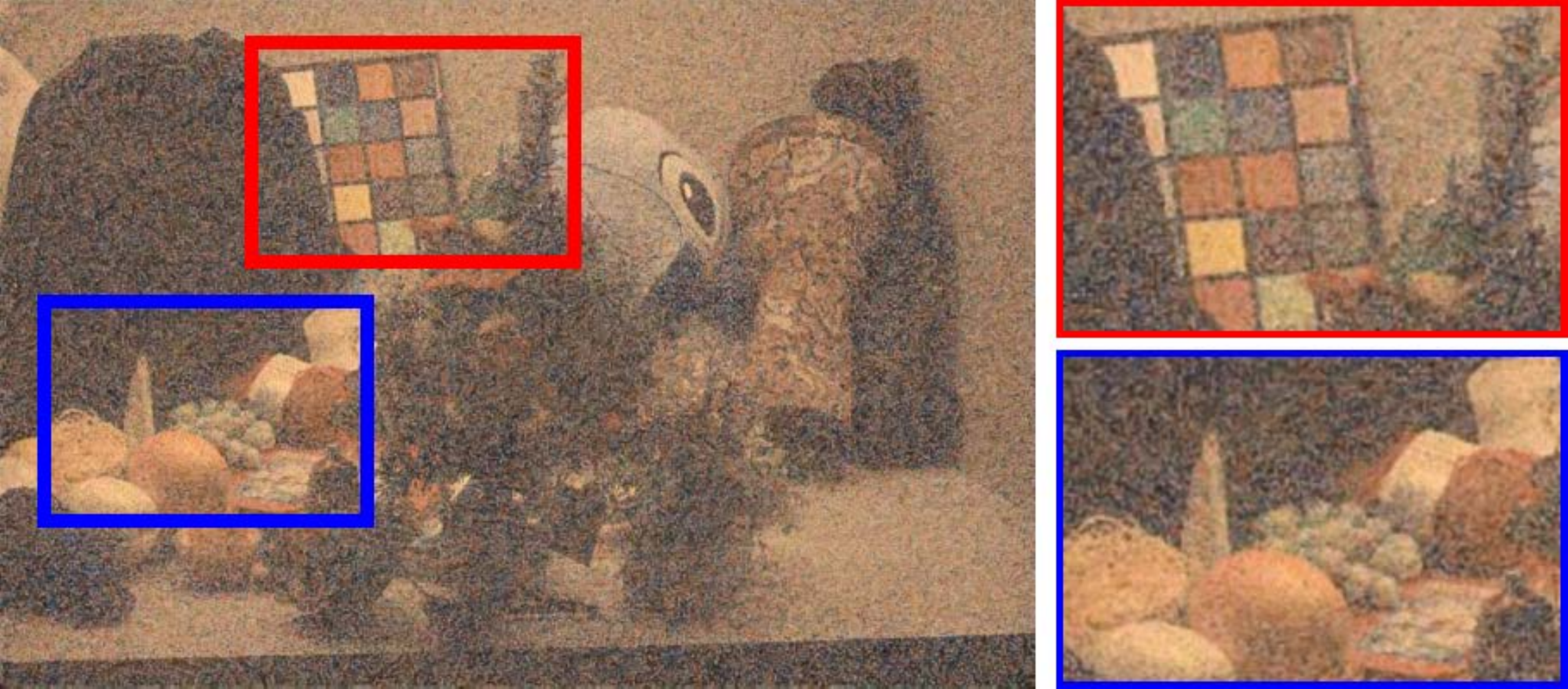}&
			\includegraphics[width=0.188\linewidth]{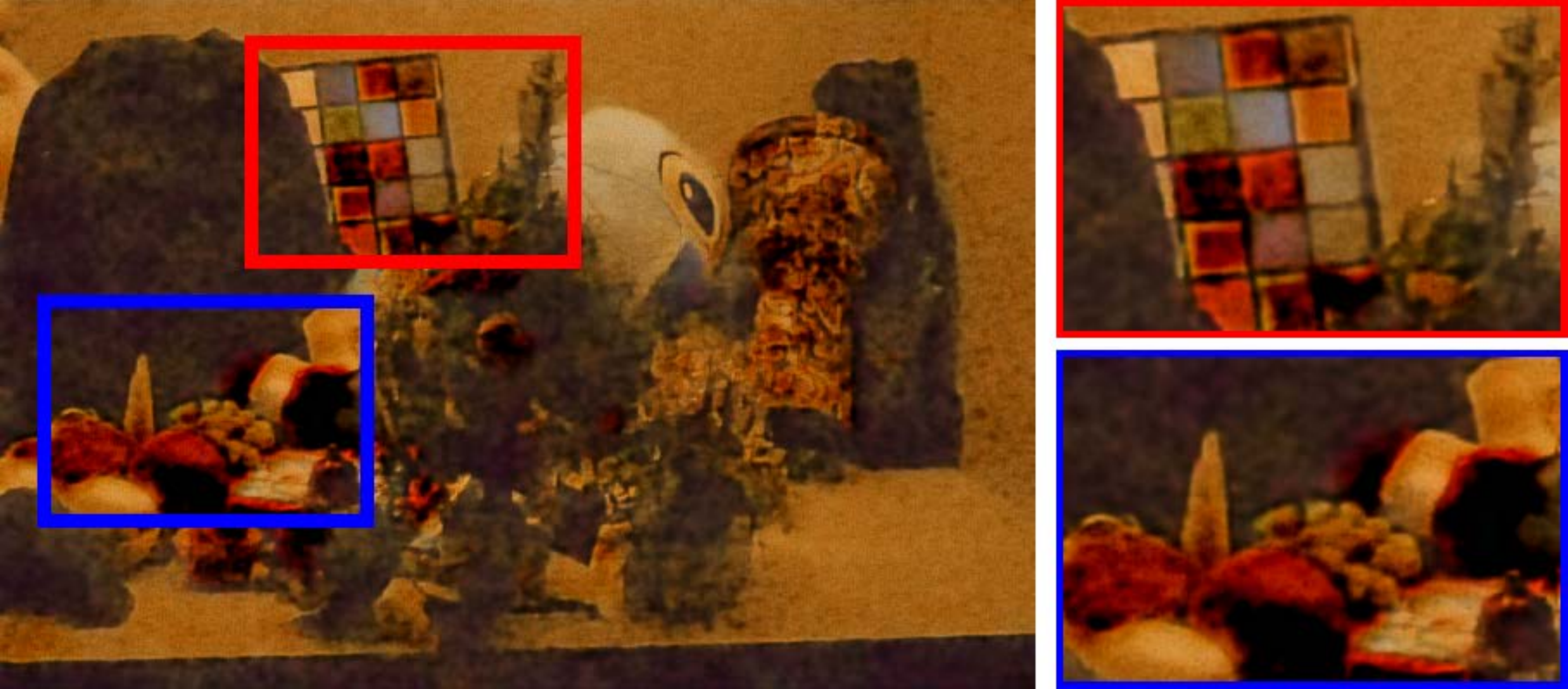}&
			\includegraphics[width=0.188\linewidth]{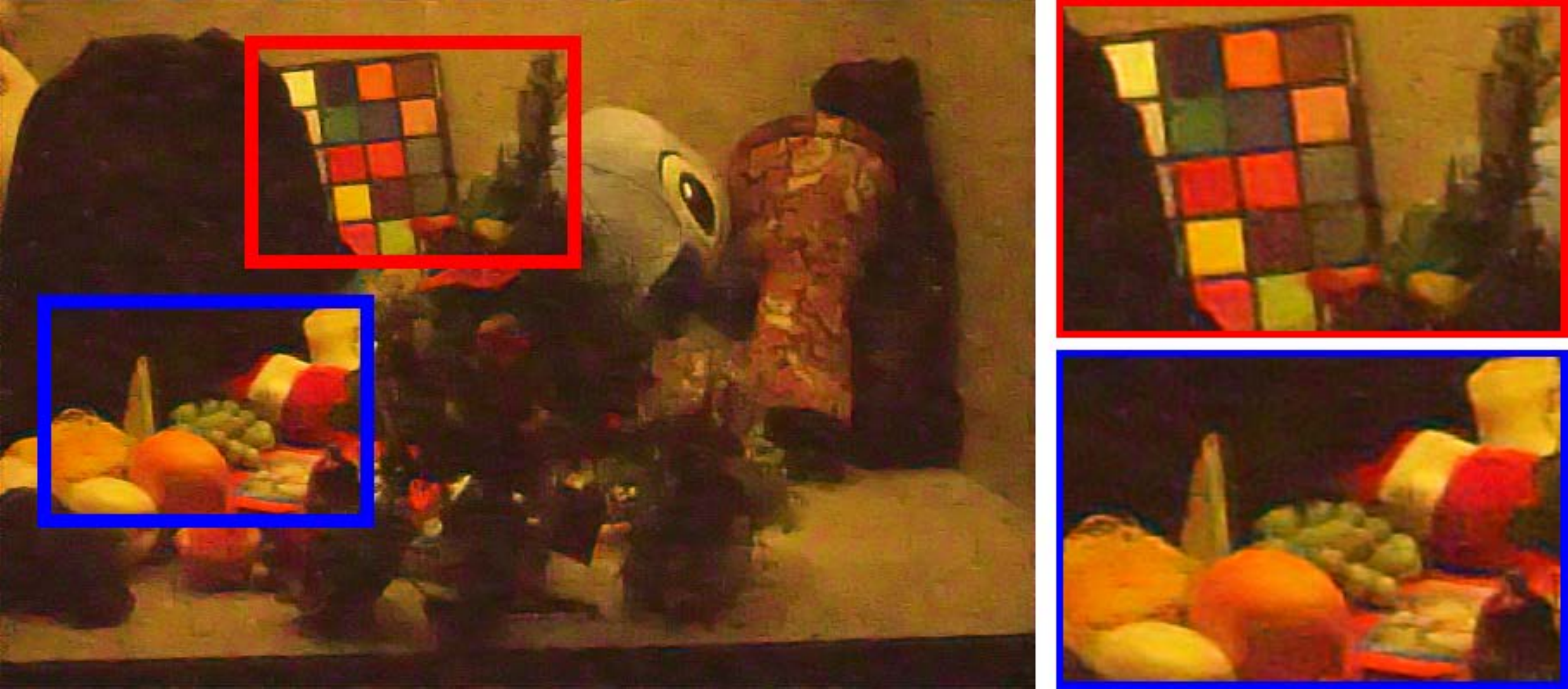}\\
			\footnotesize DeepUPE&\footnotesize{ZeroDCE}&\footnotesize FIDE&\footnotesize DRBN&\footnotesize{Ours}\\		
		\end{tabular}	
	\end{center}
	\vspace{-0.05cm}
	\caption{Visual results of state-of-the-art methods and ours on the LOL dataset. Red and blue boxes indicate the obvious differences. }
	\label{fig:LOL}
\end{figure*}

\begin{figure*}[t]
	\centering
	\begin{minipage}{0.30\textwidth}
		\vspace{0.05cm}
		\subfigure{
			\begin{minipage}{1\textwidth}
				\includegraphics[width=1\textwidth]{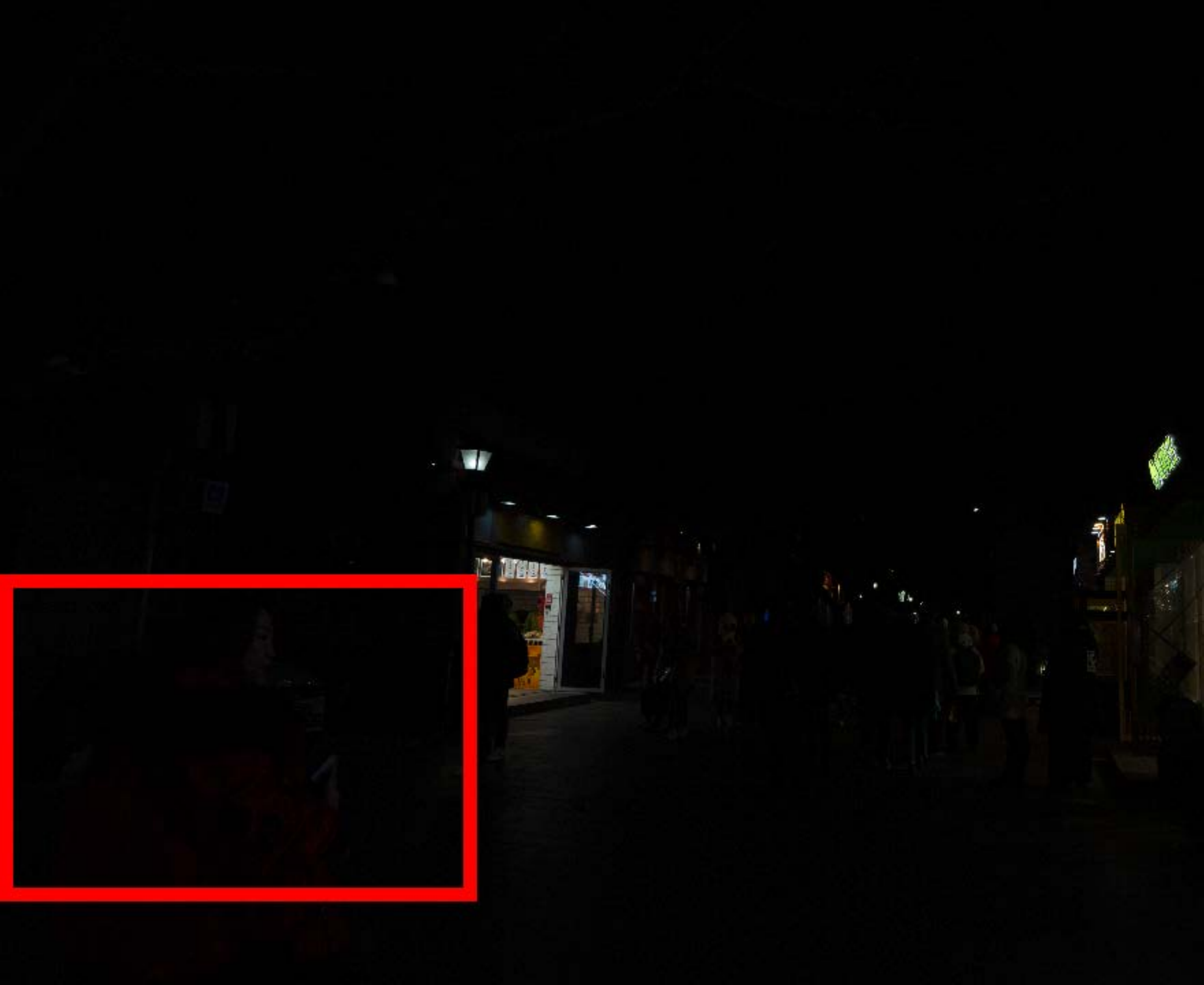}
				\centering  \footnotesize Input\\
			\end{minipage}
		}
	\end{minipage}
	\begin{minipage}{0.16\textwidth}
		\subfigure{
			\begin{minipage}{1\textwidth}
				\includegraphics[width=1\textwidth]{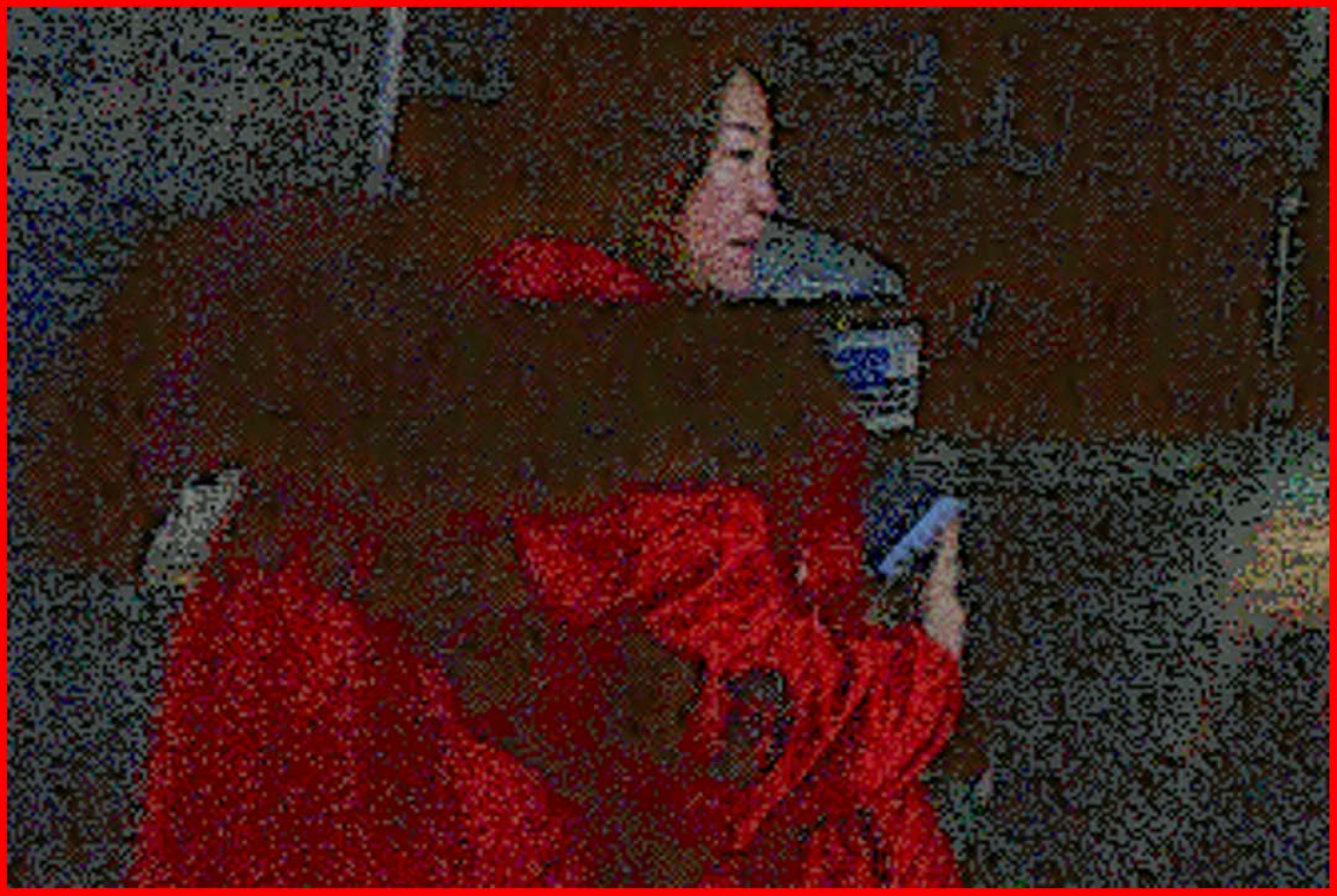}
				\centering \footnotesize RetinexNet \vspace{-0.7em}\\
			\end{minipage}
		}
		\subfigure{
			\begin{minipage}{1\textwidth}
				\includegraphics[width=1\textwidth]{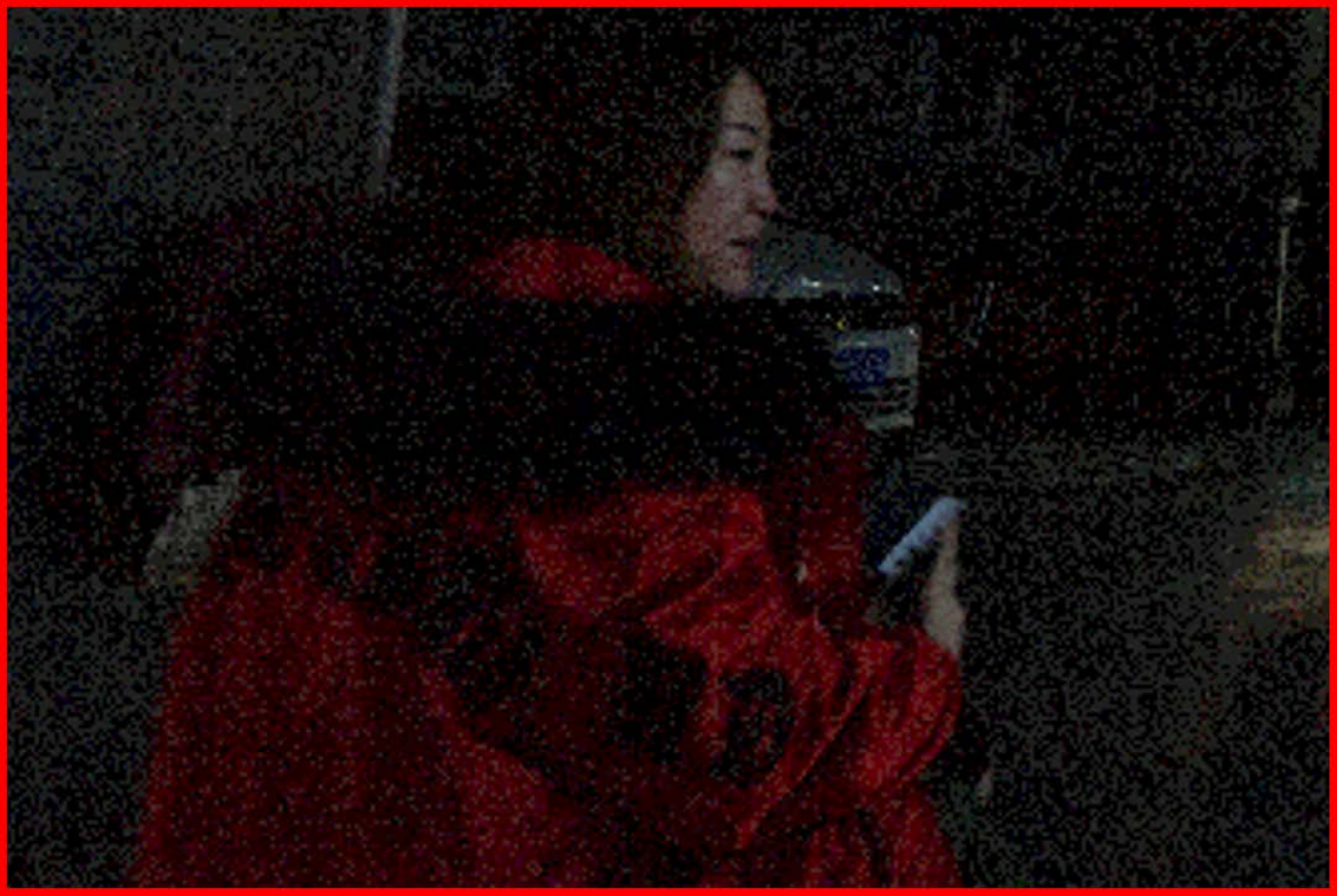}
				\centering \footnotesize ZeroDCE\\
			\end{minipage}
		}
	\end{minipage}
	\begin{minipage}{0.16\textwidth}
		\subfigure{
			\begin{minipage}{1\textwidth}
				\includegraphics[width=1\textwidth]{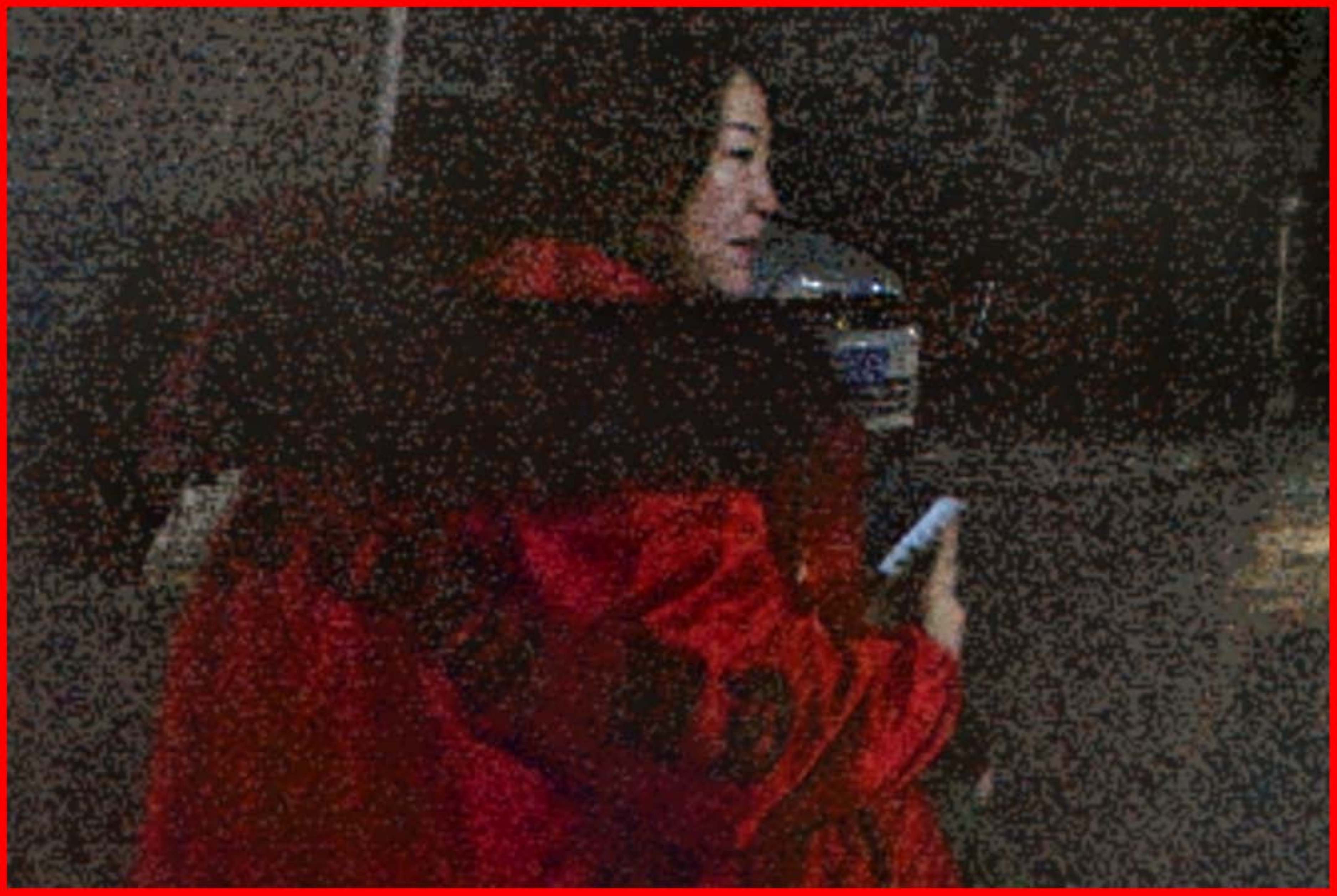}
				\centering \footnotesize  EnGAN \vspace{-0.7em}\\
			\end{minipage}
		}
		\subfigure{
			\begin{minipage}{1\textwidth}
				\includegraphics[width=1\textwidth]{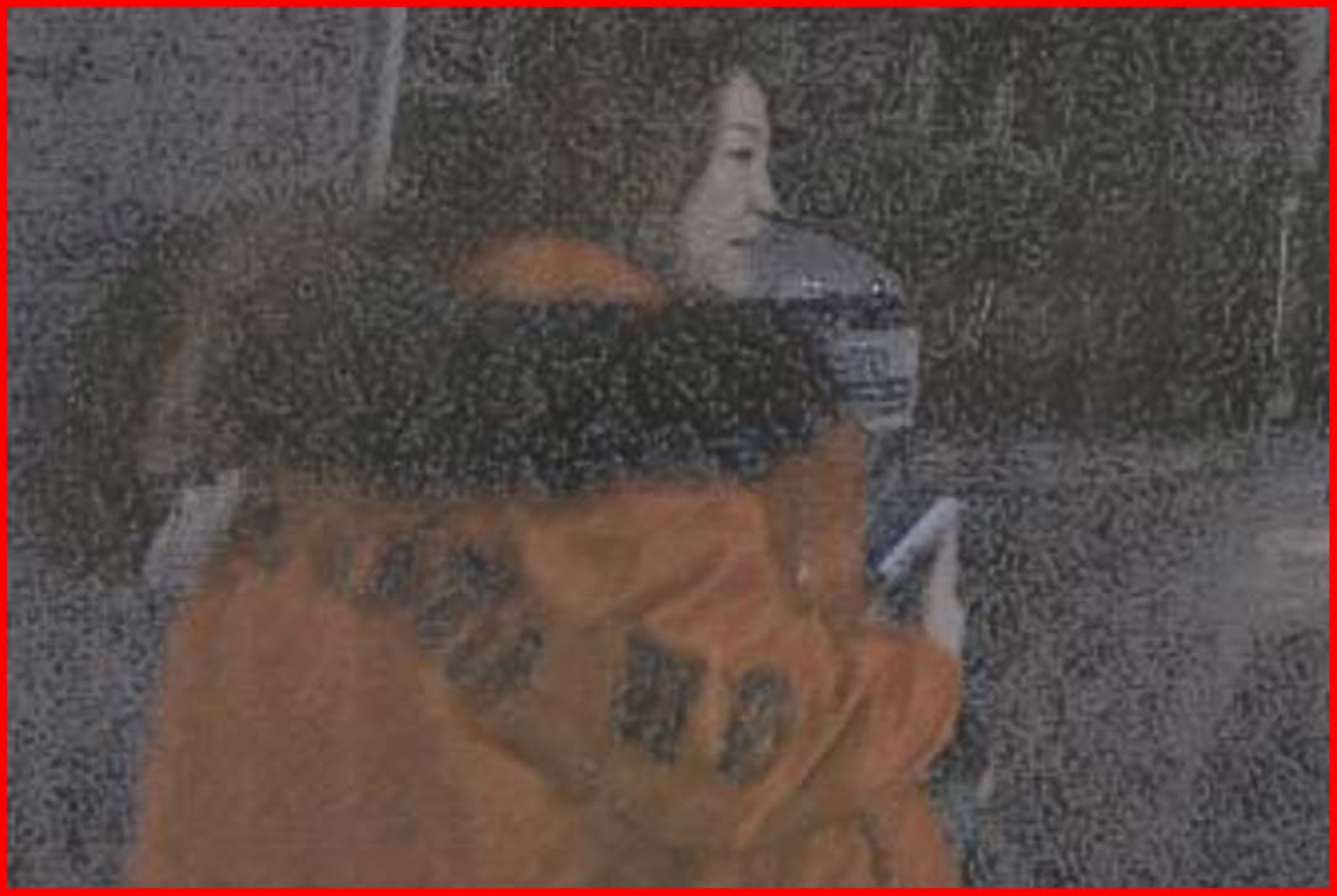}
				\centering \footnotesize  FIDE\\
			\end{minipage}
		}
	\end{minipage}
	\begin{minipage}{0.16\textwidth}
		\subfigure{
			\begin{minipage}{1\textwidth}
				\includegraphics[width=1\textwidth]{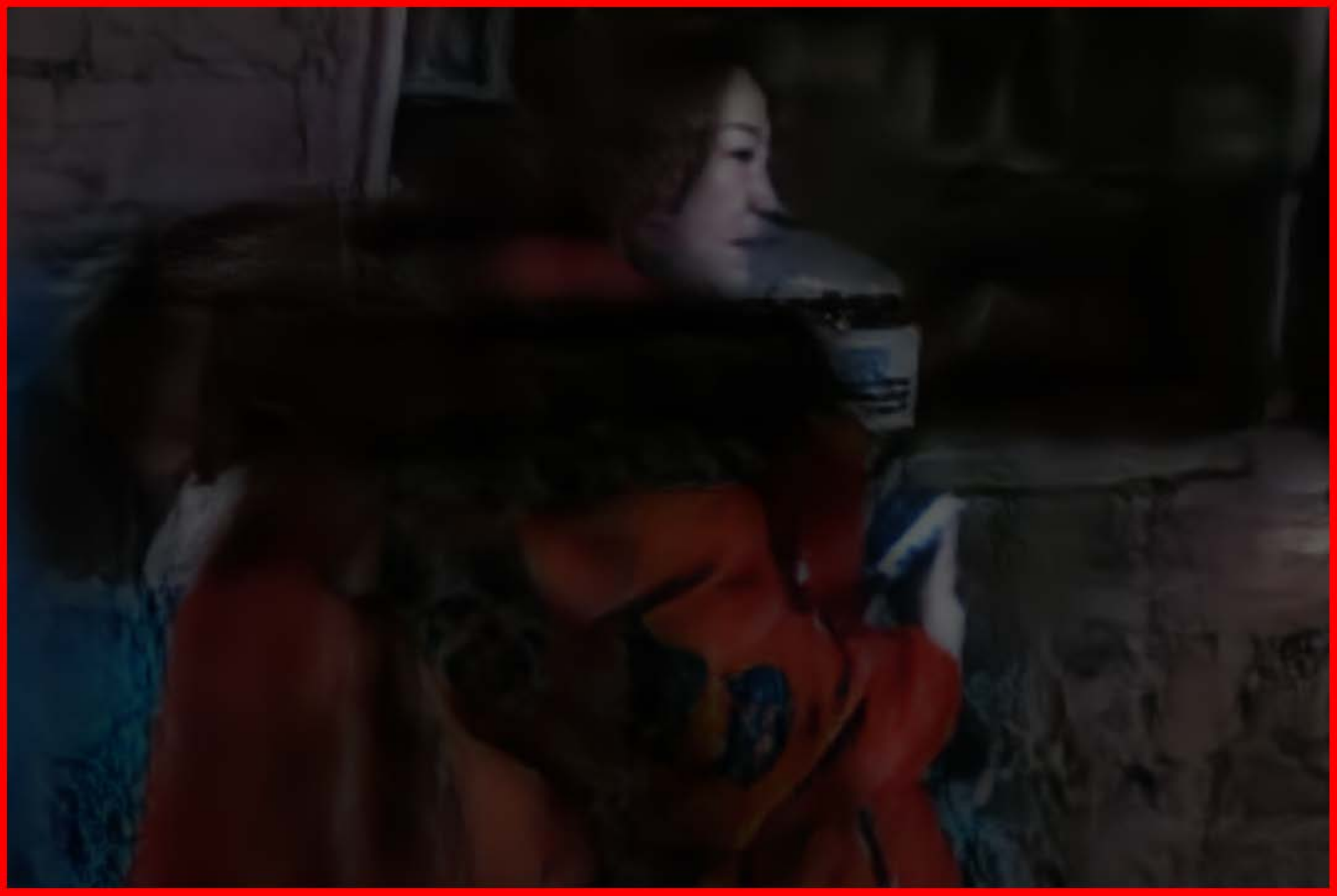}
				\centering \footnotesize KinD \vspace{-0.7em}\\
			\end{minipage}
		}
		\subfigure{
			\begin{minipage}{1\textwidth}
				\includegraphics[width=1\textwidth]{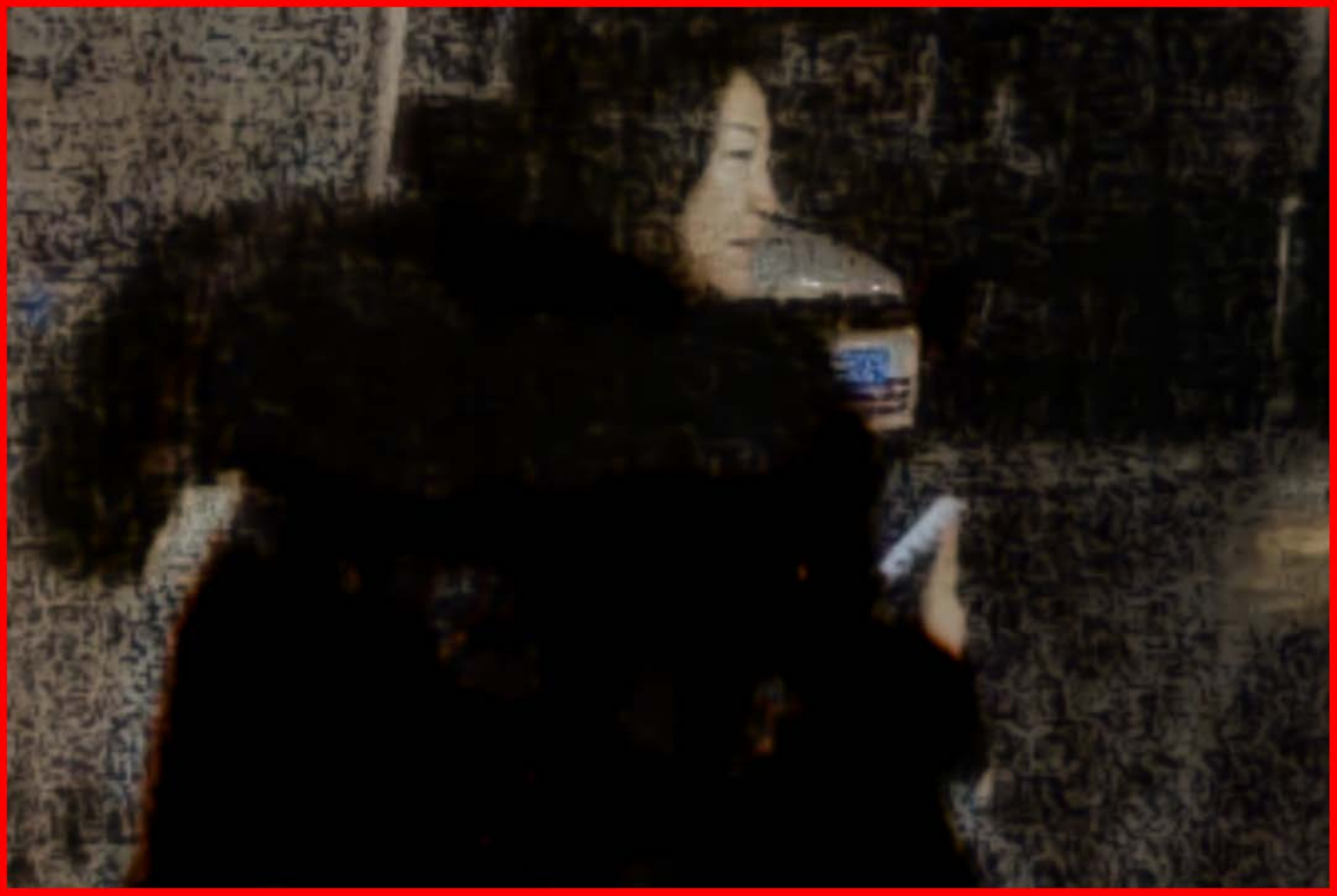}
				\centering \footnotesize DRBN\\
			\end{minipage}
		}
	\end{minipage}
	\begin{minipage}{0.16\textwidth}
		\subfigure{
			\begin{minipage}{1\textwidth}
				\includegraphics[width=1\textwidth]{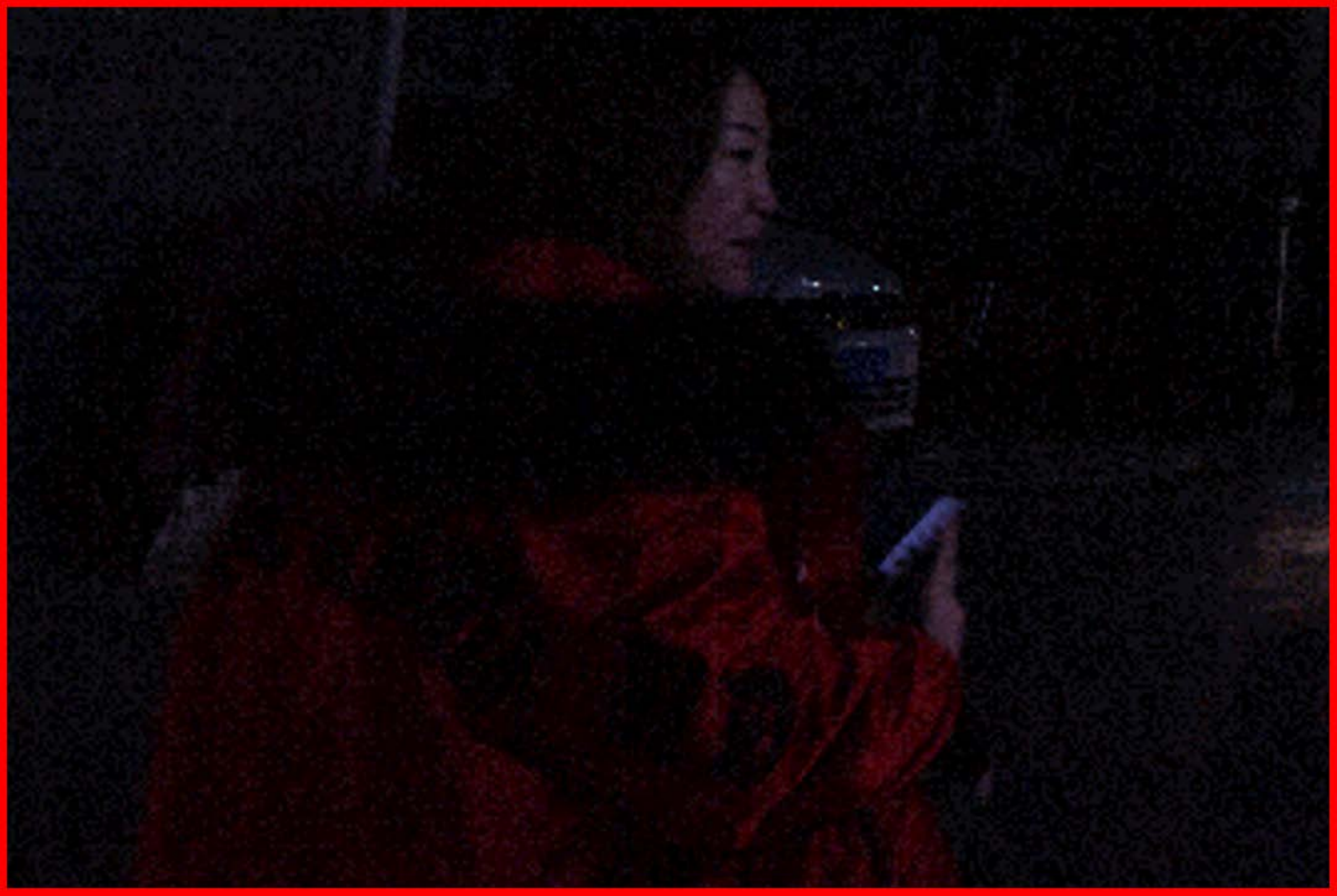}
				\centering \footnotesize DeepUPE \vspace{-0.9em}\\
			\end{minipage}
		}
		\subfigure{
			\begin{minipage}{1\textwidth}
				\includegraphics[width=1\textwidth]{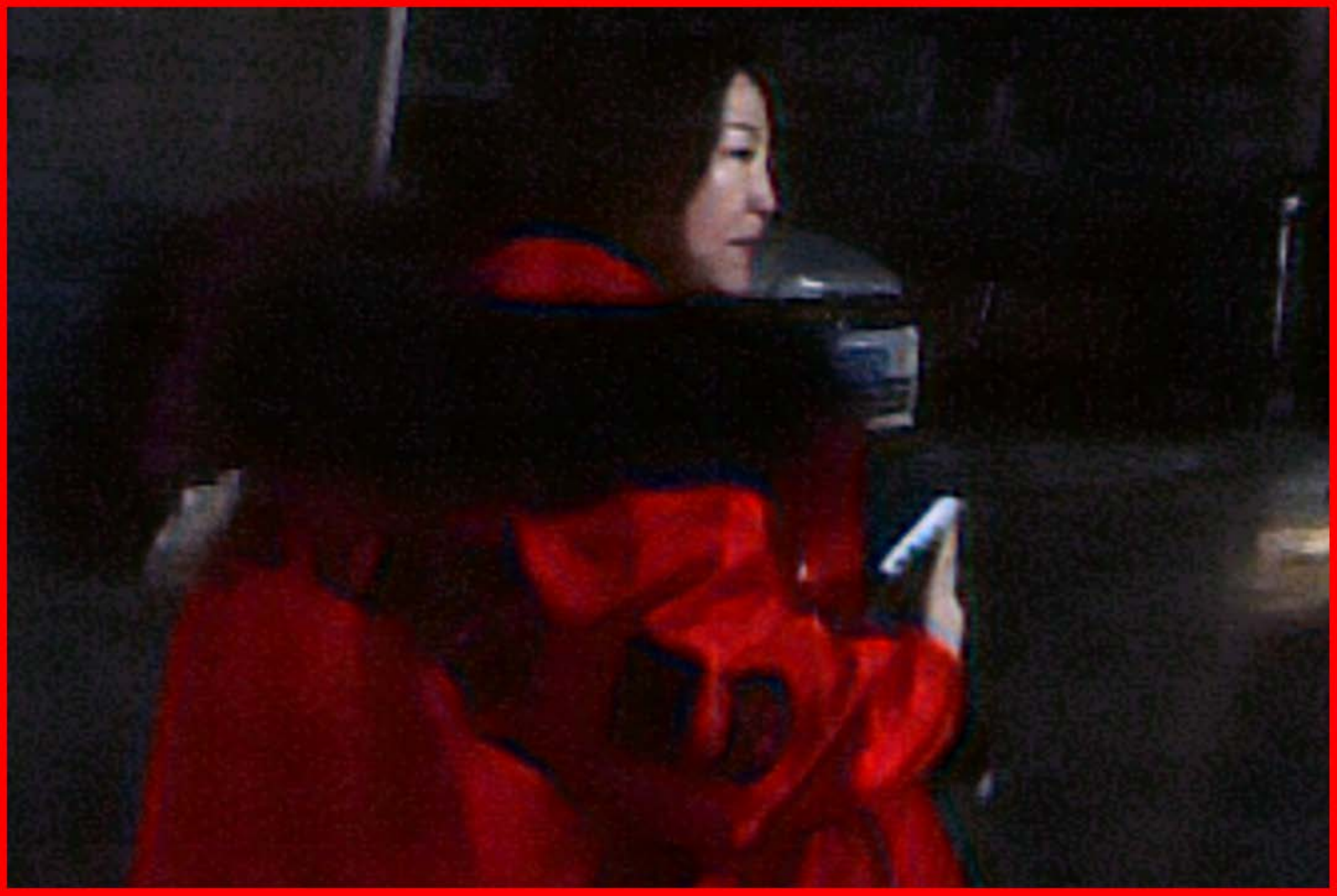}
				\centering \footnotesize Ours\\
			\end{minipage}
		}	
	\end{minipage}
	\begin{minipage}{0.3\textwidth}
		\vspace{0.05cm}
		\subfigure{
			\begin{minipage}{1\textwidth}
				\includegraphics[width=1\textwidth]{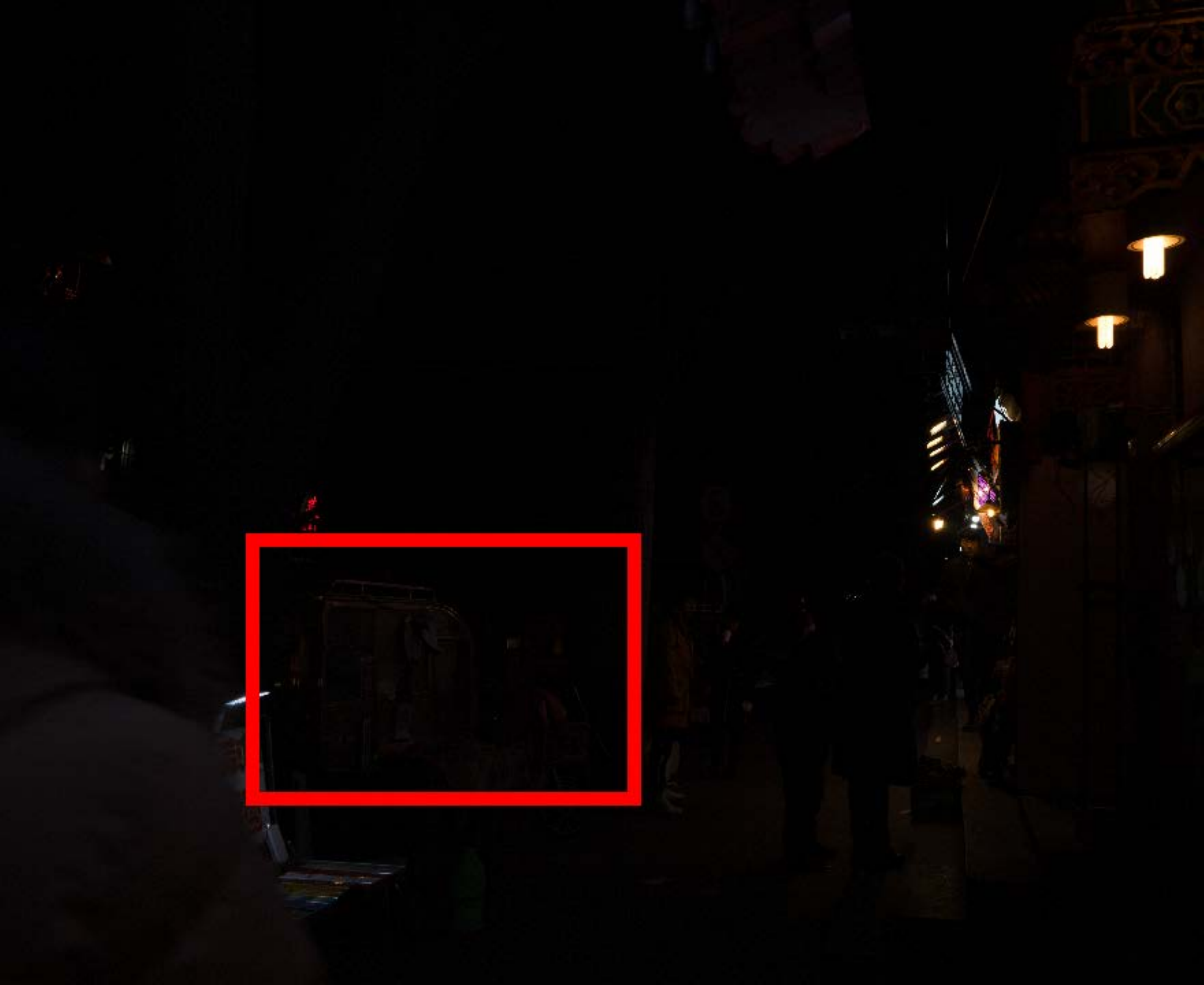}
				\centering  \footnotesize Input\\
			\end{minipage}
		}
	\end{minipage}
	\begin{minipage}{0.16\textwidth}
		\subfigure{
			\begin{minipage}{1\textwidth}
				\includegraphics[width=1\textwidth]{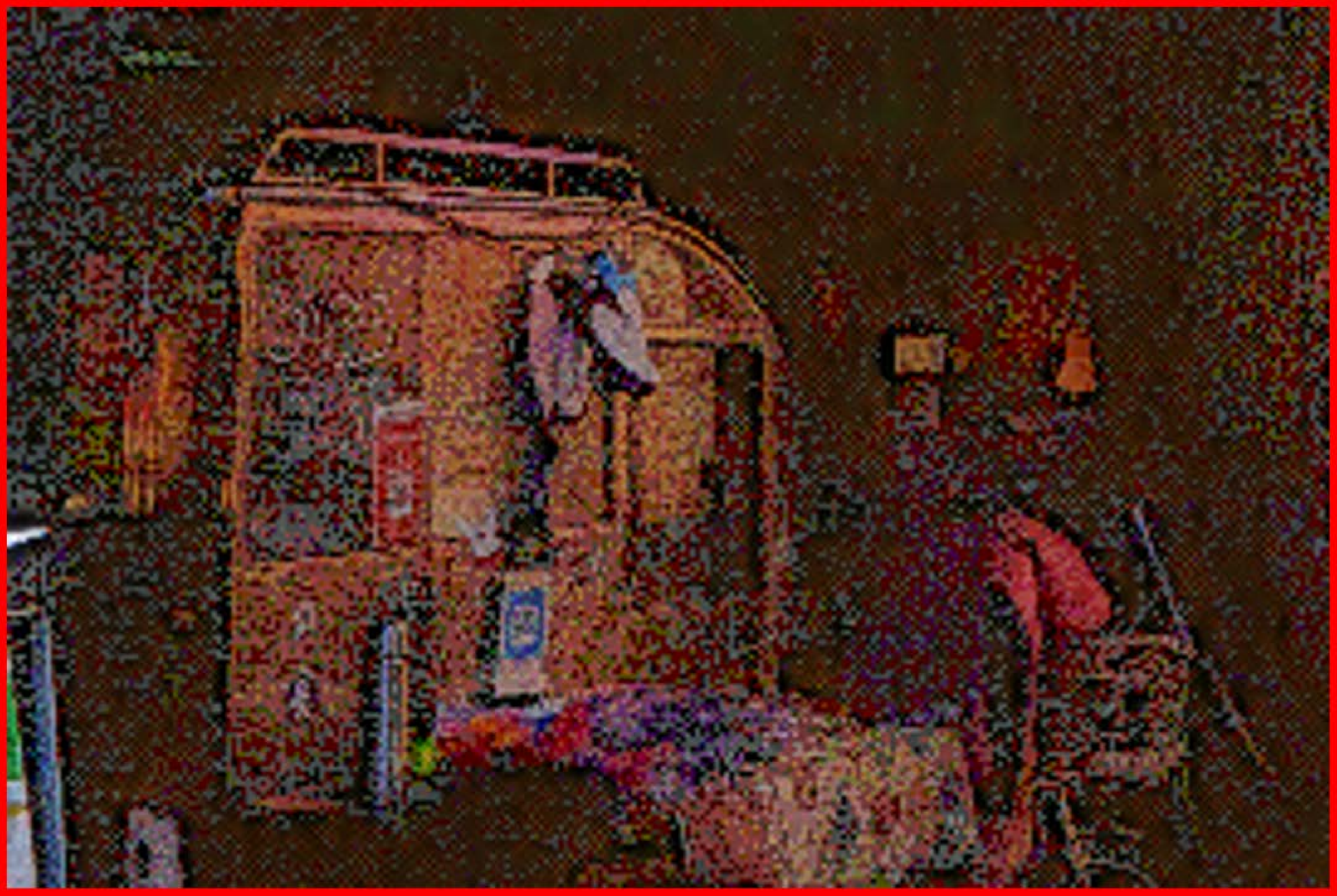}
				\centering \footnotesize RetinexNet \vspace{-0.7em}\\
			\end{minipage}
		}
		\subfigure{
			\begin{minipage}{1\textwidth}
				\includegraphics[width=1\textwidth]{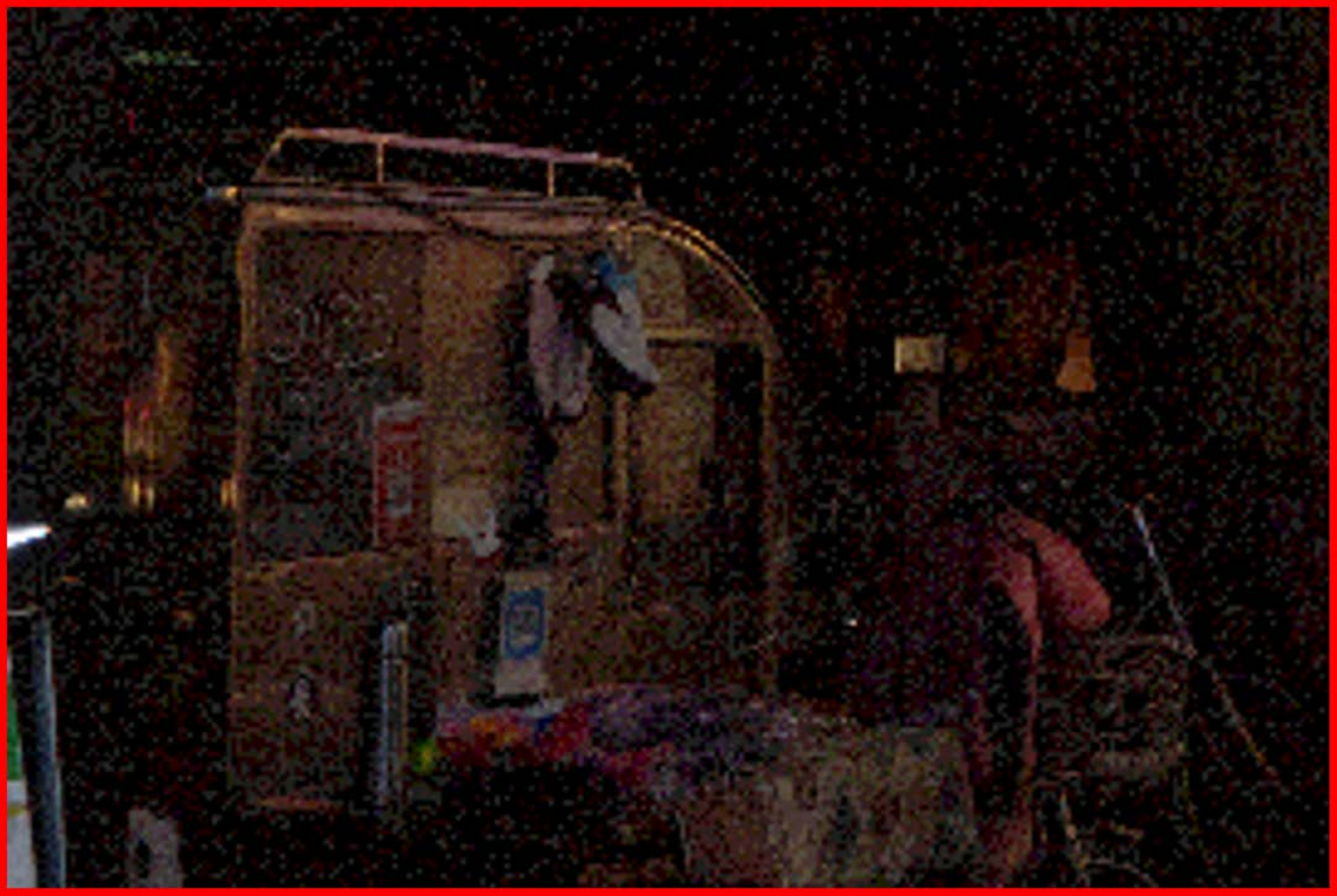}
				\centering \footnotesize ZeroDCE\\
			\end{minipage}
		}
	\end{minipage}
	\begin{minipage}{0.16\textwidth}
		\subfigure{
			\begin{minipage}{1\textwidth}
				\includegraphics[width=1\textwidth]{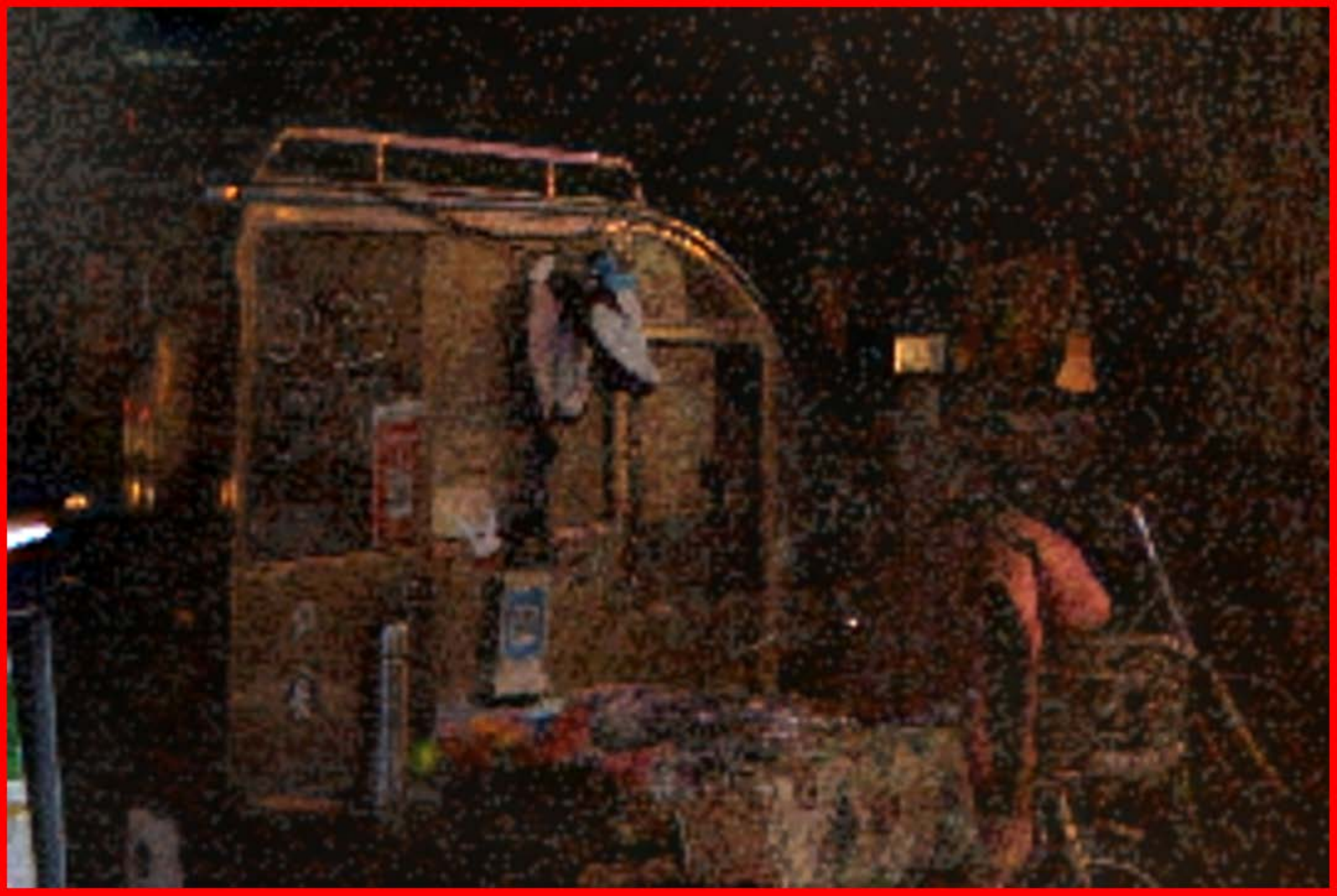}
				\centering \footnotesize  EnGAN \vspace{-0.7em}\\
			\end{minipage}
		}
		\subfigure{
			\begin{minipage}{1\textwidth}
				\includegraphics[width=1\textwidth]{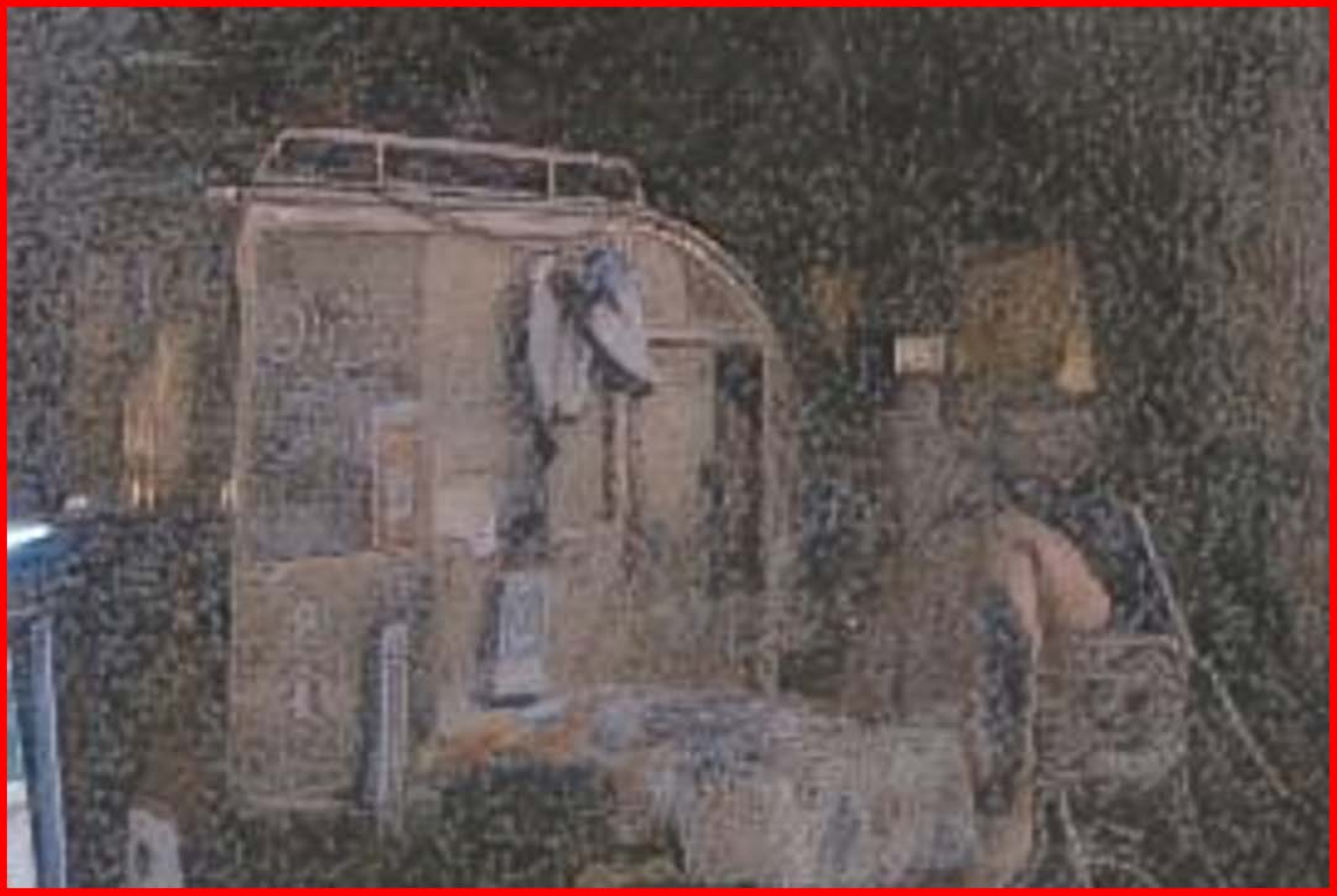}
				\centering \footnotesize  FIDE\\
			\end{minipage}
		}
	\end{minipage}
	\begin{minipage}{0.16\textwidth}
		\subfigure{
			\begin{minipage}{1\textwidth}
				\includegraphics[width=1\textwidth]{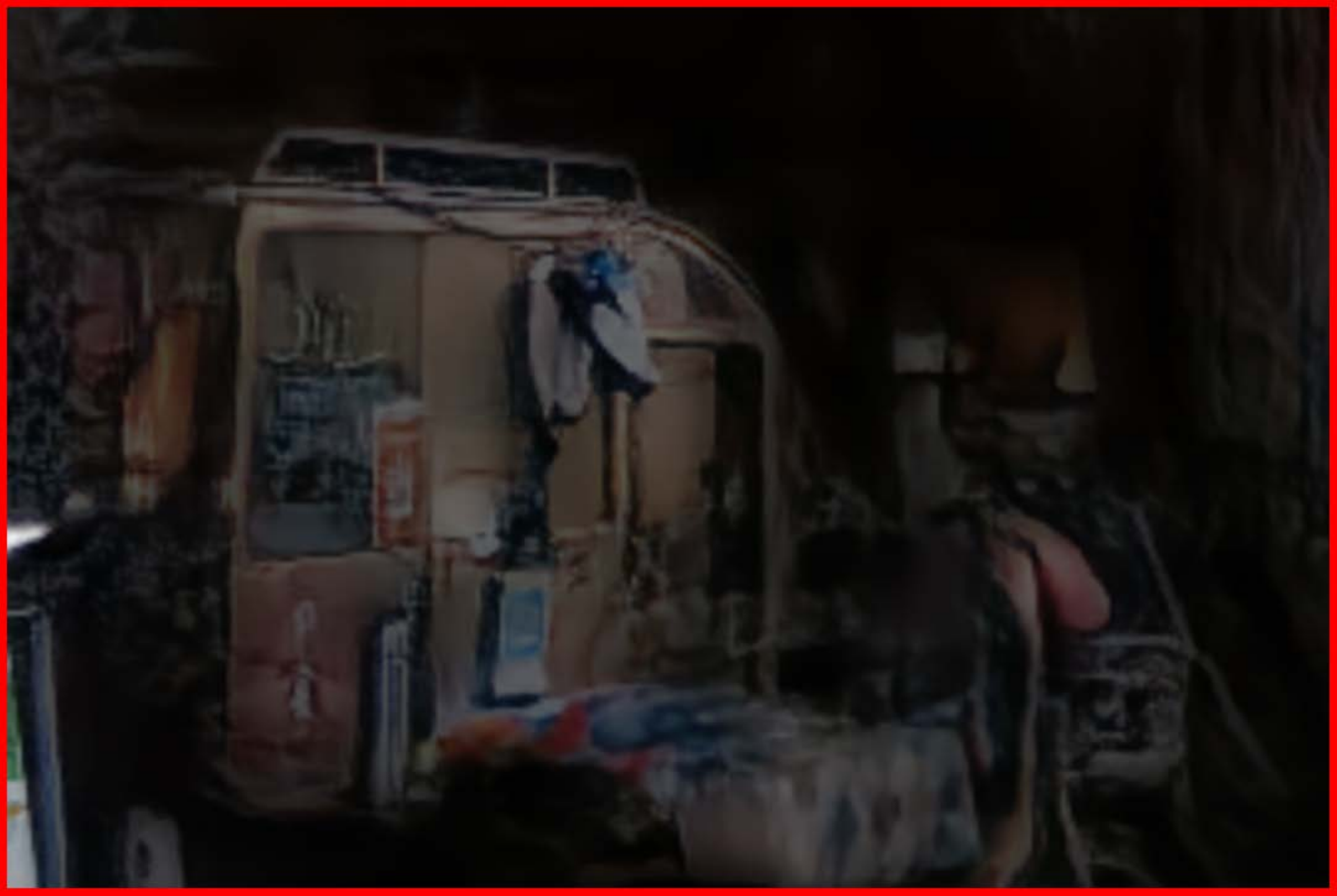}
				\centering \footnotesize KinD \vspace{-0.7em}\\
			\end{minipage}
		}
		\subfigure{
			\begin{minipage}{1\textwidth}
				\includegraphics[width=1\textwidth]{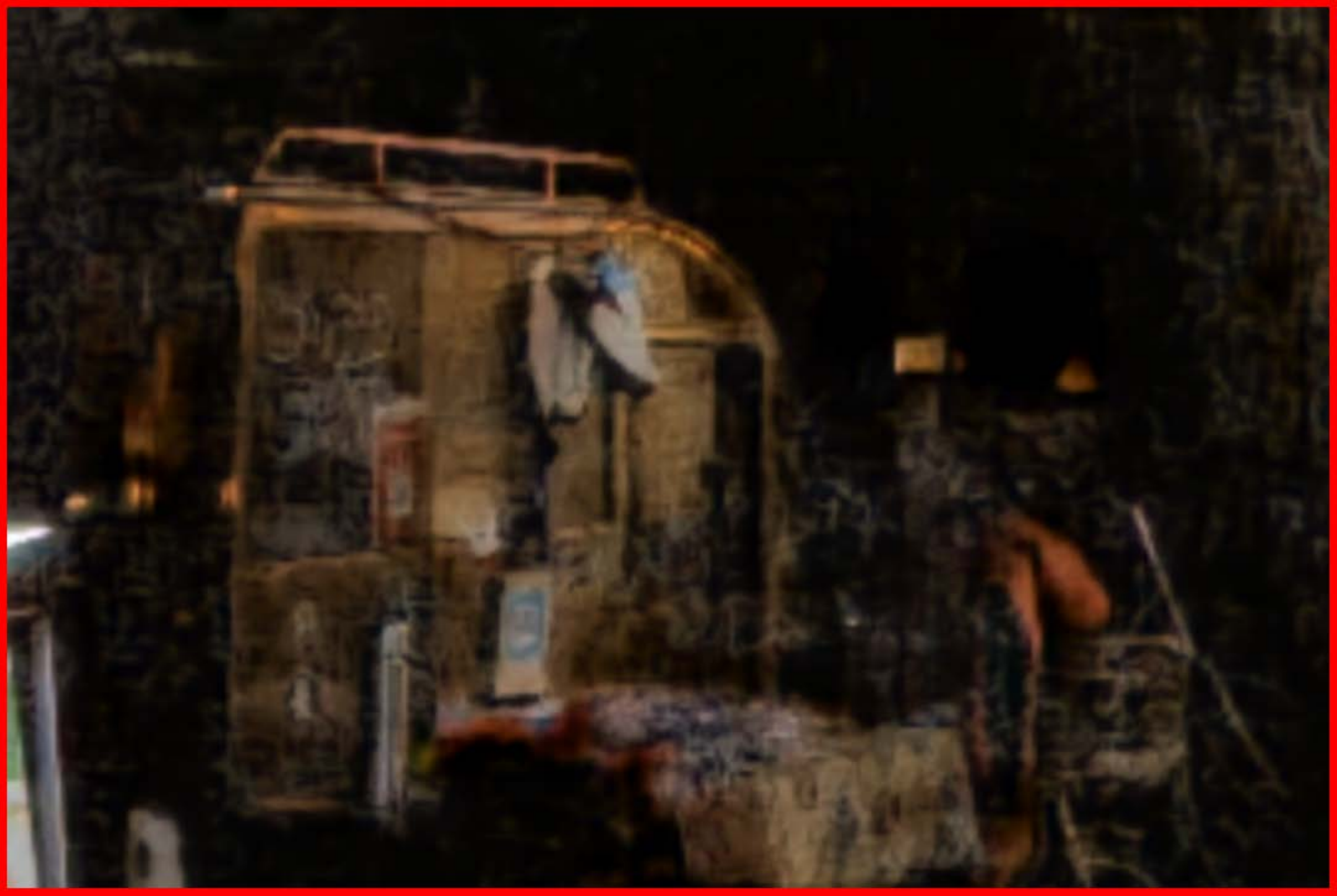}
				\centering \footnotesize DRBN\\
			\end{minipage}
		}
	\end{minipage}
	\begin{minipage}{0.16\textwidth}
		\subfigure{
			\begin{minipage}{1\textwidth}
				\includegraphics[width=1\textwidth]{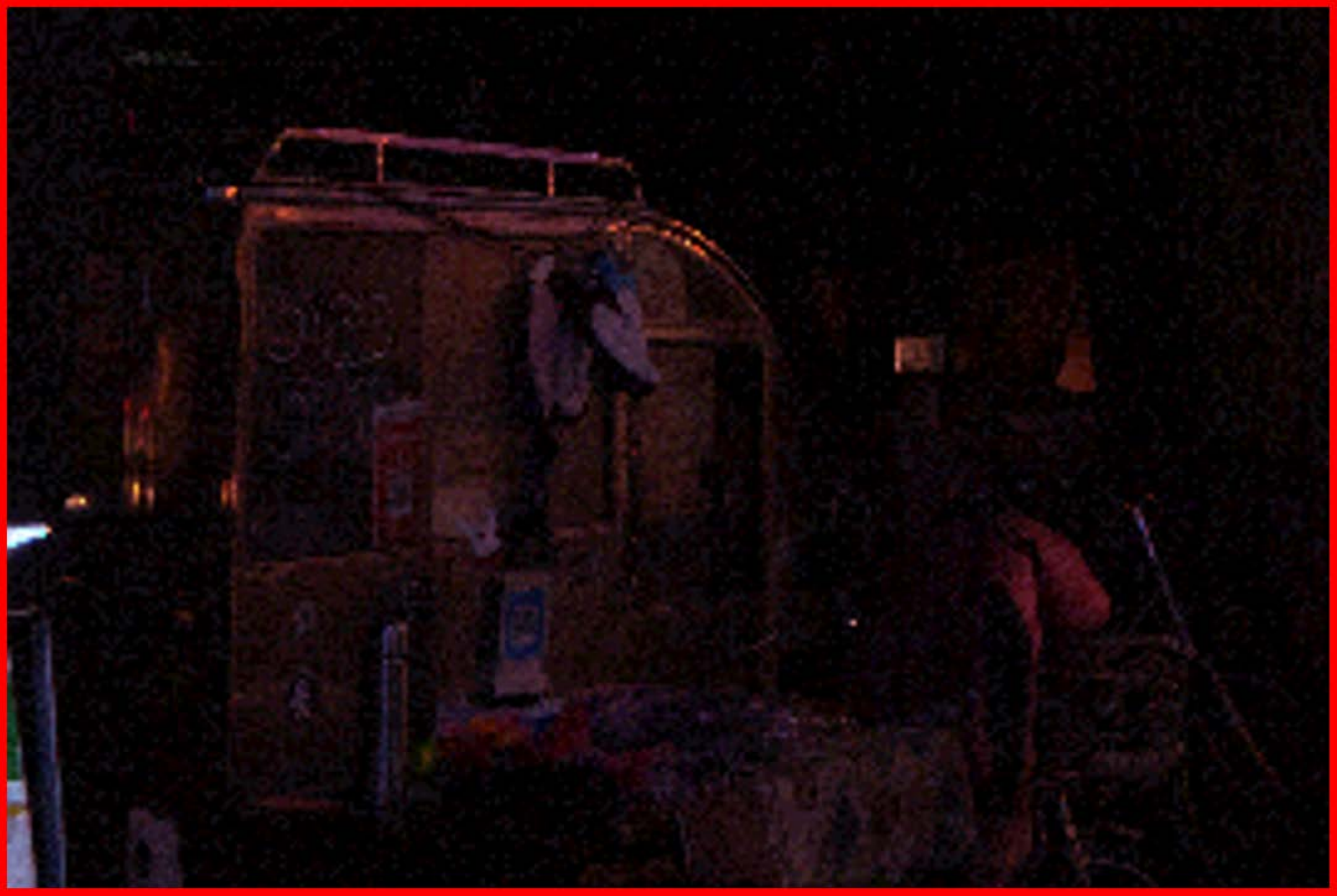}
				\centering \footnotesize DeepUPE \vspace{-0.9em}\\
			\end{minipage}
		}
		\subfigure{
			\begin{minipage}{1\textwidth}
				\includegraphics[width=1\textwidth]{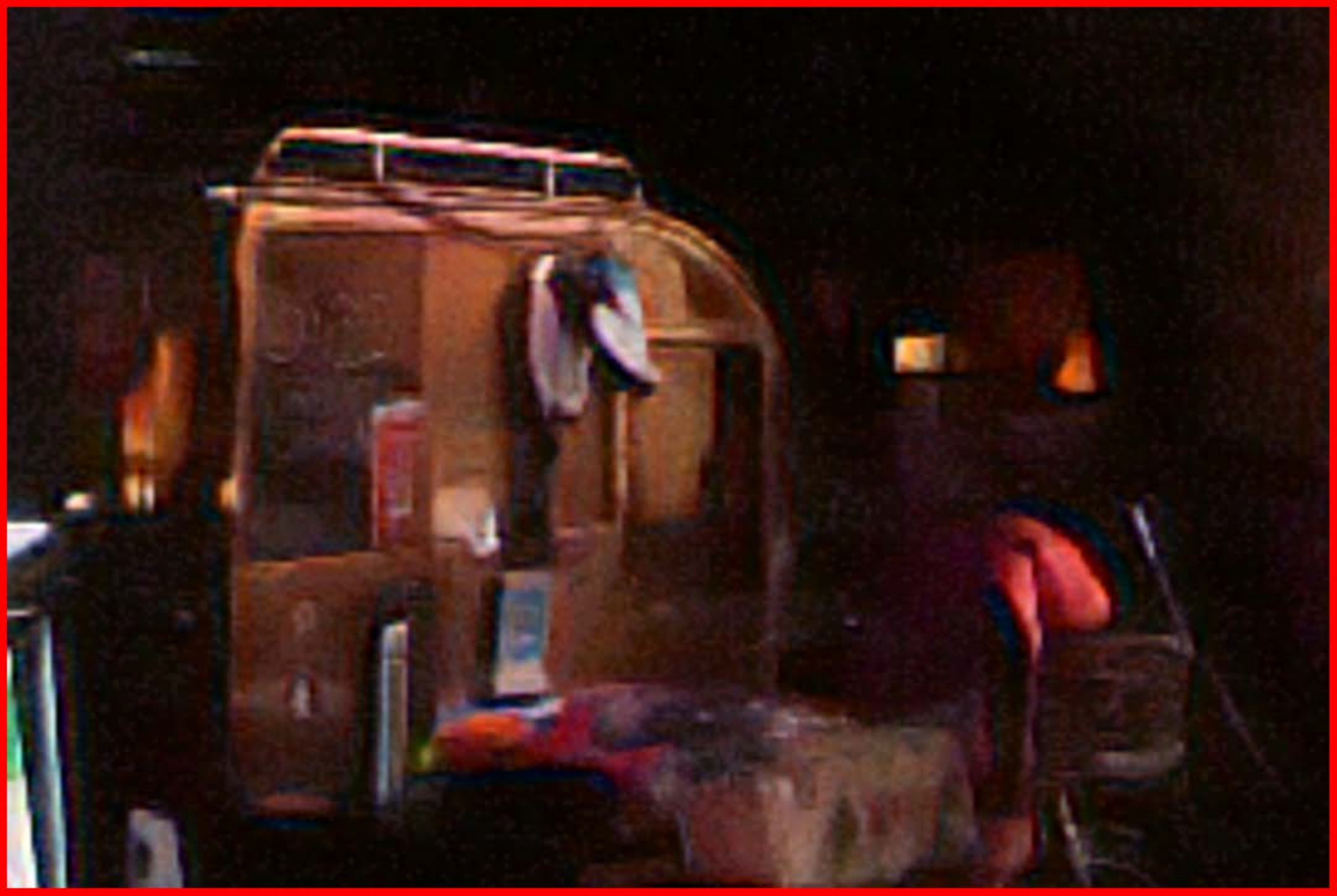}
				\centering \footnotesize Ours\\
			\end{minipage}
		}
	\end{minipage}
	\caption{Visual results of state-of-the-art methods and ours on the ExtremelyDarkFace dataset. Red boxes indicate the obvious differences.}
	\label{fig:ExtermelyDarkFace}
\end{figure*}

\begin{table*}[t]
	\caption{The model size, FLOPs and running time (GPU-seconds for inference) of some recently proposed CNN-based methods (with manually designed architectures) and our searched RUAS$_\mathtt{i}$ and RUAS$_{\mathtt{i}+\mathtt{n}}$. The FLOPs and running time are reported on the LOL dataset. The best result is in red whereas the second best one is in blue.}
	\begin{center}
		\begin{tabular}{|l|cccccccccc|}
			\hline
			\footnotesize {Methods}&\footnotesize MBLLEN&\footnotesize GLADNet &\footnotesize RetinexNet &\footnotesize EnGAN&\footnotesize SSIENet&\footnotesize KinD&\footnotesize FIDE&\footnotesize DRBN&\footnotesize RUAS$_\mathtt{i}$&\footnotesize RUAS$_{\mathtt{i}+\mathtt{n}}$\\
			\hline
			\footnotesize SIZE (M)&\footnotesize 0.450&\footnotesize 1.128&\footnotesize 0.838&\footnotesize 8.636&\footnotesize 0.682&\footnotesize 8.540&\footnotesize 8.621&\footnotesize 0.577&\footnotesize \textcolor{red}{\textbf{0.001}}&\footnotesize \textcolor{blue}{\textbf{0.003}}\\
			\hline
			\footnotesize FLOPs (G)&\footnotesize 19.956&\footnotesize 252.141&\footnotesize 136.015&\footnotesize 61.010&\footnotesize 34.607&\footnotesize 29.130&\footnotesize 57.240&\footnotesize 37.790&\footnotesize \textcolor{red}{\textbf{0.281}}&\footnotesize \textcolor{blue}{\textbf{0.832}}\\
			\hline
			\footnotesize TIME (S)&\footnotesize 0.077&\footnotesize 0.025&\footnotesize 0.119&\footnotesize \textcolor{blue}{\textbf{0.010}}&\footnotesize 0.027&\footnotesize 0.181&\footnotesize 0.594&\footnotesize 0.053&\footnotesize \textcolor{red}{\textbf{0.006}}&\footnotesize 0.016\\
			\hline
		\end{tabular}
	\end{center}
	\vspace{-0.25cm}
	\label{tab: parameters}
\end{table*}

\subsubsection{Reference-free Bilevel Learning }

We first specify our training and validation objectives based on a series of reference-free losses. Specifically, for IEM, we define a loss $\frac{1}{2}\|\mathtt{net}_{\balpha_{\ttt},\bomega_{\ttt}}(\mathbf{y})-\hat{\mathbf{t}}_0)\|^2 +  \eta_{\ttt}\mathtt{RTV}(\mathtt{net}_{\balpha_{\ttt},\bomega_{\ttt}}(\mathbf{y}))$ on the training and validation dataset as  $\mathcal{L}^{\mathtt{t}}_{\mathtt{tr}}$ and $\mathcal{L}^{\mathtt{t}}_{\mathtt{val}}$, respectively. Here the first term is the fidelity and $\mathtt{RTV}(\cdot)$ denotes the relative total variation term~\cite{tsmoothing2012} (with a parameter $\eta_{\ttt}>0$). In fact, this loss encourages IEM to output illuminations that can simultaneously preserve the overall structure and smooth the textural details. As for NRM, we introduce a similar loss $\frac{1}{2}\|\mathtt{net}_{\balpha_{\tn},\bomega_{\tn}}(\mathbf{u}_K)-\mathbf{u}_K\|^2 +  \eta_{\tn}\mathtt{TV}(\mathtt{net}_{\balpha_{\tn},\bomega_{\tn}}(\mathbf{u}_K))$ to define $\mathcal{L}^{\mathtt{n}}_{\mathtt{tr}}$ and $\mathcal{L}^{\mathtt{n}}_{\mathtt{val}}$, in which we utilize standard total variation $\mathtt{TV}(\cdot)$ as our regularization~\cite{osher2005iterative} (with a parameter $\eta_{\tn}>0$).

Then Alg.~\ref{alg:search} summarizes the overall search process\footnote{Here we skip some regular numerical parameters (e.g., initialization, learning rate and stopping criterion) to simplify our notations.}. It can be seen that IEM and NRM are searched alternatively and simultaneously. That is, we update $\balpha_{\ttt}$ (with the current $\balpha_{\tn}$) for IEM (Steps 4-7) and update $\balpha_{\tn}$ (based on the updated $\balpha_{\ttt}$) for NRM (Steps 9-12). As for each module, we just adopt the widely used one-step finite difference technique~\cite{forsythe1960finite} to approximately calculate gradients for the upper-level variables (Steps 5 and 10).

\section{Experimental Results}\label{sec: experiments}

\subsection{Implementation Details}
We sampled 500 underexposure images from MIT-Adobe 5K~\cite{fivek} for searching and training, and sampled 100 image pairs for testing. 
For LOL Dataset~\cite{Chen2018Retinex}, 100 image pairs were randomly sampled for evaluating and the remaining 689 low-light images are used for searching and training. 
We adopted the well-known PSNR and SSIM as our evaluated metrics.
We evaluated the visual performance in the DarkFace~\cite{yang2020advancing} and ExtremelyDarkFace (used as the sub-challenge in the CVPR 2020 UG2+Challenge\footnote{\url{http://cvpr2020.ug2challenge.org/dataset20_t1.html}}) datasets. We conducted all experiments on a PC with a single TITAN X GPU and Intel Core i7-7700 3.60GHz CPU.

In the prior architecture search phase, we consider the same search space (with 3 fundamental cell structures) for IEM and NRM, but define their cells with different channel widths (i.e., 3 for IEM and 6 for NRM). The gradients of the architecture and weight parameters are computed following standard differential NAS techniques~\cite{liu2018darts}. As for the numerical parameters, we set the maximum epoch as 20, the batch size as 1, and chose the initial learning rate as $3\times10^{-4}$. The momentum parameter was randomly sampled from (0.5, 0.999) and the weight decay was set as $10^{-3}$. As for the training phase (with searched architecture), we use the same training losses as that in the search phase.

\subsection{Comparison with State-of-the-arts}

We compared it qualitatively and quantitatively with twelve recently-proposed state-of-the-art methods including LIME~\cite{guo2017lime}, SDD~\cite{hao2020low}, MBLLEN~\cite{lv2019attention}, GLADNet~\cite{wang2018gladnet}, RetinexNet~\cite{Chen2018Retinex}, EnGAN~\cite{jiang2019enlightengan}, SSIENet~\cite{zhang2020self}, KinD~\cite{zhang2019kindling}, DeepUPE~\cite{wang2019underexposed}, ZeroDCE~\cite{guo2020zero}, FIDE~\cite{xu2020learning}, and DRBN~\cite{yang2020fidelity}.


\begin{figure*}[t]
	\begin{center}
		\begin{tabular}{c@{\extracolsep{0.57em}}c@{\extracolsep{0.25em}}c@{\extracolsep{0.57em}}c@{\extracolsep{0.001em}}c@{\extracolsep{0.31em}}c@{\extracolsep{0.25em}}c}
			\includegraphics[width=0.133\linewidth]{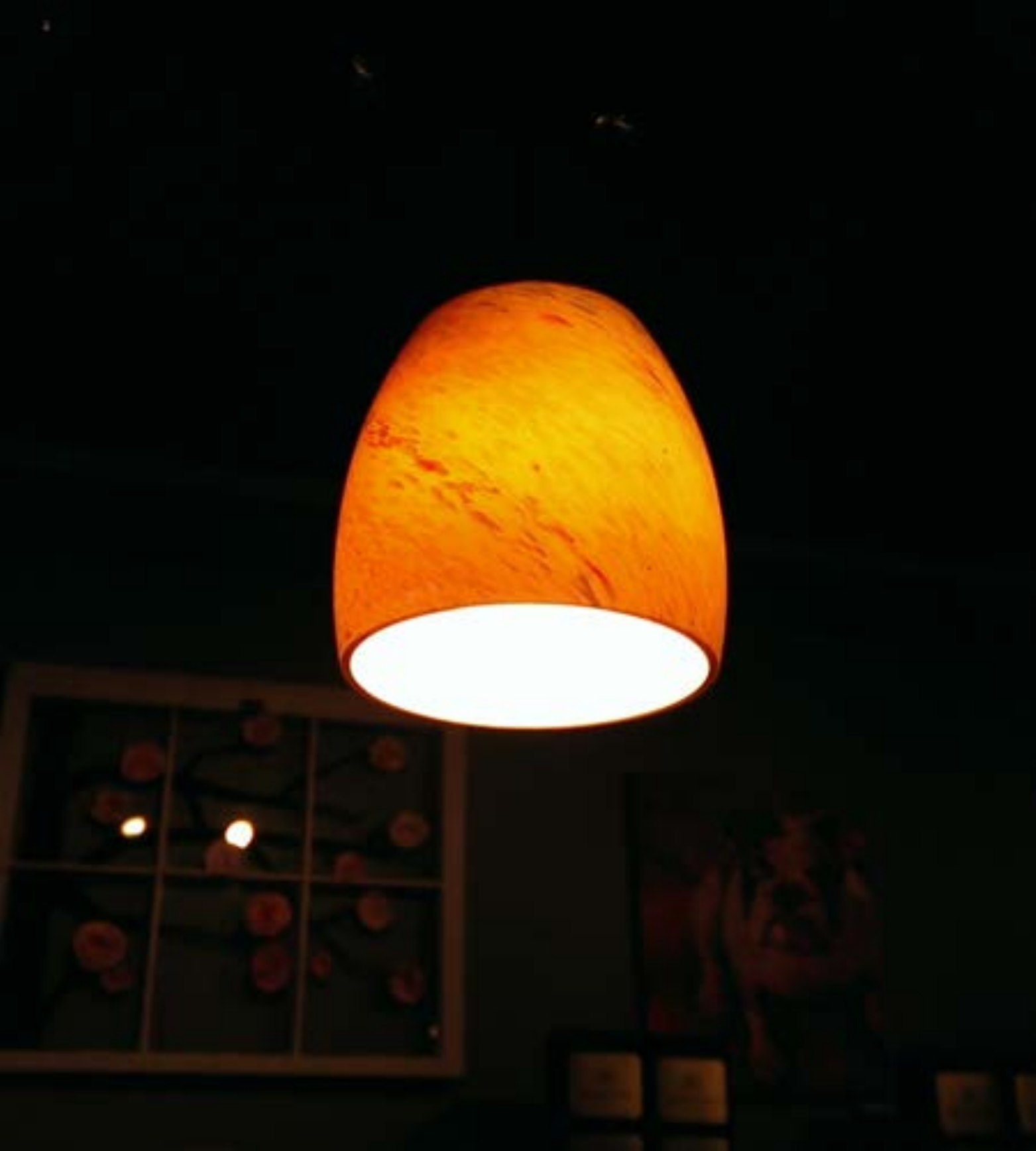}&
			\includegraphics[width=0.133\linewidth]{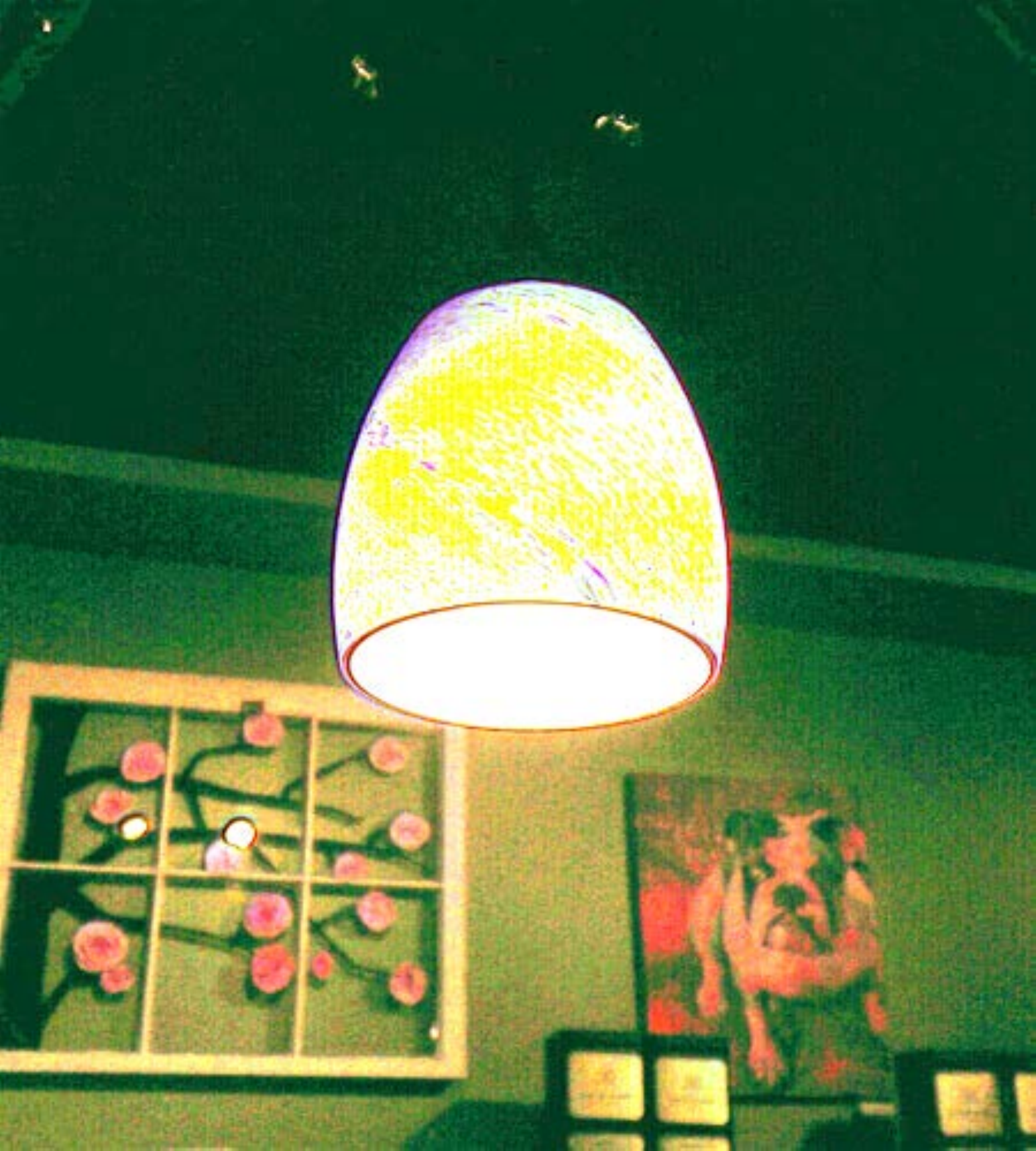}&
			\includegraphics[width=0.133\linewidth]{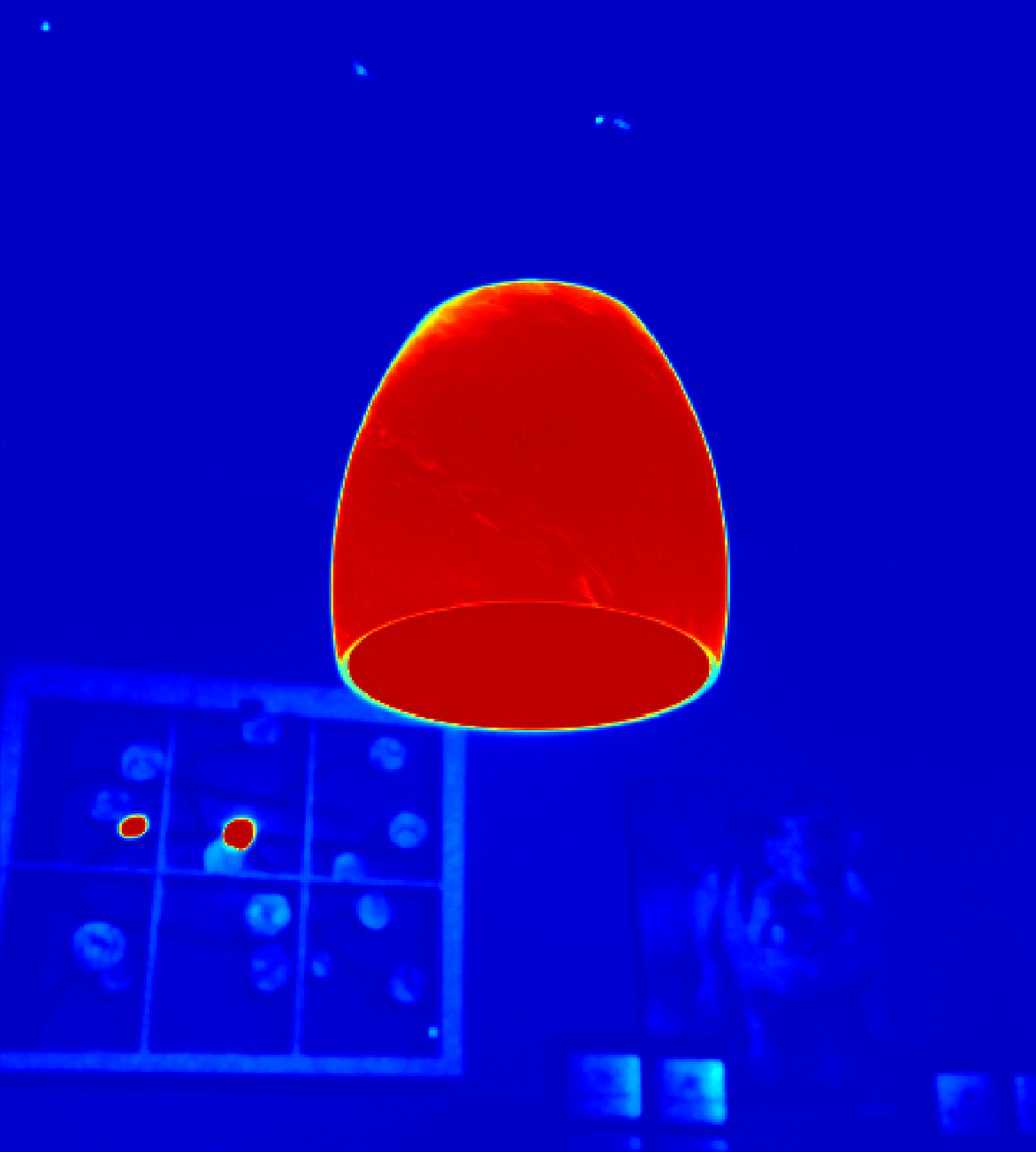}&
			\includegraphics[width=0.133\linewidth]{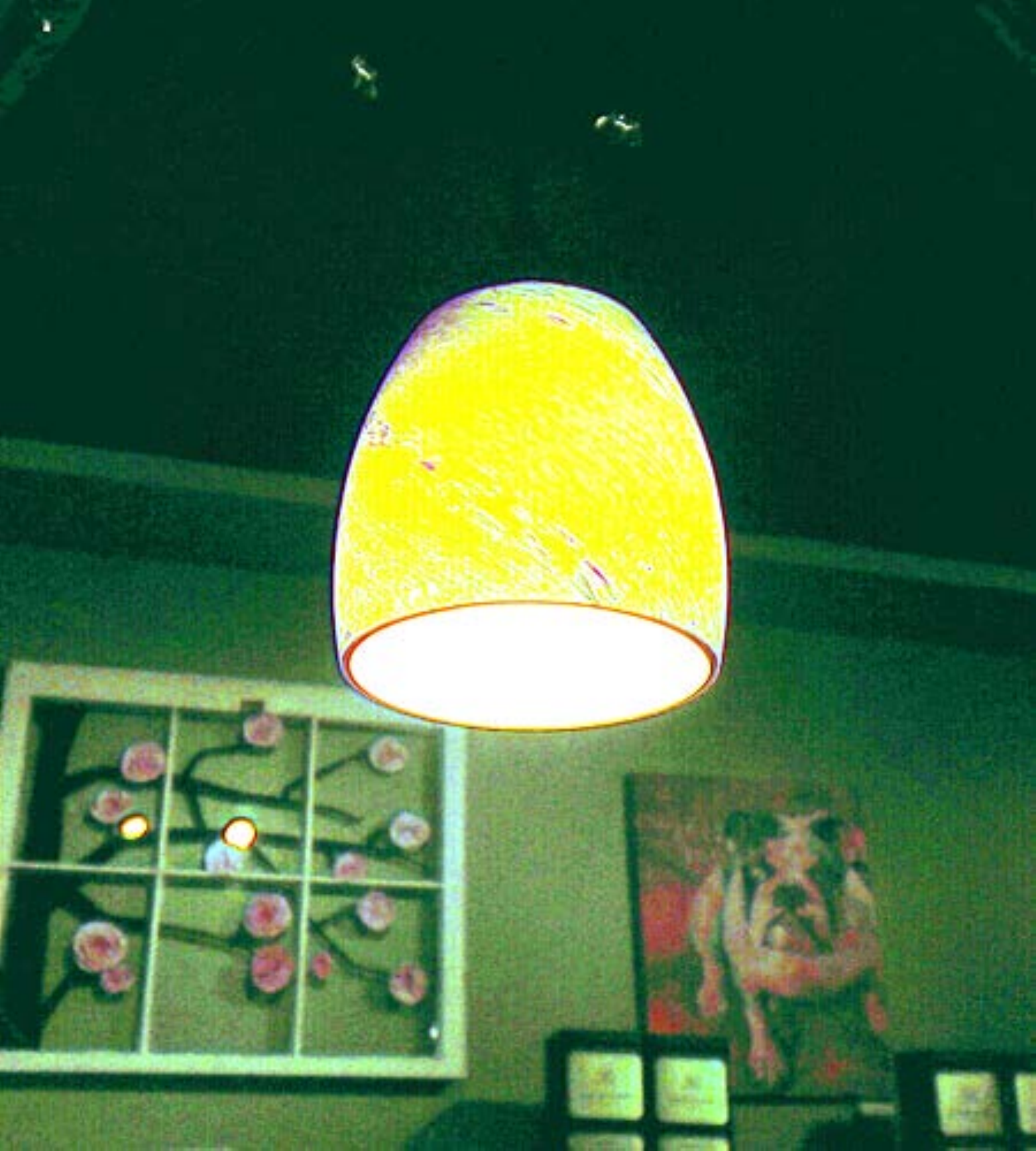}&
			\includegraphics[width=0.133\linewidth]{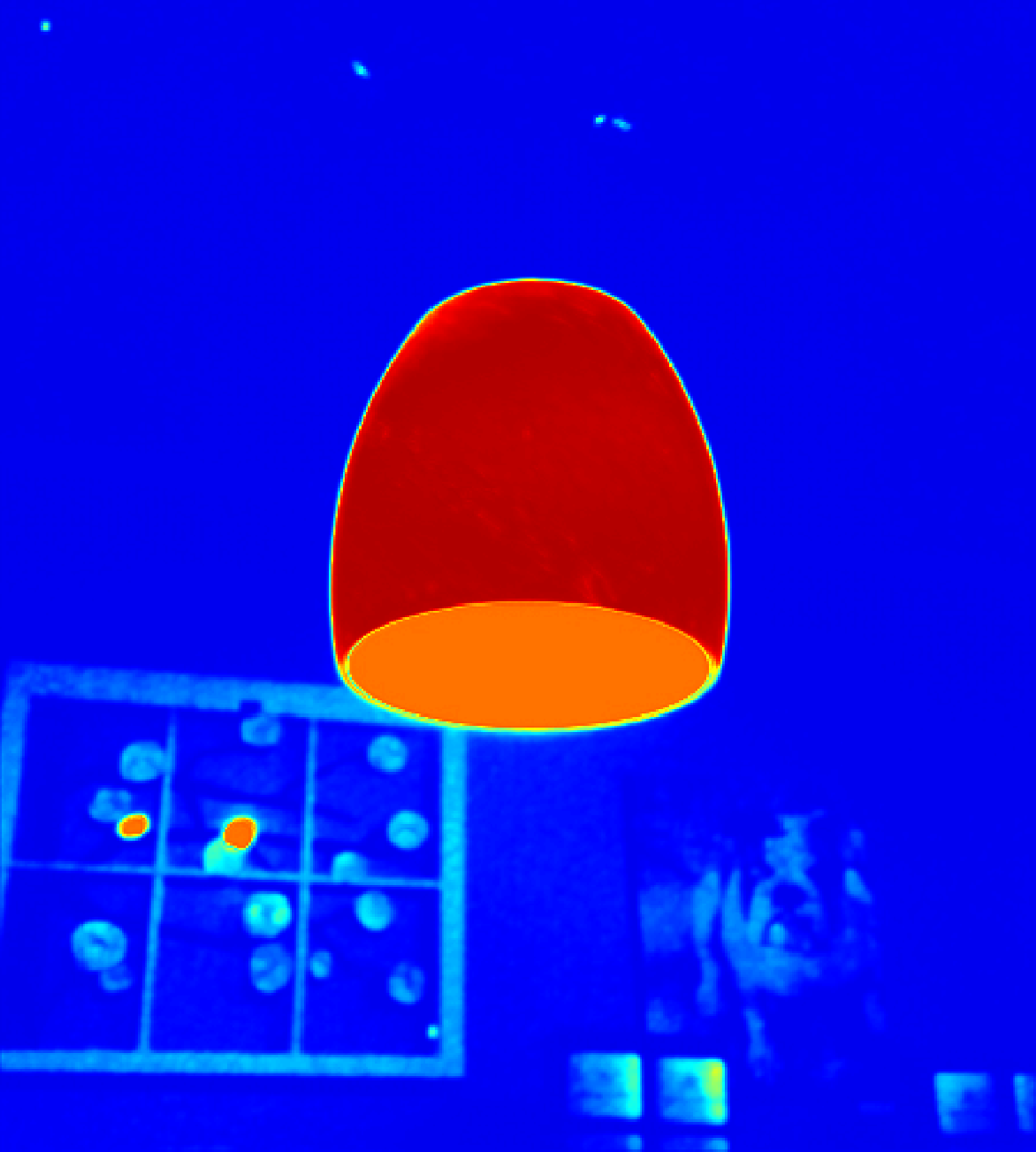}&
			\includegraphics[width=0.133\linewidth]{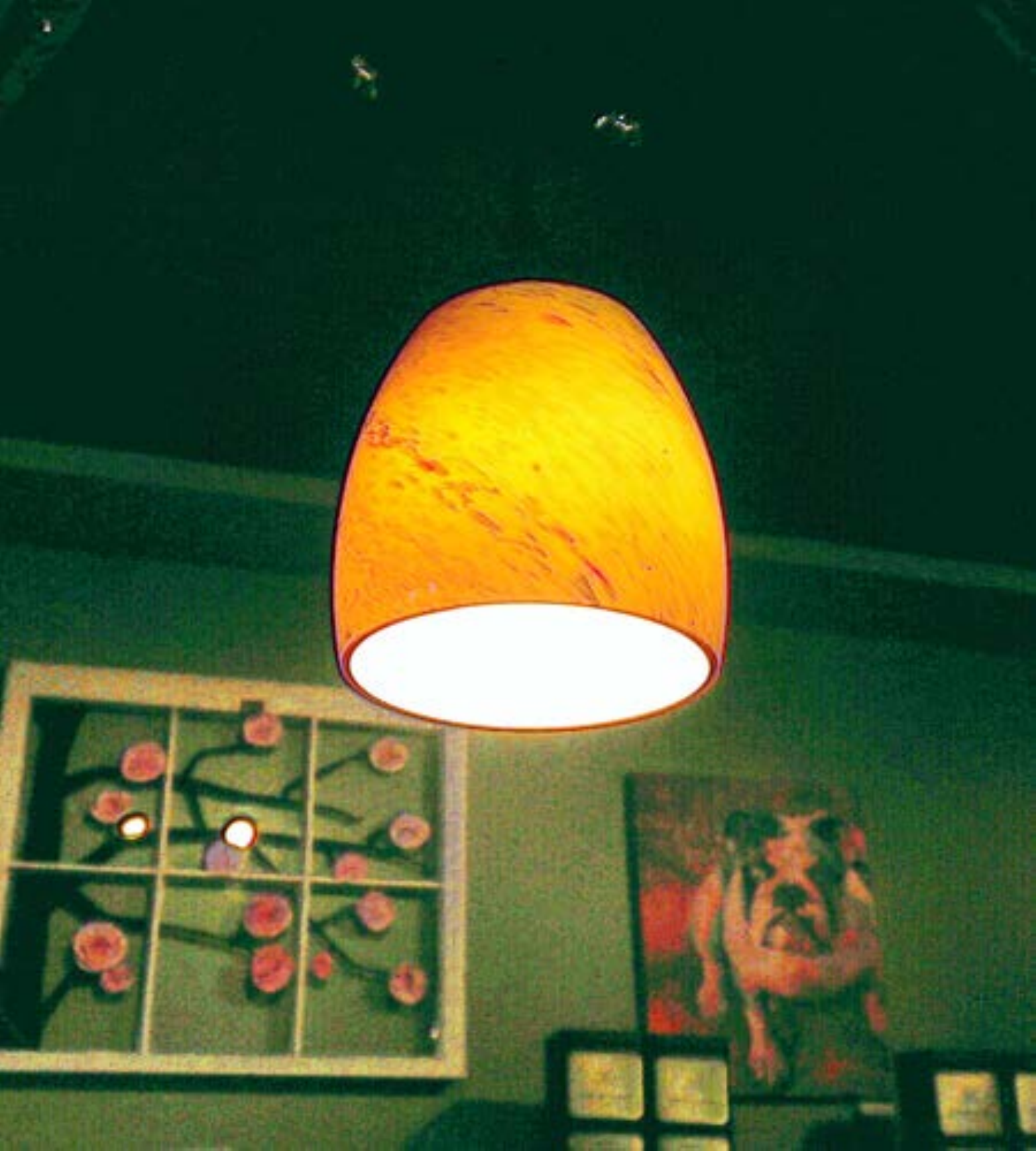}&
			\includegraphics[width=0.133\linewidth]{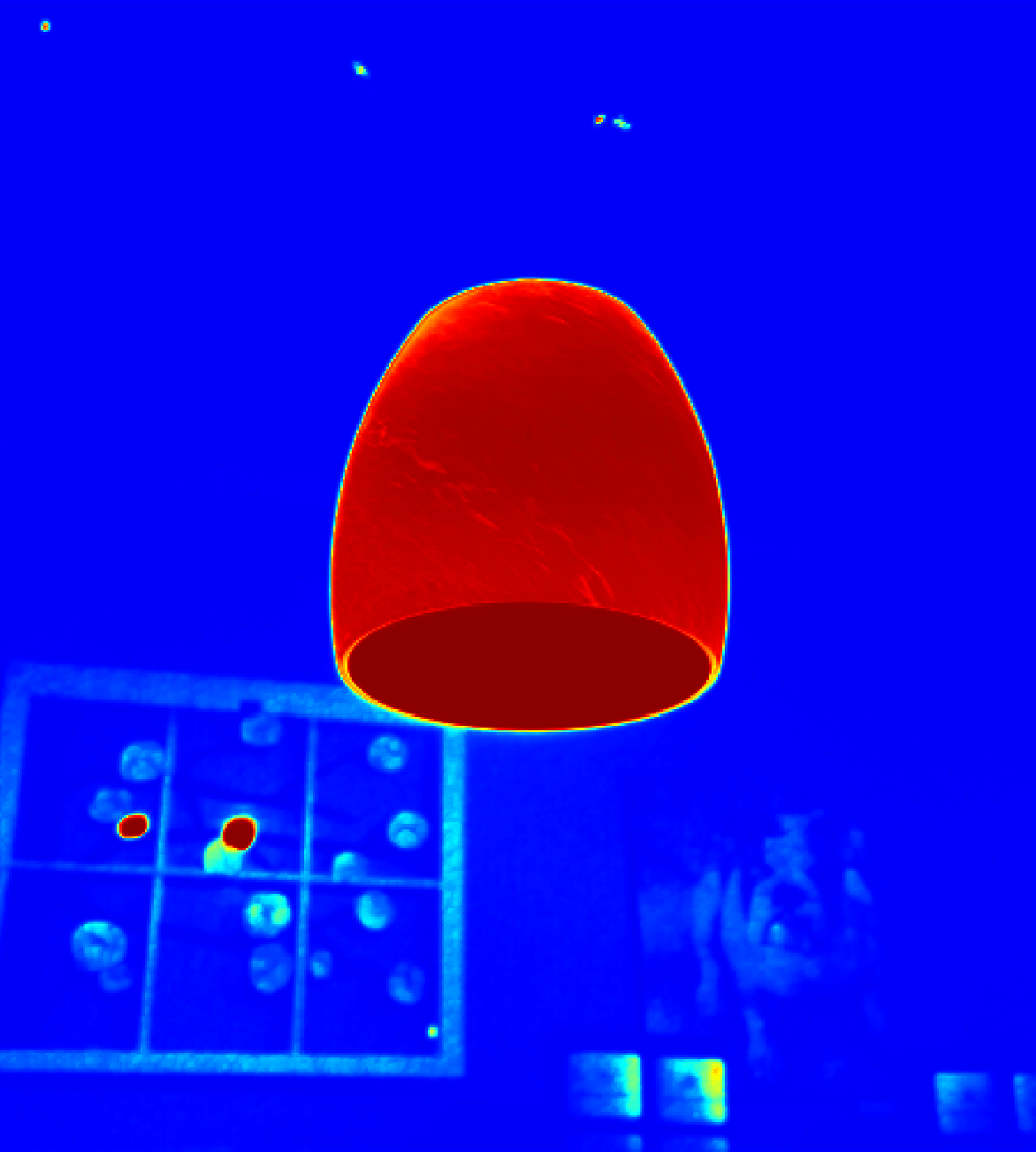}\\
			\footnotesize (a) Input&\multicolumn{2}{c}{\footnotesize (b) Fix warm-start as $\hat{\mathbf{t}}_0$}&\multicolumn{2}{c}{\footnotesize (c) Update $\hat{\mathbf{t}}_k$ w/o residual rectification}&\multicolumn{2}{c}{\footnotesize (d) Update $\hat{\mathbf{t}}_k$ w/ residual rectification}\\
		\end{tabular}
	\end{center}
	\vspace{-0.25cm}
	\caption{Ablation study of the effect of different warm-start strategies. Subfigures (b)-(d) plot the enhanced results and corresponding estimated illumination maps  for different settings. }
	\label{fig: teffect}
\end{figure*}

Firstly, we evaluated the RUAS in some simple real-world scenarios. We reported the quantitative scores on the MIT-Adobe 5K dataset. As shown in the first two rows in Table~\ref{tab: MITquantitative}, we could easily see that our method obtained the best scores. Limited to space, we provided the visual comparisons on this benchmark in the Supplemental Material. 
Then we evaluated the visual performance in some real-world scenarios. Fig.~\ref{fig:DarkFace} demonstrated three groups of visual comparisons on DarkFace dataset~\cite{yang2020advancing}. 
From the results, although some methods were able to enhance the brightness successfully, they failed to restore the textures clearly.
On the contrary, our RUAS can restore the brightness and the details perfectly at the same time.

\begin{figure}[t]
	\begin{center}
		\begin{tabular}{c@{\extracolsep{0.3em}}c@{\extracolsep{0.3em}}c}
			\includegraphics[width=0.31\linewidth]{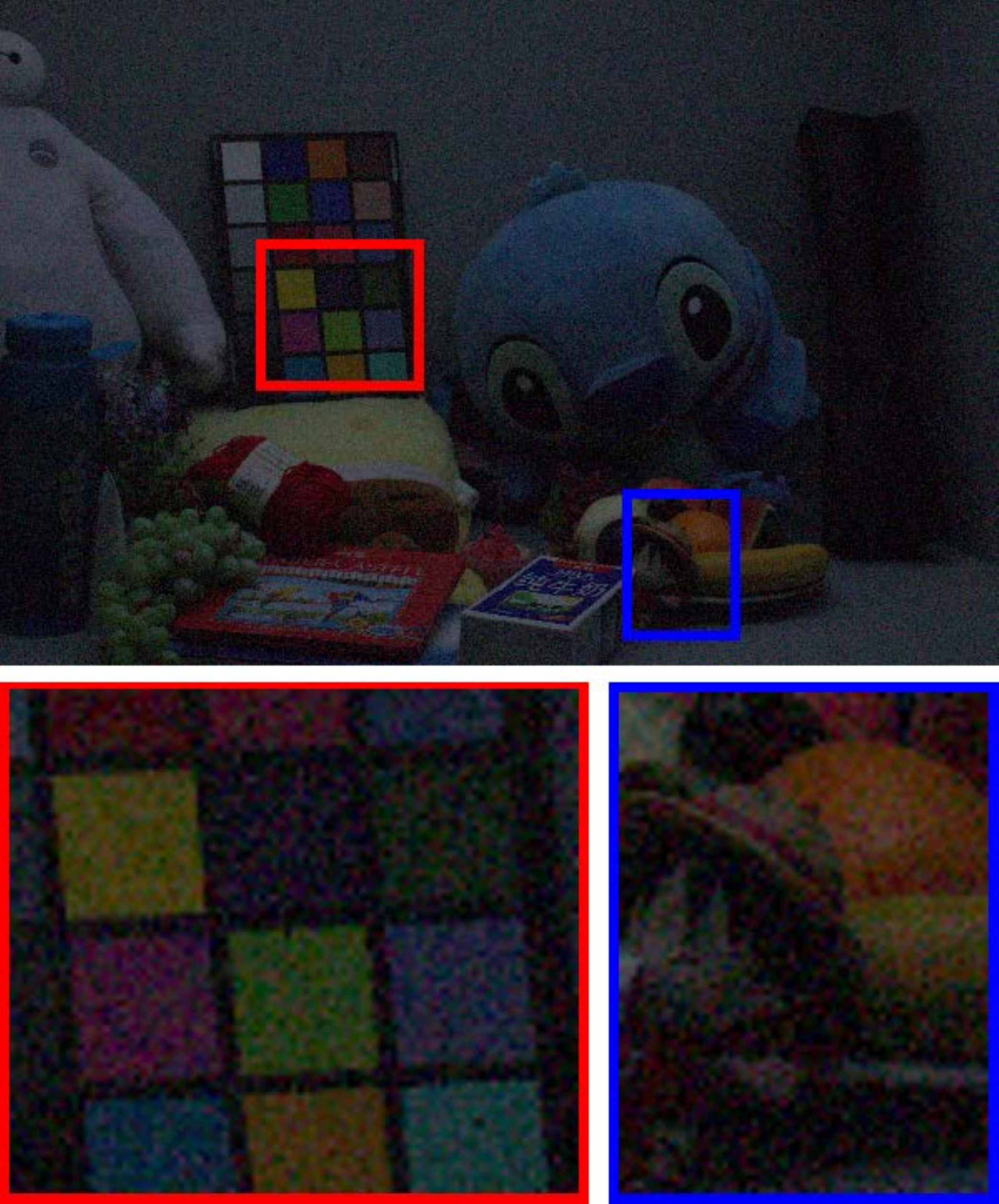}&
			\includegraphics[width=0.31\linewidth]{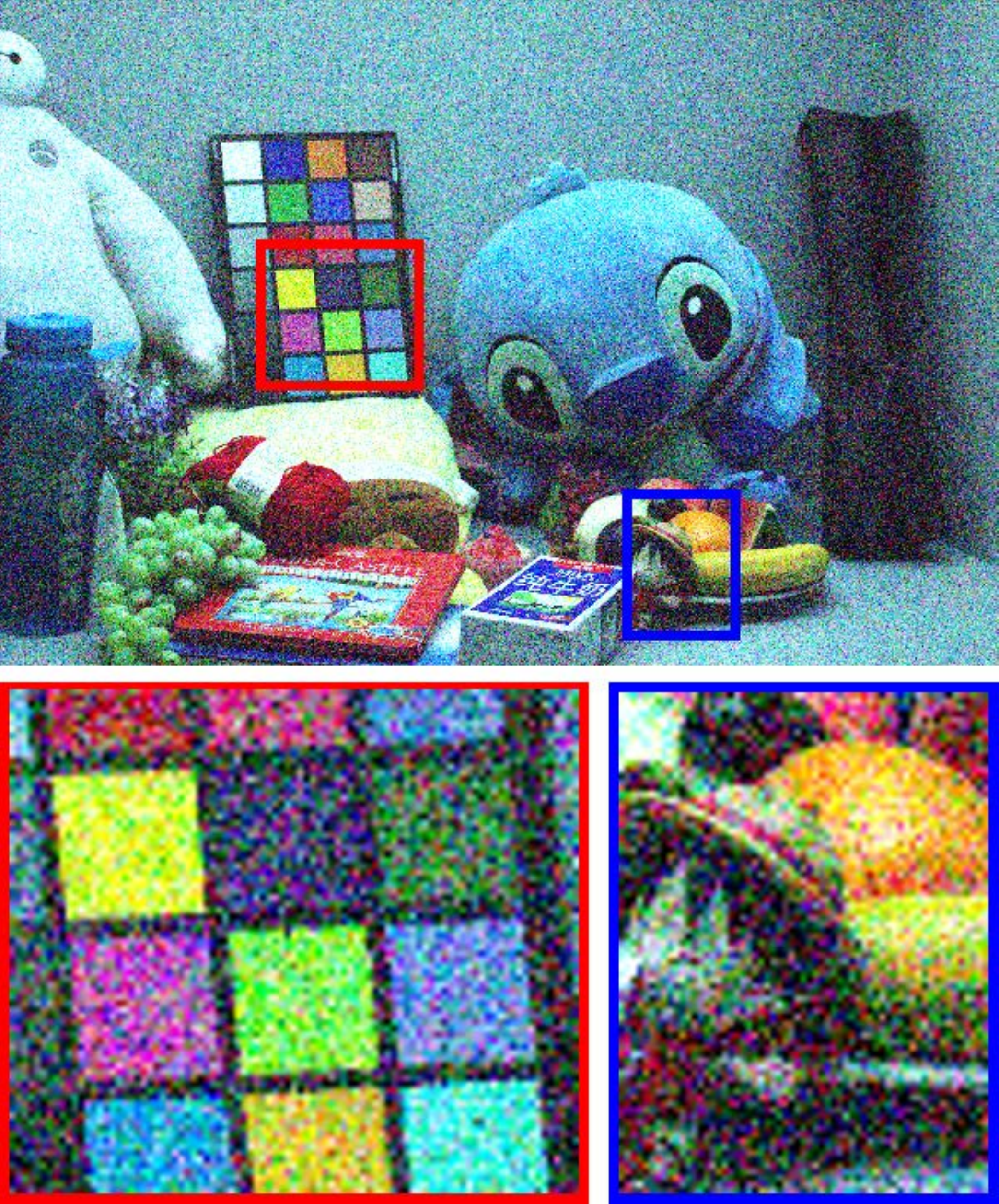}&
			\includegraphics[width=0.31\linewidth]{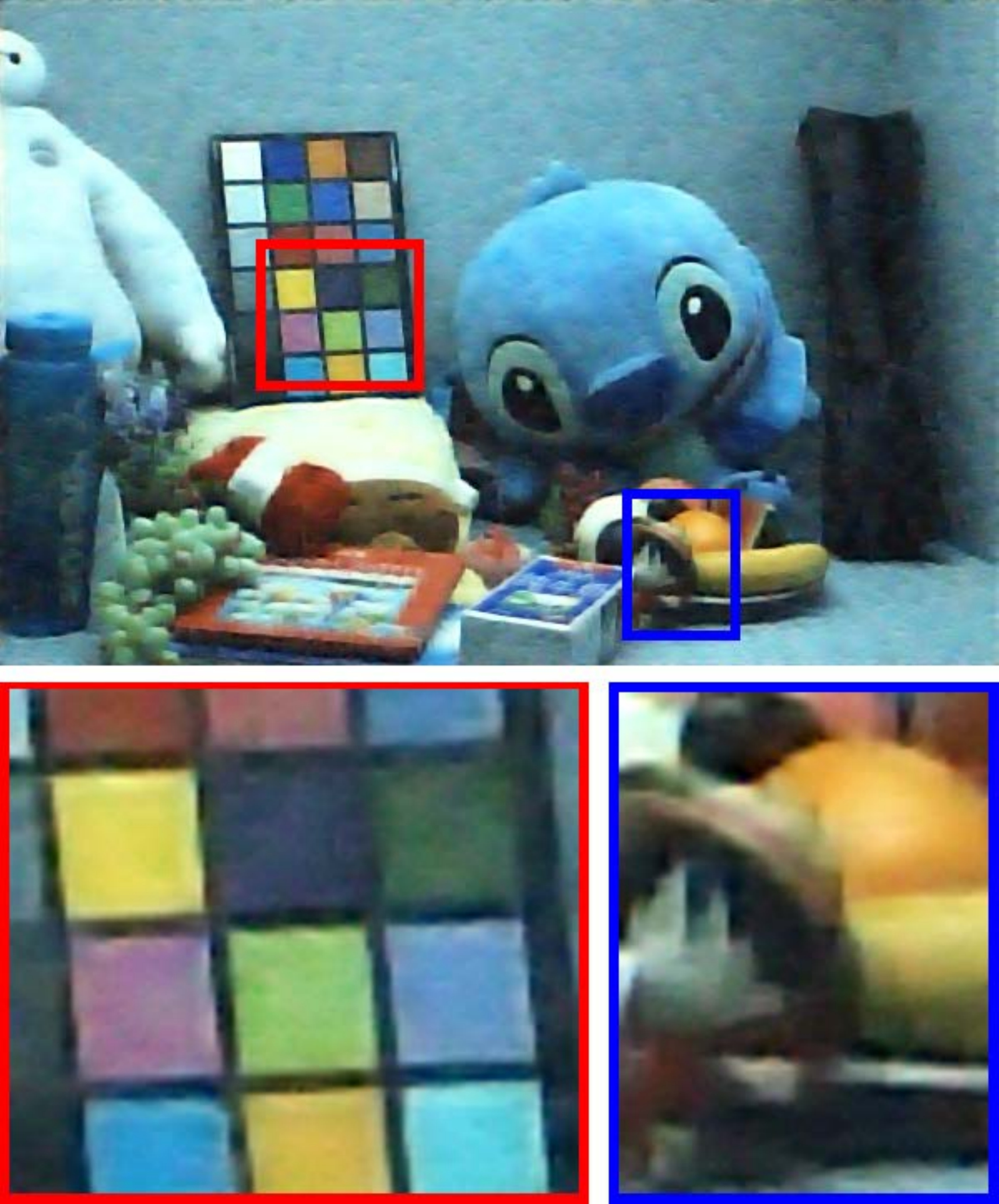}\\
			\footnotesize PSNR/SSIM&\footnotesize 13.013/0.372&\footnotesize \textbf{19.303/0.806}\\
			\includegraphics[width=0.31\linewidth]{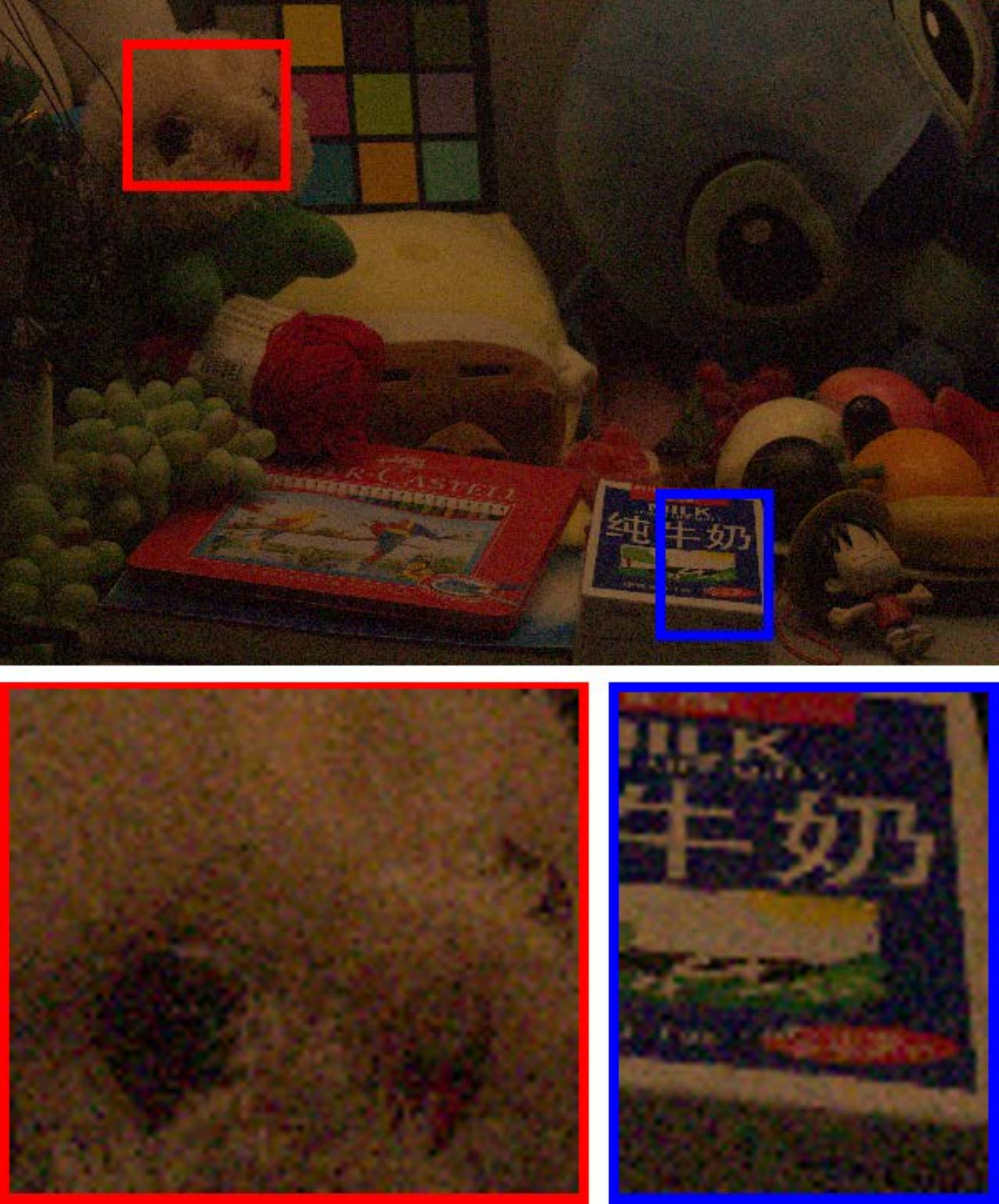}&
			\includegraphics[width=0.31\linewidth]{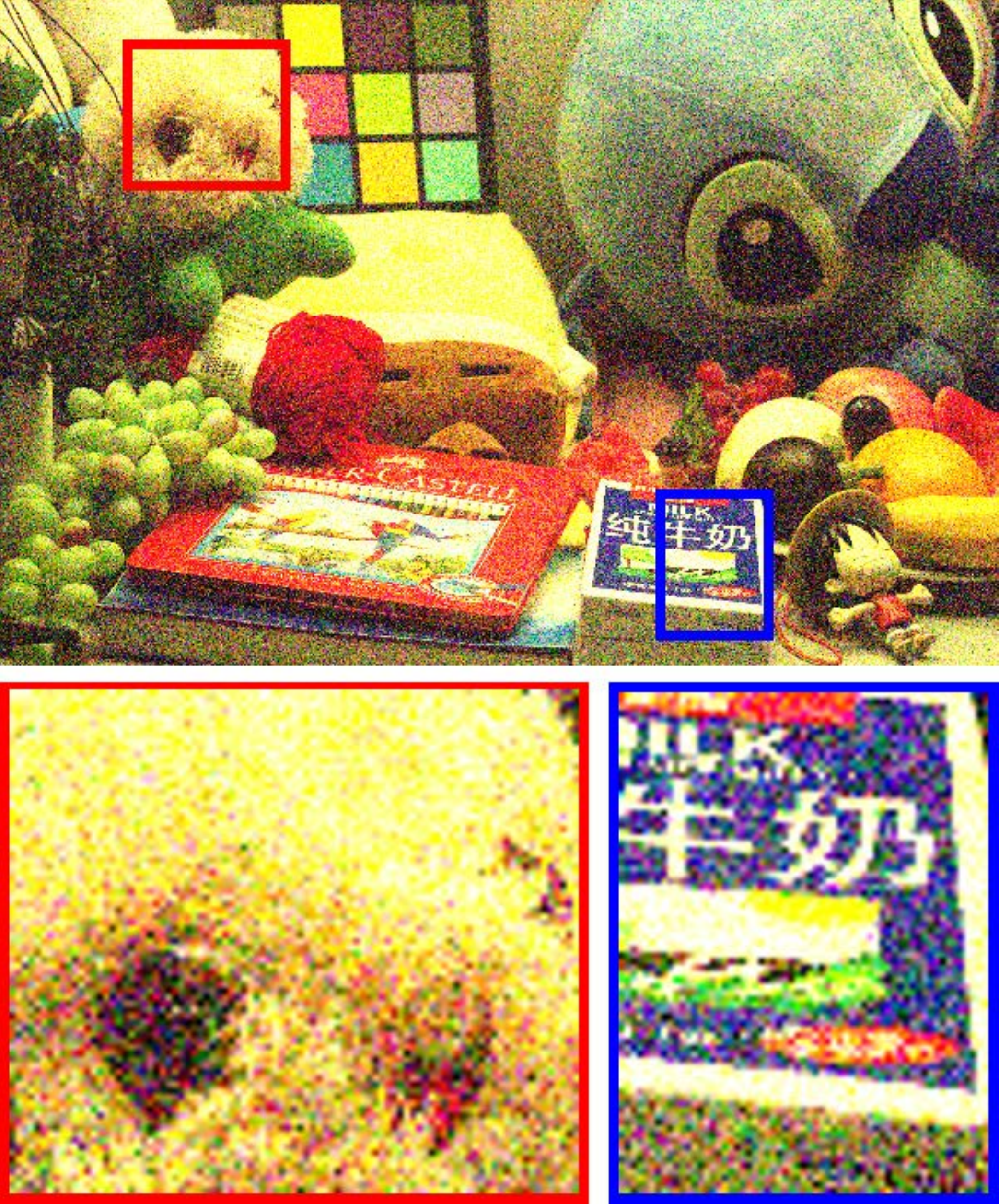}&
			\includegraphics[width=0.31\linewidth]{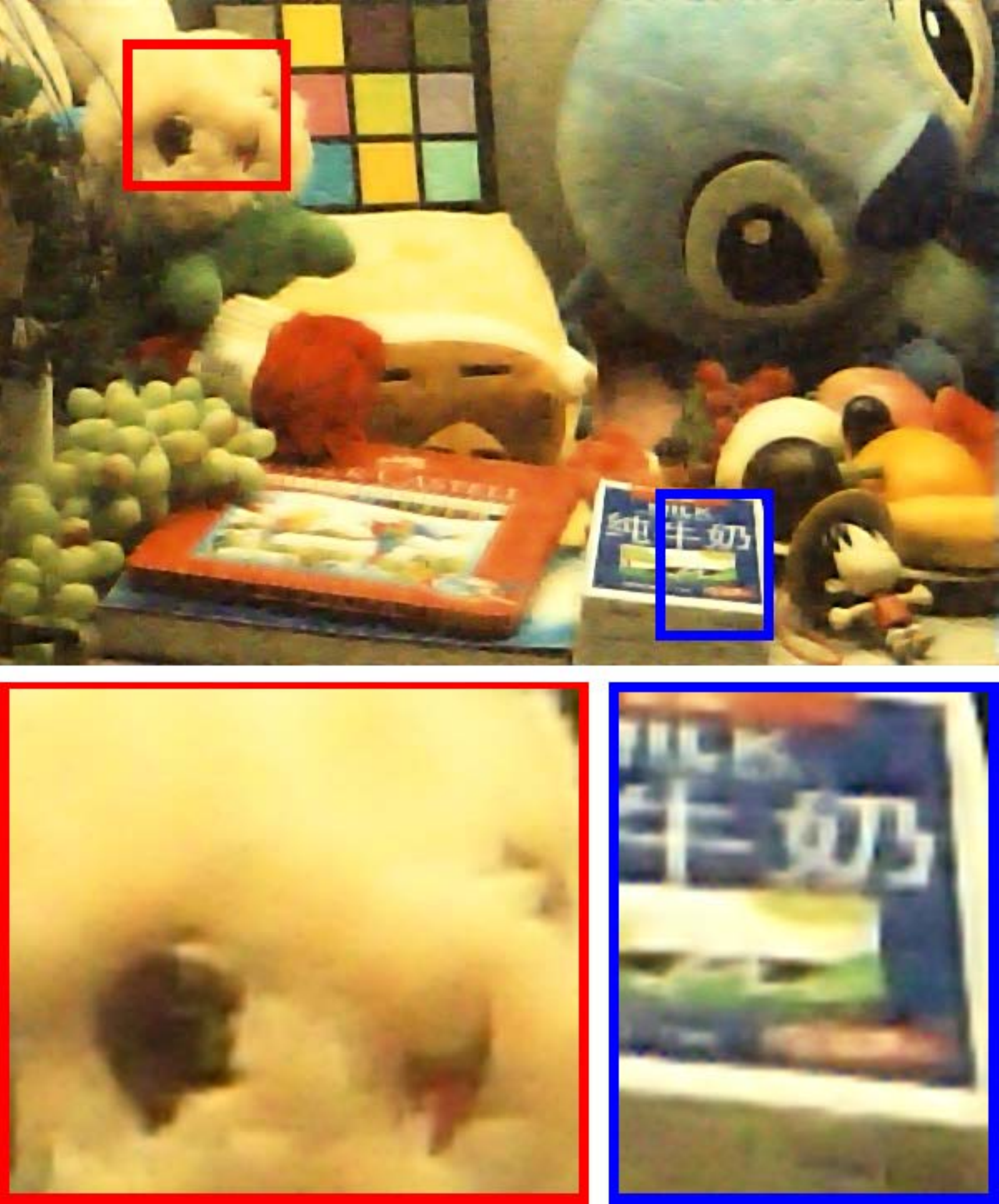}\\
			\footnotesize PSNR/SSIM&\footnotesize 10.820/0.350&\footnotesize \textbf{14.242/0.662}\\
			\footnotesize Input&\footnotesize RUAS$_{\mathtt{i}}$&\footnotesize RUAS$_{\mathtt{i+n}}$\\
		\end{tabular}
	\end{center}
	\vspace{-0.25cm}
	\caption{Ablation study of the contribution of NRM for low-light images in the noisy scenario. Quantitative results (PSNR/SSIM) are reported below each subfigure. Red and blue boxes indicate the obvious differences. }
	\label{fig: NRMeffects}
\end{figure}

Further, we made extensive evaluations on extremely challenging real-world scenarios.
We evaluated our RUAS on the LOL dataset qualitatively and quantitatively, where the LOL dataset contained sensible noises to hinder the enhancement. 
The last two rows in Table~\ref{tab: MITquantitative} illustrated the quantitative comparisons. Obviously, our RUAS obtained the competitive scores. 
Fig.~\ref{fig:LOL} provided a comparison on the LOL dataset. All compared methods failed to take on vivid and real colors. KinD and DRBN indeed removed noises but introduced some unknown artifacts. In contrast, our results presented vivid colors and removed undesired noises.
Then some extreme examples from ExtremelyDarkFace dataset were showed in Fig.~\ref{fig:ExtermelyDarkFace}. These existing advanced networks indeed improved the lightness, but lots of adverse artifacts became visible significantly. The recently-proposed methods, FIDE and DRBN even destroyed the color system, e.g., the overcoat should be red. made great advantages in both detail restoration and noise removal.

Finally, we verified the memory and computation efficiency of our RUAS. Table~\ref{tab: parameters} compared the model size, FLOPs, and running time among different state-of-the-art methods. The FLOPs and running time are calculated on 100 testing images with the size of 600$\times$400 from the LOL dataset. Fortunately, our method just needed very small model size, FLOPs, and time. Note that our RUAS$_{\mathtt{t}+\mathtt{n}}$ was our full version with the noise removal module. Even though its time consuming was a little higher than EnGAN, it is because EnGAN ignored introducing the explicit noise removal module. In a word, Table~\ref{tab: parameters} could fully verify the high-efficiency and fast speed of our method.


%
%

\begin{figure}[t]
	\begin{center}
		\begin{tabular}{c@{\extracolsep{0.4em}}c}
			\includegraphics[width=0.48\linewidth]{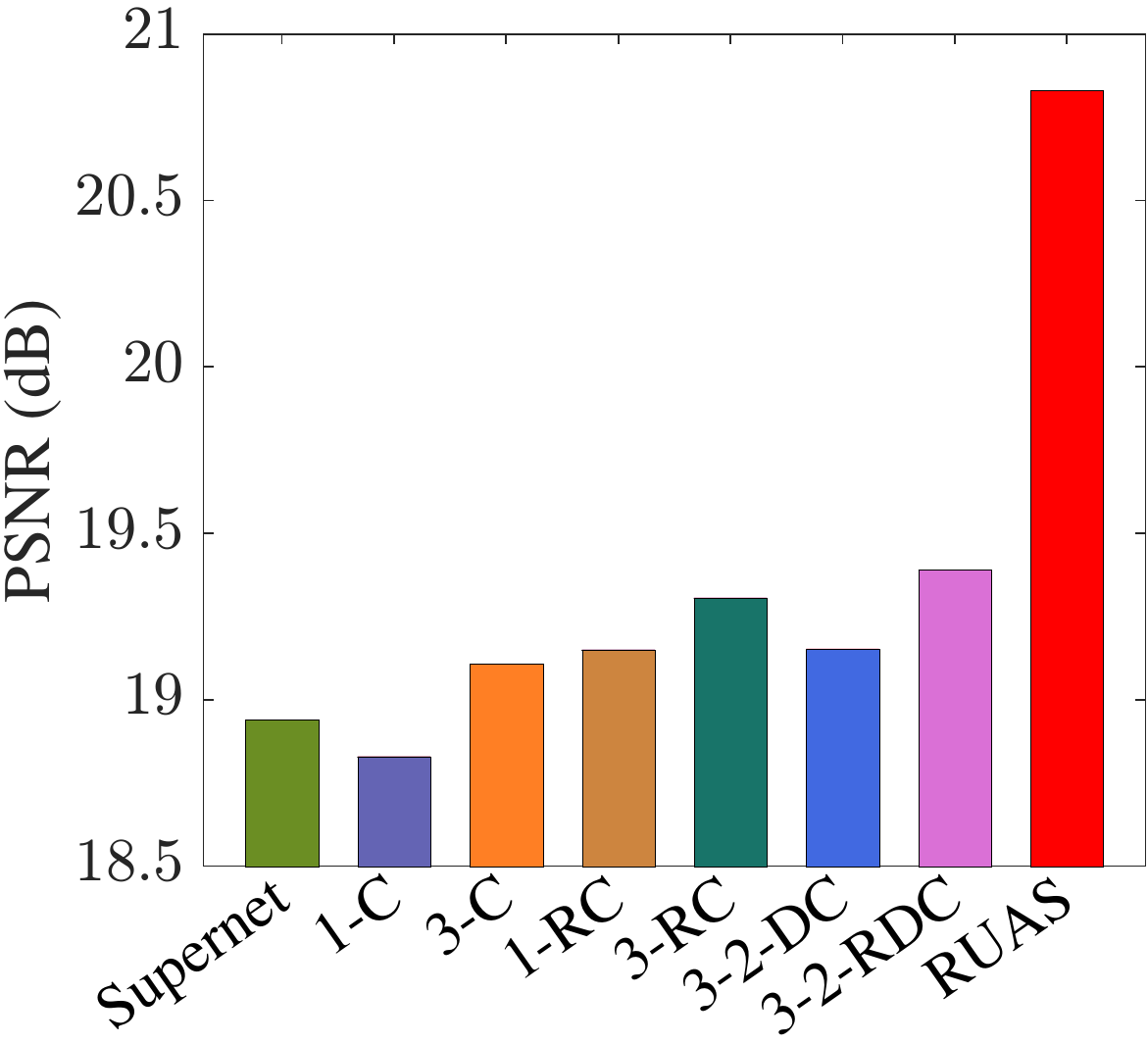}&
			\includegraphics[width=0.48\linewidth]{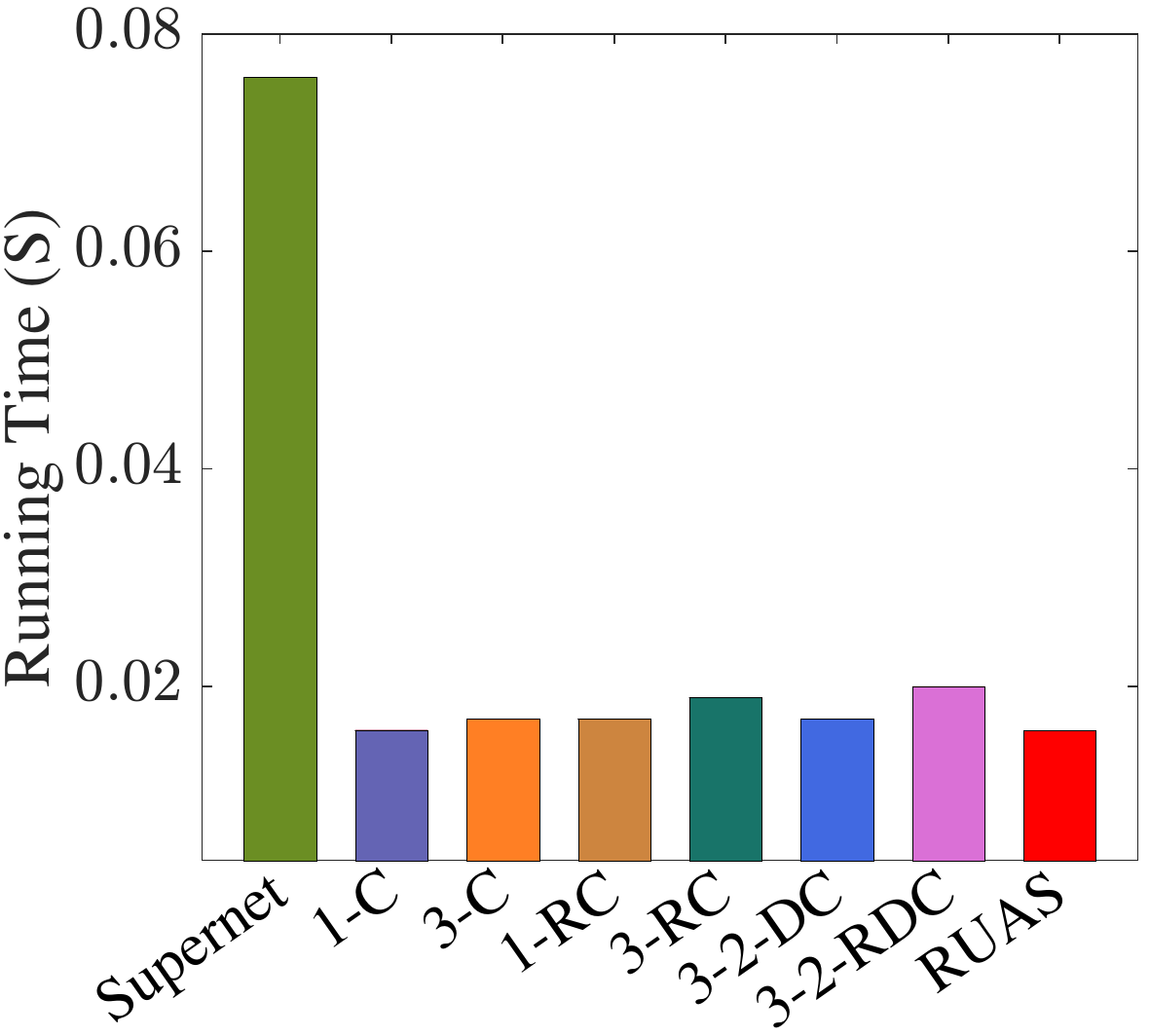}\\
			\footnotesize (a) PSNR &\footnotesize (b) Running Time (S)\\
		\end{tabular}
	\end{center}
	\vspace{-0.2cm}
	\caption{Comparison between naively determined architectures (with supernet and single type of convolution) and our searched RUAS on MIT-Adobe 5K dataset. Quantitative results (PSNR) and running time are plotted for these different architectures. }
	\label{fig:differentcells}
\end{figure}


\subsection{Analysis and Discussions}

Firstly, we evaluated the performance brought by three different warm-start strategies including fix warm-start as $\hat{\mathbf{t}}_0$, update $\hat{\mathbf{t}}_k$ w/o and w/ residual rectification. 
Fig.~\ref{fig: teffect} provided a visual comparison of these warm-start strategies in terms of different components.  
We could observe that the updating strategy of w/o residual rectification indeed suppressed the over-exposure by comparing it with the naive fix warm-start strategy. Further, by introducing the mechanism of residual rectification, the enhanced performed a more comfortable exposure (see the lampshade) than those using other strategies. In a word, we are able to confirm that our designed warm-start strategy really suppress the over-exposure during the propagations. 

We then explored the contribution of NRM, which was designed for improving the adaptability in noisy scenarios. Fig.~\ref{fig: NRMeffects} provided the visual comparison on an example that contained intensive noises hidden in the dark. After performing RUAS$_\mathtt{i}$, we enhanced the image details by significantly improving the brightness. However, the visible noises appeared during enhancement period harmfully damaged the image quality. By introducing the mechanism NRM, our RUAS$_\mathtt{i+n}$ successfully removed the undesired noises to further improved the visual quality and numerical scores. This experiment fully verified the necessity of introducing NRM in some complex noisy real-world scenarios. 

Subsequently, we analyzed the performance of different heuristically-designed architectures on the MIT-Adobe 5K dataset. As shown in Fig.~\ref{fig:differentcells}, even though adopted the complex supernet that contained massive parameters, the results performed the low PSNR and high time-consuming. As for other cases,the performance was also unideal. Briefly, these architectures may not effective enough. It was because these architectures did not integrate the task cues/principles. By comparison, our searched architecture realized the highest PSNR with less inference time. In conclusion, this experiment indicated the necessity of searching for architecture and the superiority of our searched architecture.

\begin{figure}[t]
	\begin{center}
		\begin{tabular}{c@{\extracolsep{0.5em}}c}
			\includegraphics[width=0.48\linewidth]{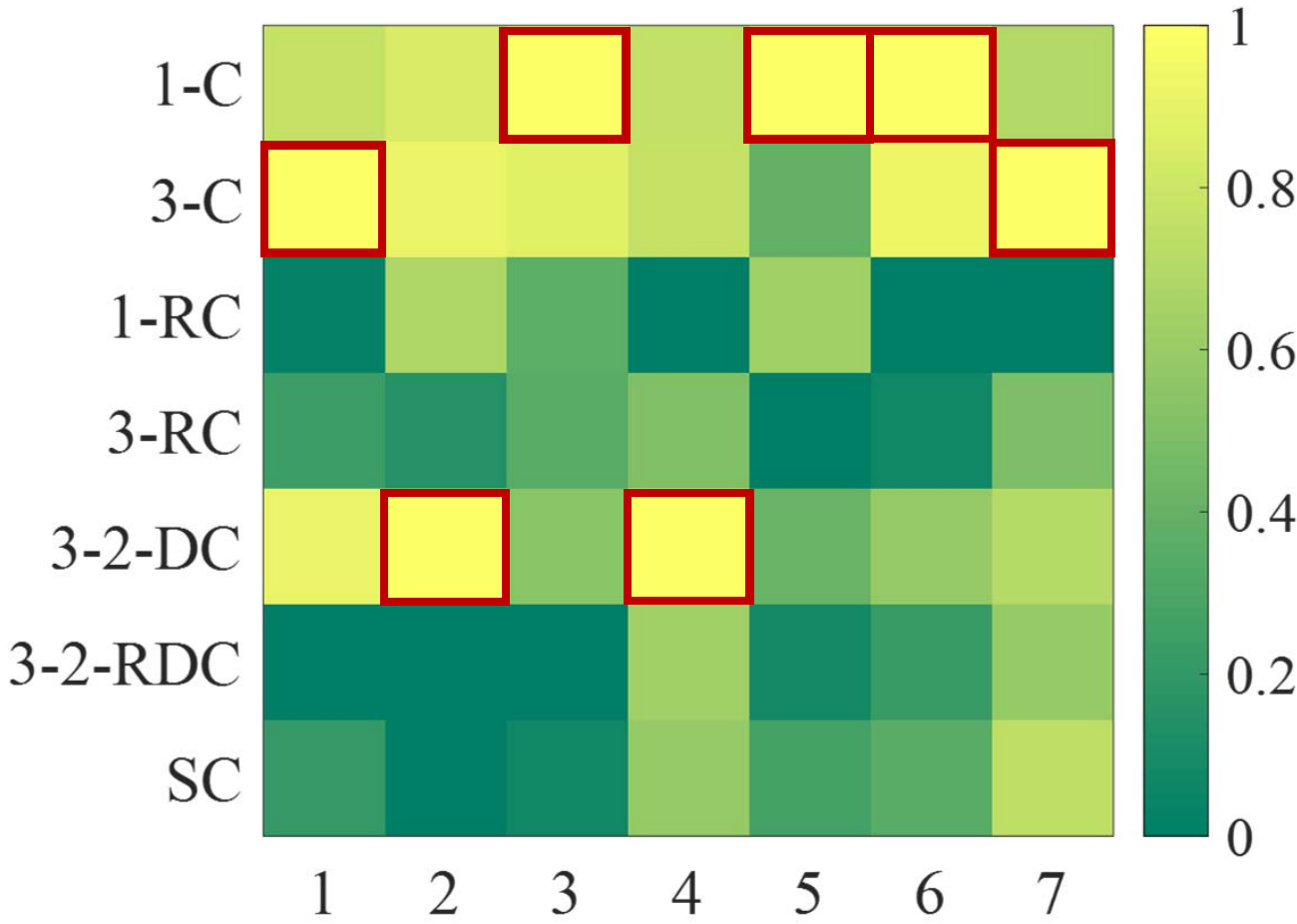}&
			\includegraphics[width=0.48\linewidth]{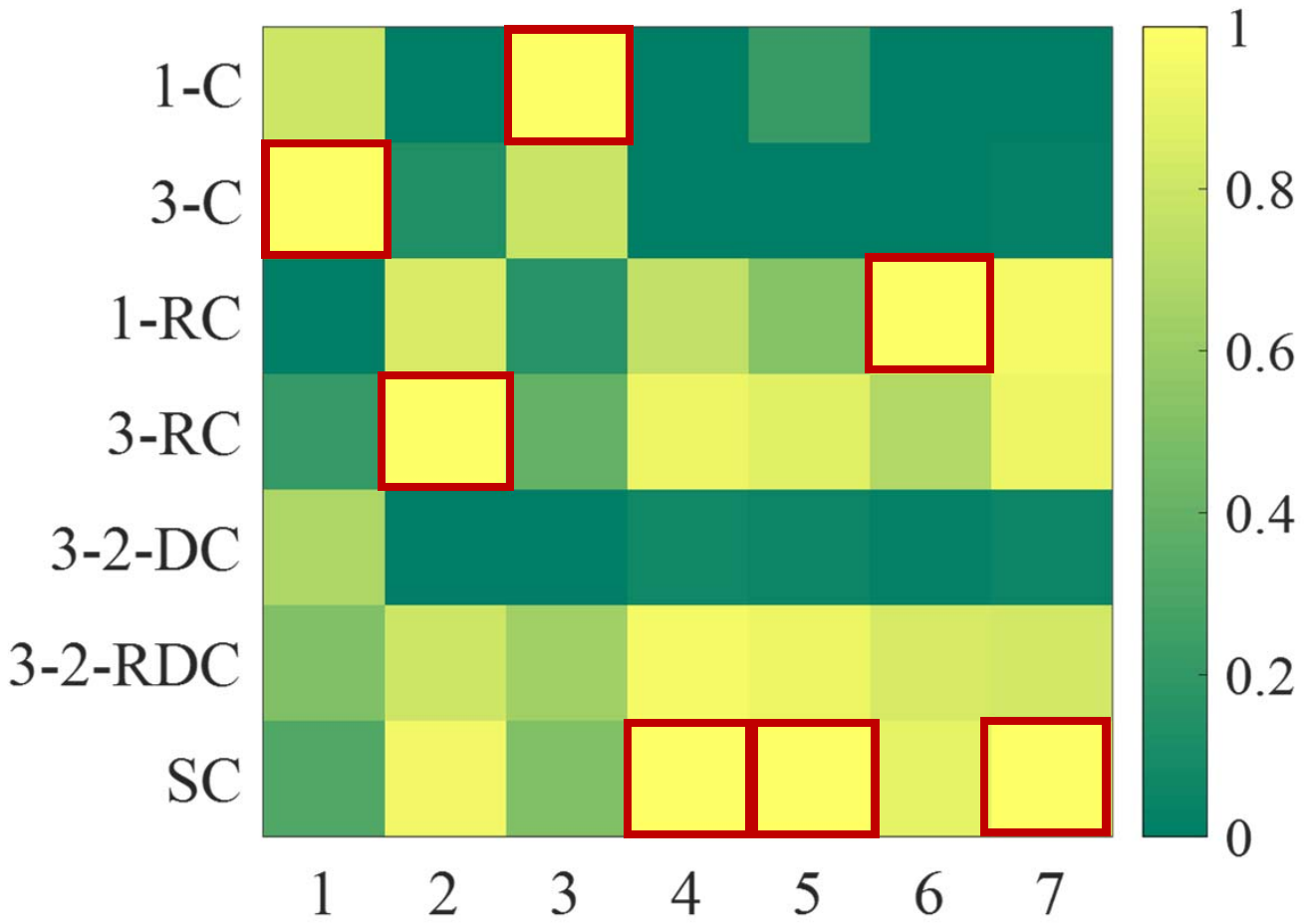}\\
			\multicolumn{2}{c}{\footnotesize (a) Separate Search (15.841/0.584)}\\
			\includegraphics[width=0.48\linewidth]{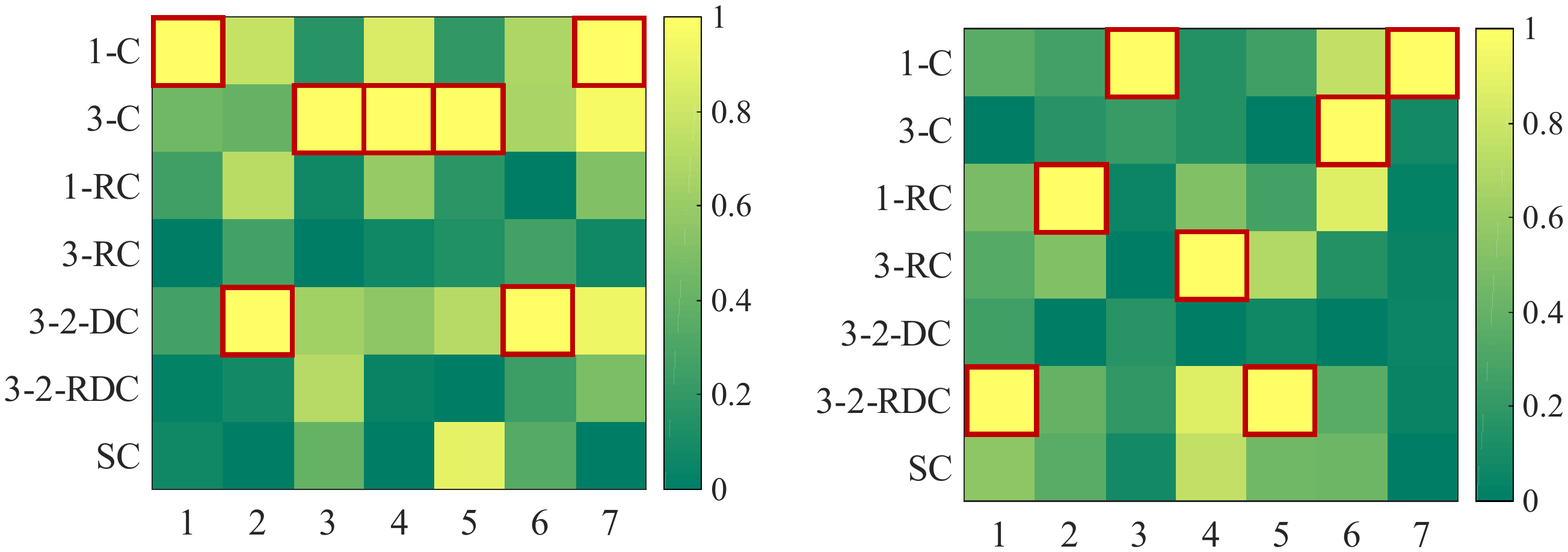}&
			\includegraphics[width=0.48\linewidth]{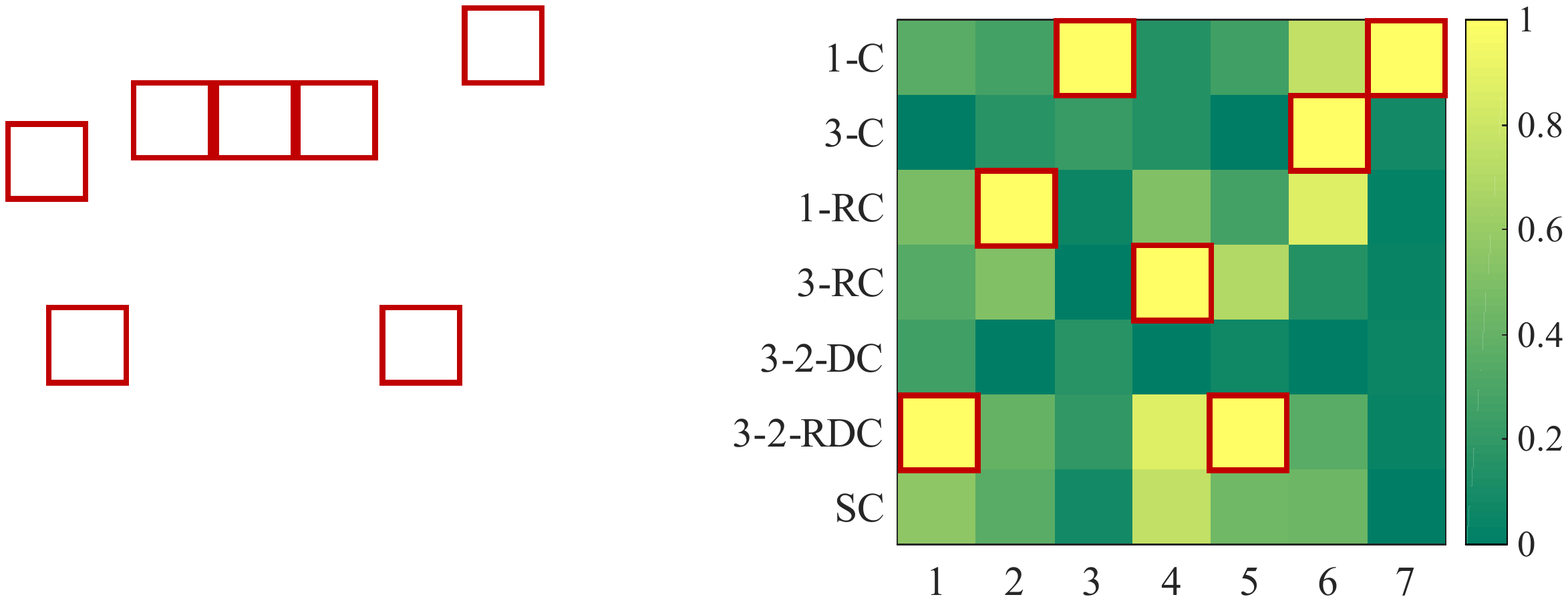}\\
			\multicolumn{2}{c}{\footnotesize (b) Naive Joint Search (14.496/0.543)}\\
			\includegraphics[width=0.48\linewidth]{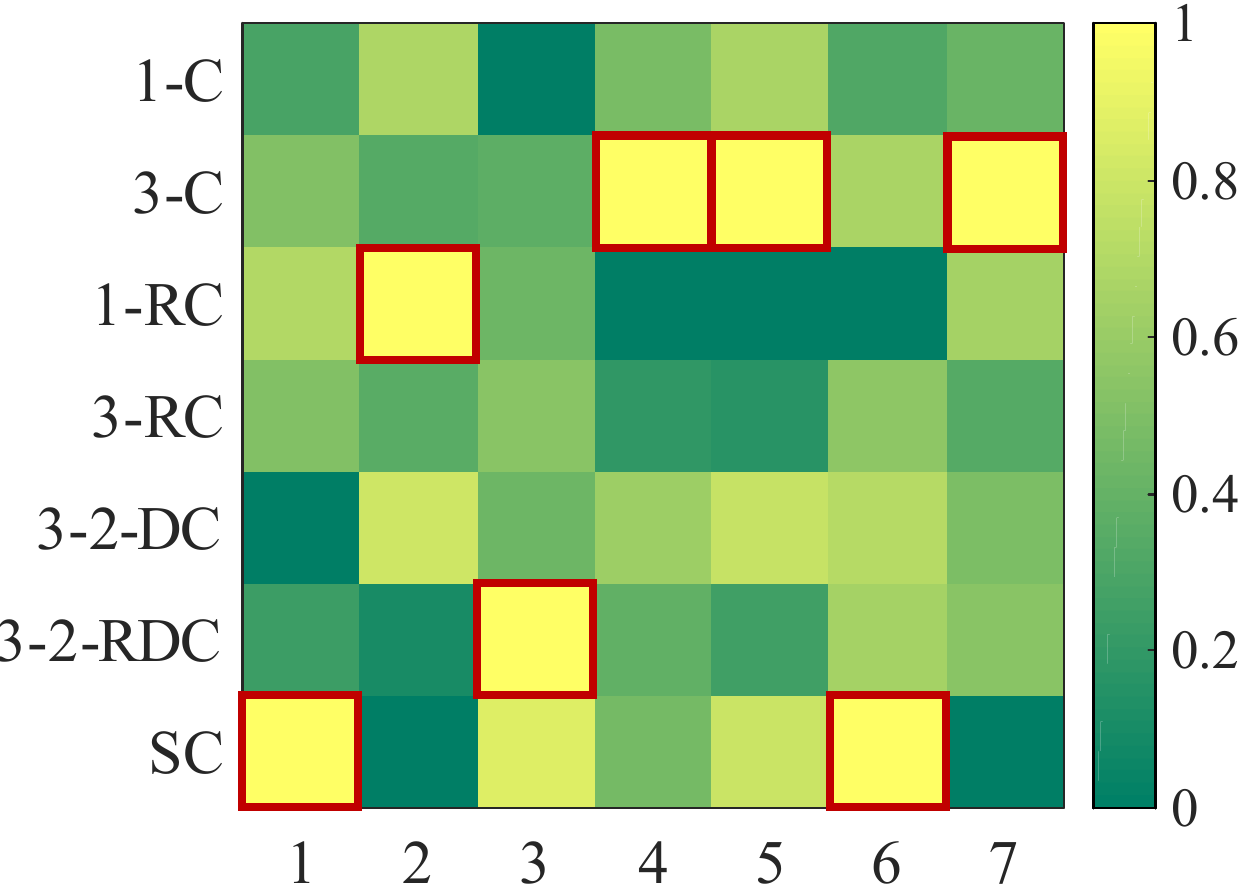}&
			\includegraphics[width=0.48\linewidth]{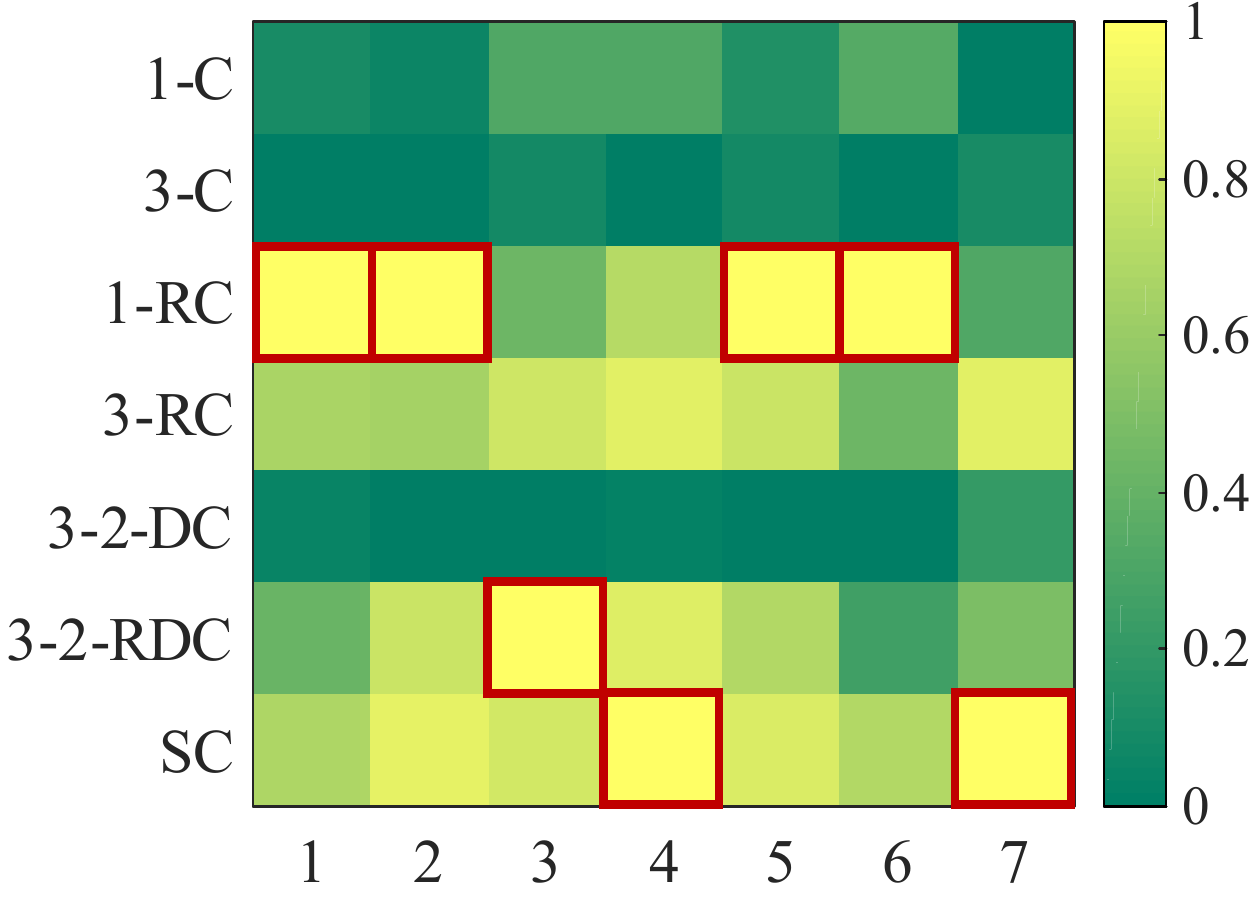}\\
			\multicolumn{2}{c}{\footnotesize (c) Cooperative Joint Search (\textbf{18.226/0.717})}\\
		\end{tabular}	
	\end{center}
	\vspace{-0.15cm}
	\caption{Heatmaps of these candidate architectures (i.e., $\balpha$) in the last searching epoch. Red boxes indicate our searched architectures (with the highest score). Since we share cell for all the stages, thus only one cell is plotted. The left and right columns are the results of IEM and NRM, respectively. Quantitative results (PSNR/SSIM) are reported accordingly.} 
	\label{fig: SearchStrategy}
\end{figure}

Actually, the search strategy was a decisive factor for the performance of the searched architecture. To this end, we made an evaluation in terms of using different searching strategies. We considered three search strategies based on how to search the IEM and NRM. The separate search strategy was to search these two parts one by one. That is, when searching the NRM, the searching procedure of the IEM had ended. The naive joint search was to view IEM and NRM as part of an entire architecture and just needed to search for all the architecture once. As shown in Fig.~\ref{fig: SearchStrategy}, our strategy was significantly effective for the numerical scores. In addition, from the searched architecture, we could see that our searched NRM contains more residual convolution and skip connection, it is reasonable because it has been proved in some existing denoising works~\cite{zhang2017beyond}. In other words, our cooperative strategy indeed bridges the illumination estimation and denoising to realize a valuable collaboration.

\section{Conclusion}
In this study, we proposed a new framework to integrate the principled optimization unrolling technique with a cooperative prior architecture search strategy for designing effective yet lightweight low-light enhancement network. We first established optimization models based the Retinex rule to formulate the latent structures of the illumination map and our desired image. By unrolling the iteration process with abstract deep priors, we can obtain the holistic structure of our enhancement network. Then we developed a cooperative and reference-free strategy to discover specific architectures from a compact search space. Our experiments were performed on a series of challenging benchmark datasets and we derived new state-of-the-art results.

{\small
\bibliographystyle{ieee_fullname}
\bibliography{egbib}
}
\end{document}